\documentclass[12pt,fleqn]{ucithesis}

\usepackage{amsmath}  
\usepackage{amsfonts} 
\usepackage{amssymb}
\usepackage{array}
\usepackage{graphicx}
\usepackage{natbib}
\usepackage{relsize}
\usepackage[titletoc]{appendix}

\usepackage{caption}
\usepackage{subcaption}  
\usepackage{multirow}
\usepackage{tabularx}
\usepackage{booktabs}
\usepackage{graphicx}
\usepackage{placeins}
\usepackage{xcolor}

\usepackage[plainpages=false,pdfborder={0 0 0}]{hyperref}

\newcommand*{\defeq}{\stackrel{\text{\tiny{}def}}{=}}

\thesistitle{Deep Latent Variable Modeling of Physiological Signals}

\documenttitle{Dissertation}

\degreename{Doctor of Philosophy}

\degreefield{Computer Science}

\authorname{Khuong An Vo}

\committeechair{Professor Nikil Dutt}
\othercommitteemembers
{
  Associate Professor Hung Cao \\
  Professor Ramesh Srinivasan \\
  Professor Erik Sudderth
}

\degreeyear{2024}

\copyrightdeclaration
{
}

\dedications
{
  This thesis is dedicated to my parents, sister, and brother for their unwavering love, support, and encouragement.
}

\acknowledgments
{
This PhD journey has been the most challenging yet highly fulfilling experience of my life. Achieving this milestone would not have been possible without the individuals who have become part of it along the way. 

First of all, I would like to thank the members of my thesis committee for their invaluable feedback, constructive criticism, and unwavering support throughout this process. I am grateful to Prof. Nikil Dutt for his warm welcome to UCI and for his guidance in research from day one. His advice on presentations, paper writing, and networking has significantly improved my skills as a researcher. I thoroughly enjoyed our discussions on energy-efficient deep learning. I extend my heartfelt thanks to Prof. Hung Cao for offering me numerous opportunities for both academic and personal growth. Prof. Cao inspired my work on AI in healthcare, involving me in multidiscplinary teams where I learned extensively, leading to many fruitful collaborations and ideas for commercializing research projects. I am thankful to Prof. Ramesh Srinivasan for his guidance on challenging yet intellectually stimulating neurocognitive projects. Working with him allowed me to delve deeply into the most complex problems in AI and neuroscience. His patience, amiable advising approach, and expert knowledge have been immensely helpful in overcoming numerous significant challenges in the research. I deeply appreciate Prof. Erik Sudderth for his insightful feedback and suggestions that have greatly influenced my research. His course on Learning in Graphical Models provided the essential groundwork for the main topic of this thesis.

I also would like to express my gratitude to my supervisors at Samsung Device Solution Research America, Dr. Mostafa El-Khamy and Dr. Yoojin Choi, for their extensive guidance and support, which extended beyond the duration of my internship. Portions of this dissertation were carried out during my time as an intern.

Before embarking on my PhD journey, I was privileged to have met and been inspired by several accomplished scholars, particularly Prof. Khanh Dang, Prof. Tho Quan, Mao Nguyen, and Minh Truong. Their influence motivated me to engage in research and pursue graduate studies in the United States.

A special thanks goes to my friends and fellow Ph.D. candidates at UCI, whose collaboration and camaraderie made this journey a memorable one. These include Michale Lee, Tai Le, Jenny Sun, Manoj Vishwanath, Emad Kasaeyan Naeini, Isaac Menchaca, Floranne Ellington, Amir Naderi, Alex Huang, Sadaf Sarafan, Ramses Torres, Anh Nguyen, Steven Cao, and many others.

I am grateful to the National Science Foundation (NSF) and the National Institutes of Health (NIH) for providing the financial support that enabled me to pursue and complete this research. Without their generosity, this work would not have been possible. Funding resources: NSF grant \#1917105, \#1658303, \#1850849, \#2051186, \#2126976, and NIH SBIR grant \#R44OD024874.

}

\newcommand{\mypubentry}[3]{
  \begin{tabular*}{1\textwidth}{@{\extracolsep{\fill}}p{4.5in}r}
    \textbf{#1} & \textbf{#2} \\ 
    \multicolumn{2}{@{\extracolsep{\fill}}p{.95\textwidth}}{#3}\vspace{6pt} \\
  \end{tabular*}
}

\curriculumvitae
{

\pagebreak

\textbf{REFEREED JOURNAL PUBLICATIONS}

  \mypubentry{Deep latent variable joint cognitive modeling of neural signals and human behavior}{2024}{NeuroImage}
  \mypubentry{A novel wireless ECG system for prolonged monitoring of multiple zebrafish for heart disease and drug screening studies}{2022}{Biosensors and Bioelectronics}
  \mypubentry{Deep learning-based framework for cardiac function assessment in embryonic zebrafish from heart beating videos}{2021}{Computers in Biology and Medicine}
  \mypubentry{Continuous non-invasive blood pressure monitoring: a methodological review on measurement techniques}{2020}{IEEE Access}
  \mypubentry{An efficient and robust deep learning method with 1-D octave convolution to extract fetal electrocardiogram}{2020}{Sensors}

\textbf{REFEREED CONFERENCE PUBLICATIONS}

  \mypubentry{PPG-to-ECG signal translation for continuous atrial fibrillation detection via attention-based deep state-space modeling}{2024}{International Conference of the IEEE Engineering in Medicine and Biology Society}
  \mypubentry{Composing graphical models with generative adversarial networks for EEG signal modeling}{2022}{IEEE International Conference on Acoustics, Speech and Signal Processing}
  \mypubentry{Decision SincNet: Neurocognitive models of decision making that predict cognitive processes from neural signals}{2022}{International Joint Conference on Neural Networks}
  \mypubentry{P2E-WGAN: ECG waveform synthesis from PPG with conditional Wasserstein generative adversarial networks}{2021}{ACM Symposium on Applied Computing}
  \mypubentry{STINT: Selective transmission for low-energy physiological monitoring}{2020}{ACM/IEEE International Symposium on Low Power Electronics and Design}

}

\thesisabstract
{ \vspace{-25pt}
  A deep latent variable model is a powerful method for capturing complex distributions. These models assume that underlying structures, but unobserved, are present within the data.
  In this dissertation, we explore high-dimensional problems related to physiological monitoring using latent variable models.
  First, we present a novel deep state-space model to generate electrical waveforms of the heart using optically obtained signals as inputs. This can bring about clinical diagnoses of heart disease via simple assessment through wearable devices. 
  Second, we present a brain signal modeling scheme that combines the strengths of probabilistic graphical models and deep adversarial learning. The structured representations can provide interpretability and encode inductive biases to reduce the data complexity of neural oscillations. The efficacy of the learned representations is further studied in epilepsy seizure detection formulated as an unsupervised learning problem.
  Third, we propose a framework for the joint modeling of physiological measures and behavior. Existing methods to combine multiple sources of brain data provided are limited. Direct analysis of the relationship between different types of physiological measures usually does not involve behavioral data. 
  Our method can identify the unique and shared contributions of brain regions to behavior and can be used to discover new functions of brain regions.
  The success of these innovative computational methods would allow the translation of biomarker findings across species and provide insight into neurocognitive analysis in numerous biological studies and clinical diagnoses, as well as emerging consumer applications.
}

\hypersetup{
	pdftitle={\Thesistitle},
	pdfauthor={\Authorname},
	pdfsubject={\Degreefield},
}

\begin{document}

\preliminarypages

\chapter{Introduction}

\section{Motivation}

The modeling of physiological signals stands as a cornerstone in contemporary neuroscience and cardiology research, providing a powerful framework for understanding the intricate processes underlying the generation of signals in the heart and brain. These models, often constructed using computational methods, offer invaluable insights into the complex interactions among biological variables and serve as invaluable tools for hypothesis testing and experimental design. 
At the heart of physiological modeling lie equations that describe the interrelations among variables governing the dynamics of synthetic time series. These equations, typically derived from existing data or informed by domain expertise, describe the temporal evolution of physiological phenomena. By systematically altering parameters within these equations, researchers can explore a wide range of scenarios, uncovering novel insights and validating theoretical predictions.
Previous studies in neuroscience and cardiology have leveraged computational models \citep{niederer2019computational, glomb2020computational} providing principal laws, empirically validated rules, or other domain expertise, typically presented as general, time-dependent, and nonlinear partial differential equations.


With regard to electroencephalogram (EEG) signals, neural mass models represent a prominent paradigm for understanding the complex dynamics of large populations of neurons. One such influential model is the Jansen-Rit model \citep{jansen1993neurophysiologically, jansen1995electroencephalogram}. This model emerged as a lumped parameter representation of a cortical column, specifically designed to capture the dynamics of human EEG rhythms and visual evoked potentials. Building upon earlier work by Lopes da Silva and Katznelson, the Jansen-Rit model retains non-linearities within the cortical column, thus offering a more biologically realistic description of neuronal activity. At its core, the Jansen-Rit model is based on the interaction between two distinct populations of neurons within the cortical column: pyramidal cells and local excitatory/inhibitory interneurons. Each population contributes to the generation and modulation of EEG signals through intricate synaptic connections and feedback loops. The model describes the dynamics of postsynaptic potentials using a set of coupled differential equations, which are formulated to capture the complex interplay between excitatory and inhibitory neuronal populations. These equations are typically rewritten as a system of first-order differential equations, resulting in a six-dimensional dynamical system that encapsulates the behavior of the cortical column.

In terms of the electrocardiogram (ECG) signal, the \citet{mcsharry2003dynamical} model is a significant contribution, providing a framework for understanding the complex dynamics of cardiac rhythms. This model provides a mathematical description of the electrical activity of the heart through a system of three ordinary differential equations. Central to the model is the recognition of the distinct components of the ECG waveform, each of which corresponds to specific cardiac events and functions. The typical sequence begins with the P wave, representing atrial depolarization, followed by the QRS complex, reflecting ventricular depolarization, and concludes with the T wave, indicating ventricular repolarization. By capturing the temporal relationships between these waveform components, the model provides insights into the underlying physiological processes driving cardiac activity. Moreover, the constructed simulator offers the flexibility to manipulate various attributes of the produced ECG signals. Researchers can adjust parameters such as the interval between waves, the magnitude of P-waves and Q-waves, and the average and standard deviation of heart rate patterns. Additionally, the model allows for the exploration of frequency-domain aspects of heart rate variability, providing insights into the dynamic regulation of cardiac rhythm.

Although electrophysical models based on differential equations have been instrumental in understanding physiological signals, they come with inherent limitations. These models often rely on strong assumptions, can be computationally intensive, and may suffer from model misspecification issues. Consequently, there is a growing emphasis on developing more powerful and flexible models capable of handling diverse types of medical time series data. Medical time series data exhibit complex multidimensional dependencies, including spatio-temporal dependencies in biosignals and multimodal dependencies across physiological measures and behaviors. One of the key challenges is effectively modeling spatio-temporal dependencies in biosignals. Biosignals such as EEG and ECG are inherently dynamic and exhibit spatial variations across different body regions. Another challenge lies in integrating multiple modalities of medical data. In modern healthcare settings, patient data often comprise a diverse array of measurements from different sensors and modalities, including physiological signals, imaging data, and clinical observations. These presents significant challenges for modeling and analysis in biomedical research and clinical practice. 

In recent decades, the exponential growth of data collection in various medical applications has paved the way for the emergence of data-driven approaches capable of unlocking valuable insights and addressing complex challenges in healthcare. These approaches leverage advanced computational techniques to analyze large volumes of data, uncover underlying structures, and utilize extracted information for tasks such as predictive modeling and pattern recognition. A significant breakthrough in this domain has been the integration of probabilistic modeling and deep learning techniques. This fusion of methodologies combines the expressive power of deep neural networks with the probabilistic framework, enabling the parameterization of rich probabilistic distributions over latent variables. By incorporating established or desired inductive biases, these models can effectively capture the complex relationships and uncertainties inherent in medical data. Two prominent classes of models that have propelled recent progress are variational autoencoders (VAEs) and generative adversarial networks (GANs), both belonging to the broader category of deep generative models. These models offer scalable and efficient solutions for unsupervised learning of complex, high-dimensional data distributions.

This thesis aims to explore a diverse category of sequential and multimodal models, with a particular focus on their application to complex spatio-temporal physiological measures. By leveraging these models, we seek to unlock valuable insights from unlabeled datasets, paving the way for a deeper understanding of dynamic physiological processes.

\section{Outline and contributions}

\textbf{Chapter 2} first presents the pertinent techniques employed in both clinical practice and research for physiological monitoring, offering crucial insights into the body's internal condition. We focus on biosignals obtained through sensors either on or inside the body, like surface ECG and EEG. Subsequently, we introduce probability theory, graphical models, and latent variable models, which underpin all the methods discussed later in this dissertation.

\textbf{Chapter 3} presents a sequential modeling of ECG signals from photoplethysmography (PPG). PPG is a cost-effective and non-invasive technique that utilizes optical methods to measure cardiac physiology. PPG has become increasingly popular in health monitoring and is used in various commercial and clinical wearable devices. Compared to electrocardiography (ECG), PPG does not provide substantial clinical diagnostic value, despite the strong correlation between the two. Here, we propose a subject-independent attention-based deep state-space model (ADSSM) to translate PPG signals to corresponding ECG waveforms. The model is not only robust to noise but also data-efficient by incorporating probabilistic prior knowledge. To evaluate our approach, 55 subjects' data from the MIMIC-III database were used in their original form, and then modified with noise, mimicking real-world scenarios. Our approach was proven effective as evidenced by the PR-AUC of 0.986 achieved when inputting the translated ECG signals into an existing atrial fibrillation (AFib) detector. ADSSM enables the integration of ECG's extensive knowledge base and PPG's continuous measurement for early diagnosis of cardiovascular disease.

\textbf{Chapter 4} presents a deep generative model of EEG signals that provides not only a stochastic procedure that directly generates data but also insights to further understand the neurological mechanisms. Specifically, we propose a generative and inference approach that combines the complementary benefits of probabilistic graphical models and GANs for EEG signal modeling. We investigate the method's ability to jointly learn coherent generation and inverse inference models on the CHI-MIT epilepsy multi-channel EEG dataset. We further study the efficacy of the learned representations in epilepsy seizure detection formulated as an unsupervised learning problem.

\textbf{Chapter 5} introduces a method for joint cognitive modeling of neural signals and human behavior. As the field of computational cognitive neuroscience continues to expand and generate new theories, there is a growing need for more advanced methods to test the hypothesis of brain-behavior relationships. Recent progress in Bayesian cognitive modeling has enabled the combination of neural and behavioral models into a single unifying framework. However, these approaches require manual feature extraction, and lack the capability to discover previously unknown neural features in more complex data. Consequently, this would hinder the expressiveness of the models. To address these challenges, we propose a Neurocognitive Variational Autoencoder (NCVA) to conjoin high-dimensional EEG with a cognitive model in both generative and predictive modeling analyses. Importantly, our NCVA enables both the prediction of EEG signals given behavioral data and the estimation of cognitive model parameters from EEG signals. This novel approach can allow for a more comprehensive understanding of the triplet relationship between behavior, brain activity, and cognitive processes.

\textbf{Chapter 6} concludes the main contributions of this dissertation and discusses some directions for future work.
\chapter{Background}

\section{Physiological Signals}

\subsection{Electrocardiogram (ECG or EKG)}
Electrocardiography (ECG) is fundamental in diagnosing and managing cardiac health . By recording the heart's electrical activity, the ECG provides crucial insights into its rhythm, rate, and muscular functionality. This non-invasive tool quantifies voltage differences between points on the body's surface over time, enabling clinicians to assess the heart's performance with precision. At its core, an ECG captures the electrical signals generated by the heart with each beat. These signals, represented graphically as waves and complexes on the ECG tracing, reflect the coordinated sequence of events during the cardiac cycle, including atrial depolarization, ventricular depolarization, and ventricular repolarization. Key among its features is the identification of the R-wave, marking the onset of ventricular contraction that propels blood from the heart into the aorta, a pivotal event in the cardiac cycle. In clinical settings, traditional 12-lead ECG devices provide a comprehensive view of the heart's electrical activity from multiple angles. By delivering 12 concurrent ECG signals, these devices offer invaluable insights into irregularities and pinpoint specific areas of concern within the heart. Recent advancements have seen the integration of single-lead ECG sensors into wrist-worn devices, bringing cardiac monitoring closer to everyday life. These compact gadgets, featuring electrodes on the wrist and an additional point of contact, offer a simplified yet effective means of evaluating heart rhythm and certain functional aspects.

\subsection{Photoplethysmogram (PPG)}

Photoplethysmogram (PPG) signals are obtained through the emission of light from a light-emitting diode (LED) onto the skin, followed by the assessment of the light either reflected back from the skin surface or transmitted through bodily tissues. This fundamental principle underpins its application in a variety of devices, from wrist-worn wearables to medical-grade pulse oximeters. The PPG signal tracks variations in blood volume over time, particularly in arterial blood. As the arterial pulse wave reaches the measurement site, typically at the fingertip or earlobe, it triggers detectable changes in blood volume, generating characteristic waveforms in the PPG signal. Each heartbeat manifests as a distinct peak in the PPG waveform, reflecting the pulsatile nature of blood flow and pressure within the arteries. A major benefit of PPG sensors is their capability to record physiological signals passively, without the need for user involvement, in contrast to wrist-worn ECG sensors which require active user engagement during the collection of signals. This ease of use, coupled with their non-invasive nature, has propelled the widespread adoption of PPG sensors in consumer wearable devices for tracking heart rate, heart rate variability, and cardiac rhythm. However, it is important to recognize that the PPG signal is susceptible to noise, stemming from factors such as motion artifacts, ambient light interference, and variations in skin perfusion.

\subsection{Electroencephalogram (EEG)}

Electroencephalography (EEG) provides invaluable insights into the electrical activity of the brain. By recording the synchronized post-synaptic currents primarily in cortical pyramidal neurons \citep{nunez2006electric}, EEG offers a unique window into cognitive processes, neural dynamics, and neurological disorders. Over the years, EEG has become a cornerstone in both clinical diagnostics and cognitive research, enabling researchers and clinicians to delve deep into the complexities of brain function. Broadly categorized into spontaneous potentials, such as sleep rhythms, and evoked potentials, which are time-locked responses to external stimuli, EEG captures brain activity with high temporal resolution. Operating on the scale of milliseconds, EEG is capable of detecting rapid changes in neural activity, making it a powerful tool for studying dynamic brain processes. However, despite its temporal precision, EEG possesses inherent limitations in spatial resolution. The electrical signals detected at the scalp originate from currents that propagate through head tissues via volume conduction, resulting in a low spatial resolution. While EEG can provide insights into broad patterns of brain activity, its ability to localize specific neural sources is limited. Nevertheless, EEG has found extensive applications in both clinical and research settings. In clinical practice, EEG is used to diagnose and manage various neurological conditions, including epilepsy, sleep disorders, stroke, and Alzheimer's disease. In the realm of cognitive sciences, EEG offers insights into sensorimotor pathways, memory, language processing, and general intelligence. One of the key advantages of EEG lies in its affordability, portability, and suitability for real-time observation. Unlike other brain imaging techniques that require specialized equipment and expertise, EEG can be deployed with minimal resources, making it accessible to a wide range of researchers and clinicians. However, EEG analysis is not without challenges. EEG signals are non-stationary, exhibit a poor signal-to-noise ratio, and exhibit high variability among individuals, posing significant obstacles to the development of generalized models for EEG analysis.

\subsection{Frequency Domain Representation}

A univariate time series signal of length $T$ consists of a sequence of real-valued data points $\boldsymbol{x}=\left(x_0, \ldots, x_{T-1}\right) \in \mathbb{R}^T$, each representing observations of a specific phenomenon. Observations in a time series are generally interdependent, and grasping this dependency is crucial for recognizing various phenomena as they appear. In the frequency domain, a time series is analyzed by decomposing it into sinusoids that vary in amplitude and phase.

Assuming the periodic nature of the time series $\boldsymbol{x} = (x_0, \ldots, x_{T-1})$, we can represent each element $x_t$ with an equation

\begin{equation}
x_t=\sum_{k=0}^{T-1} X_k e^{\frac{j 2 \pi k t}{T}}
\end{equation}

where $X_k \in \mathbb{C}, k=0, \ldots, T-1$, represent the Fourier coefficients. Each element $x_t$ in the time series is broken down into $T$ frequency components $e^{j 2 \pi k t / T}, k=0, \ldots, T-1$, each scaled by the Fourier coefficients. In the discrete FT (DFT), the time series $\boldsymbol{x}$ is expressed via the Fourier coefficients $X_k=(1 / T) \boldsymbol{v}_k^* \boldsymbol{x}, \quad k=0, \ldots, T-1$, where $\boldsymbol{v}_k:=\left[1, e^{j 2 \pi k(1) / T}, e^{j 2 \pi k(2) / T}, \ldots, e^{j 2 \pi k(T-1) / T}\right]$. The collection of vectors $\left\{\boldsymbol{v}_0, \ldots, \boldsymbol{v}_{T-1}\right\}$ forms an orthogonal basis for the $T$-dimensional complex vector space. Each Fourier coefficient thus serves as an independent representation of a subcomponent of the entire time series. The conversion of the time series into the frequency domain can be efficiently performed using the fast FT (FFT) algorithm, while the transformation from the frequency domain back to the time domain is accomplished using the inverse DFT.

\section{Probabilistic Graphical Models}
\subsection{Random Variables and Probabilities}

\textit{Definition 1 (Random Variable).}

In a random experiment with a sample space \(\mathcal{S}\), a function \(X\) maps each element \(s \in \mathcal{S}\) to a single real number \(X(s)=x\). This function \(X\) is known as a random variable (r.v.). 

Typically, random variables are represented by capital letters, while the values they take are denoted by lowercase letters. The term univariate distribution will be used to describe the distributions of a single random variable, indicated by the non-bold $x$. The term multivariate distributions will apply to distributions involving multiple random variables, typically represented as a vector with bold $\boldsymbol{x}$.

\textit{Definition 2 (Discrete Random Variable and Probability Mass Function).}

\begin{enumerate}
\item A random variable \(X\) is termed discrete if its range consists of countable values.
\item If \(X\) is a discrete r.v., the function \(P(X=x)\) is called the probability mass function (PMF) of \(X\).
\item The PMF of any discrete r.v. \(X\) must adhere to these two criteria:
    \begin{itemize}
    \item Nonnegativity: \(P(X=x) > 0\) if \(x = x_i\) for some \(i\), and \(P(X=x) = 0\) otherwise.
    \item Summation to 1: \(\sum_{i=1}^{\infty} P(X=x) = 1\).
    \end{itemize}
\end{enumerate}

\textit{Definition 3 (Continuous Random Variable and Probability Density Function).}

\begin{enumerate}
\item A r.v. \(X\) is defined as continuous if there is a function \(p(.)\) such that for every real number \(x\), the cumulative distribution function (CDF) is given by \(F_X(x) = P(X \leqslant x) = \int_{-\infty}^{x} p(x) \, dx\).
\item For a continuous r.v. \(X\), the function \(p(.)\) in \(F_X(x) = \int p(x) \, dx\) is known as the probability density function (PDF).
\item The PDF of any continuous r.v. \(X\) must meet the following two criteria:
    \begin{itemize}
   \item Nonnegativity: \(p(x) \geq 0\).
   \item Integration to 1: \(\int_{-\infty}^{\infty} p(x) \, dx = 1\).
    \end{itemize}
\end{enumerate}

The term "probability distribution" will be applied to both discrete probability mass functions and continuous probability density functions in our discussions. The specific use of the term will be inferred based on the context.

\textit{Definition 4 (Exponential Family of Probability Distributions).}

\begin{enumerate}
\item The exponential family of probability distributions is a set of PDFs or PMF's characterized by the following form:

\begin{equation}
p(\boldsymbol{x} \mid \boldsymbol{\eta})=h(\boldsymbol{x}) \exp \left\{T(\boldsymbol{x})^T \boldsymbol{\eta}-A(\boldsymbol{\eta})\right\}
\end{equation}

Here, $\boldsymbol{x}$ represents a specific value of a r.v. $X, T(\boldsymbol{x})$ is the sufficient statistic, and $\boldsymbol{\eta}$ is the natural parameter. The function $A(\boldsymbol{\eta})$ is known as the log-partition function and $h(\boldsymbol{x})$ is the base measure.

\item A fundamental property of PDFs or PMFs in the exponential family is expressed by:
\begin{equation}
\mathbb{E}[T(\boldsymbol{x})]=\nabla A(\boldsymbol{\eta})
\end{equation}
Here, $\nabla A(\boldsymbol{\eta})$ denotes the gradient of $A(\boldsymbol{\eta})$. This property highlights the relationship between the expected value of the sufficient statistic and the gradient of the log-partition function.
\end{enumerate}

The exponential family includes many well-known distributions such as the normal, exponential, Poisson, and gamma distributions, among others \citep{wainwright2008graphical}. One of the key advantages of distributions in the exponential family is their mathematical tractability, which simplifies parameter estimation, hypothesis testing, and model interpretation.

\subsection{Graphical Models}

\begin{figure}[h!]
    \centering
    
    \begin{subfigure}{0.83\columnwidth}
     \centering
     \includegraphics[width=\columnwidth]{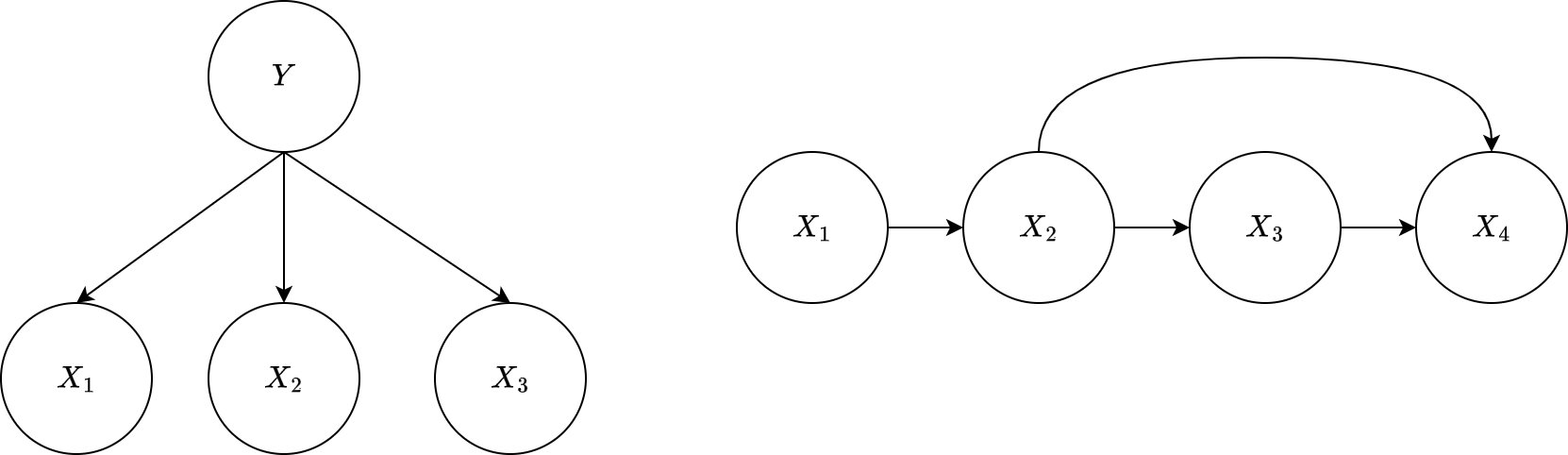}
     \caption{Directed graphical models}
     \label{fig:pgm1}
    \end{subfigure} 
  
  \vspace{5mm}

  \begin{subfigure}{0.33\columnwidth}
    \centering
    \includegraphics[width=\columnwidth]{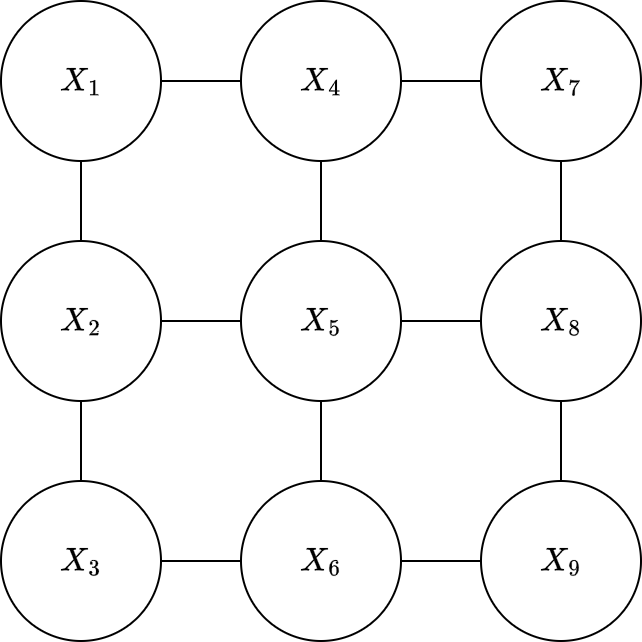}
    \caption{Undirected graphical models}
    \label{fig:pgm2}
  \end{subfigure}
  
  \caption{Examples of different probabilistic graphical models.}
  \label{fig:pgm}
\end{figure}

Probabilistic Graphical Models (PGMs) \citep{koller2009probabilistic} serve as a structured framework for representing and reasoning about complex probabilistic relationships among multiple variables. One of the key advantages of PGMs lies in their ability to handle uncertainty and variability efficiently. As the number of variables grows, managing probability distributions becomes increasingly complex. 
In many real-world scenarios, understanding these relationships and making informed decisions based on uncertain data is essential.

At their core, PGMs leverage graphs to visually depict probabilistic relationships between variables. These graphs consist of nodes, which represent random variables, and edges, which denote probabilistic dependencies between variables. Random variables can be classified as either observed, with their values determined by the problem, or latent, with their values remaining unknown.

There are two main types of PGMs: directed graphical models, also known as Bayesian networks, and undirected graphical models, also called Markov Random Fields. Bayesian networks utilize directed edges to represent influential relationships between variables, while Markov Random Fields capture the concept of local interactions among variables through undirected edges.

\subsubsection{Representing domain knowledge through graphical structures}

One notable aspect of PGMs is the inherent structure of graph representations, which directly implies a factorization of the joint distribution over random variables. In a graphical model, every graph structure entails a specific factorization of the joint distribution. For instance, in a directed graphical model, such as the one shown in Figure \ref{fig:pgm1} (right), the joint distribution factors according to conditional probabilities associated with each node:
\begin{equation}
P\left(X_{1}, \ldots, X_{4}\right)=P\left(X_{1}\right) P\left(X_{2} \mid X_{1} \right) P\left(X_{3} \mid X_{2}\right) P\left(X_{4}  \mid X_{2}, X_{3} \right)
\end{equation}
Similarly, considering the model depicted in Figure \ref{fig:pgm2} (left), the joint distribution can be expressed as a product of clique potentials, where each clique represents a set of variables that are directly connected:
\begin{equation}
P\left(X_{1}, \ldots, X_{9}\right)=\frac{1}{Z} \prod_{c_{i} \in \mathcal{C}} \phi_{i}\left(x_{c}\right)
\end{equation}
where $\mathcal{C}$ represents all cliques within the graph , specifically every pair of connected nodes, $\phi_i(x_c)$ refers to the clique potentials, which are scalar values given to each possible combination of variables within the clique $x_c$, and $Z$ is the normalization constant.

This factorization has practical implications, as we need only track parameters associated with each conditional probability or clique potential. By exploiting this structured representation, significant parameter savings can be achieved compared to a full join distribution. Additionally, the development of a graphical model typically requires integration with domain experts. Such integration encourages careful deliberation of the choice and interconnections of random variables in the model. For example, the grid structure observed in Figure \ref{fig:pgm2} suggests spatial correlations among variables, akin to those found in image pixels. Furthermore, graphical models enable the utilization of structural properties of the underlying data distribution to simplify the computation of probabilistic queries. One notable advantage is the simplification of independence properties, facilitated by the graph's topology.

\subsubsection{Independence properties}

By leveraging insights from graph theory and probability theory, graphical models offer a structured approach to probabilistic reasoning. An essential aspect of understanding these relationships lies in studying independence properties among the distributions over random variables within the graphical model:

\begin{itemize}
    \item \textit{Marginal Independence}. Marginal independence occurs when two random variables are independent of each other, irrespective of the values of other variables. In graphical models, this independence is evident when there is no direct connection (edge) between the variables. For instance, in a directed graphical model, if there is no directed path between two variables, they are considered marginally independent. Mathematically, marginal independence implies that the joint distribution factorizes into the product of marginal distributions for each variable.

    \item \textit{Conditional Independence}. Conditional independence arises when two random variables become independent given the values of a third set of variables. In graphical models, conditional independence statements reveal when observing certain variables render others independent. This concept is crucial for understanding the influence and interaction between variables. Conditional independence relationships are often inferred from the graph's structure, where paths between variables indicate possible dependencies or influences.
\end{itemize}

In directed graphical models, a unique type of conditioning exists that causes variables, which are marginally independent, to become dependent. This concept of "explaining away" refers to a phenomenon where the evidence observed for one variable increases or decreases the probability of another variable being responsible for the same observation, depending on their shared dependencies.
Consider a simple scenario represented by a DGM where variables $A$ and $B$ capture potential causes of chest pain (variable $C$): heart attack and indigestion, respectively, as shown in the graph $A \rightarrow C \leftarrow B$.
As the doctors gather more information about the patient's symptoms and medical history, they can start to narrow down the potential causes. For example, if the patient also shows symptoms like sweating and shortness of breath, which are commonly associated with heart attacks, this evidence increases the likelihood of a heart attack being the cause of the chest pain. Consequently, the probability of indigestion decreases.

The Markov blanket of a random variable refers to a set of variables in a graphical model that, when conditioned on, renders the random variable independent of all other variables in the graph. Specifically, for any random variables \( X \) and \( Y \) in the graphical model \( G \), the Markov blanket (MB) of \( X \) is defined as the minimal set of variables such that conditioning on this set renders the conditional independence of \( X \) and \( Y \):
\begin{equation}
   P(X \mid \mathrm{MB}(X), Y) = P(X \mid \mathrm{MB}(X)) 
\end{equation}
In undirected graphical models, the Markov blanket of a random variable consists of all its neighboring variables. In directed graphical models, the Markov blanket includes a node's parents, its children, and its children's co-parents.

\section{Latent Variable Models}

In the realm of machine learning, one of the fundamental challenges is to accurately model and understand the underlying probability distributions $p(\boldsymbol{x})$ governing high-dimensional data. High-dimensional data, such as images, text documents, and sensor readings, often exhibit complex structures and dependencies that are not easily captured by simple parametric models. Modeling these complex probability distributions is crucial for various tasks, including generative modeling, anomaly detection, and density estimation. Generative models aim to learn the underlying data distribution and generate new samples that resemble the original data. Anomaly detection algorithms rely on accurate probability estimates to identify deviations from normal behavior. Density estimation techniques seek to estimate the probability density function of the data, enabling various downstream tasks such as sampling and likelihood evaluation.

Introducing an unobserved latent variable $\boldsymbol{z}$  with lower dimensionality than observed vectors and defining a conditional distribution $p(\boldsymbol{x} \mid \boldsymbol{z})$ for the data is a powerful approach in probabilistic modeling, particularly in scenarios involving high-dimensional data. This framework allows us to capture complex correlations in the observed variable $\boldsymbol{x}$ by leveraging the latent variable $\boldsymbol{z}$, which serves as a compact representation of underlying factors influencing the data. In this framework, the latent variable encodes meaningful information about the structure and content of the observed data. For example, in the context of modeling medical data, $\boldsymbol{z}$ could encapsulate latent representations of various attributes such as disease subtypes or phenotypes, biomarker signatures, and other relevant features. 
To formalize this probabilistic model, we introduce a prior distribution $p(\boldsymbol{z})$ over the latent variables, representing our beliefs about the likely configurations of $\boldsymbol{z}$. This prior distribution encodes any prior knowledge or assumptions about the latent space. We then compute the joint distribution over observed and latent variables, denoted as $p(\boldsymbol{x}, \boldsymbol{z})$, which describes the probability of observing a particular data point $\boldsymbol{x}$ along with its corresponding latent representation $\boldsymbol{z}$:

\begin{equation}
p(\boldsymbol{x}, \boldsymbol{z})=p(\boldsymbol{x} \mid \boldsymbol{z}) p(\boldsymbol{z})
\end{equation}

Introducing a latent variable in the model enables us to express the complex marginal distribution $p(\boldsymbol{x})$ as a more tractable joint distribution, which consists of the conditional distribution $p(\boldsymbol{x} \mid \boldsymbol{z})$ and the prior distribution $p(\boldsymbol{z})$. Typically, simpler distributions such as exponential family distributions are used to define the conditional distribution $p(\boldsymbol{x} \mid \boldsymbol{z})$ and the prior distribution $p(\boldsymbol{z})$. Exponential family distributions have desirable properties, including tractable normalization constants, which make them computationally efficient for modeling purposes. Once we have the joint distribution $p(\boldsymbol{x}, \boldsymbol{z})$, we can obtain the desired data distribution $p(\boldsymbol{x})$ by marginalizing over the latent variables:

\begin{equation}
p(\boldsymbol{x})=\int p(\boldsymbol{x}, \boldsymbol{z}) \mathrm{d} \boldsymbol{z}=\int p(\boldsymbol{x} \mid \boldsymbol{z}) p(\boldsymbol{z}) \mathrm{d} \boldsymbol{z}
\end{equation}

By applying Bayes' theorem, we can calculate the posterior distribution $p(\boldsymbol{z} \mid \boldsymbol{x})$ as

\begin{equation}
p(\boldsymbol{z} \mid \boldsymbol{x})=\frac{p(\boldsymbol{x} \mid \boldsymbol{z}) p(\boldsymbol{z})}{p(\boldsymbol{x})}
\end{equation}

which allows inference of the latent variable given the observation.

Latent variable models offer a framework to describe the generative process behind the observed data. This process can be interpreted as follows:

\begin{enumerate}
    \item Sampling Latent Variables: To generate a new data point, we first sample a latent variable $\boldsymbol{z}^{(s)}$ from the prior distribution $p(\boldsymbol{z})$. This latent variable captures unobserved factors or features that influence the generation of the data.
    \item Generating Observations: Once we have sampled $\boldsymbol{z}^{(s)}$, we use it to sample a new observation $\boldsymbol{x}^{(s)}$ from the conditional distribution $p(\boldsymbol{x} \mid \boldsymbol{z}^{(s)})$. This conditional distribution captures the relationship between the latent variables and the observed data, allowing us to generate realistic data points.
\end{enumerate}

Latent variable models (LVMs) are particularly effective when the data lie in a manifold, a lower-dimensional structure embedded within the higher-dimensional data space. By capturing the essential characteristics of the data manifold, LVMs can effectively model the underlying data distribution while reducing the dimensionality of the representation. LVMs serve not only as black-box density models but also as interpretable frameworks for incorporating prior knowledge about the generative process underlying the data. Probabilistic graphical models, such as Bayesian networks or Markov random fields, provide a principled way to encode dependencies among variables and incorporate domain knowledge into the joint distribution $p(\boldsymbol{x}, \boldsymbol{z})$.

This dissertation is centered on non-linear LVMs, specifically those utilizing deep neural networks, termed Deep Latent Variable Models (DLVMs). DLVMs are adept at modeling complex, high-dimensional data distributions, yet they necessitate approximate inference due to the intractability of the integral in Equation (2.8), which lacks an analytic solution. Subsequent chapters will explore variational auto-encoders (VAEs) and generative adversarial networks (GANs), which combine principles from deep learning and latent variable models to create highly flexible distributions using deep neural networks.

\subsection{Posterior Inference}

In latent variable models, the posterior distribution updates our understanding of the latent variables based on the data observed. It is essential for probabilistic reasoning, facilitating prediction, inference, and the learning of model parameters. To approximate the complex posterior distribution, two main types of methods are utilized, each balancing accuracy with computational efficiency:

\begin{enumerate}
\item \textit{Sampling}. Sampling techniques, including Markov Chain Monte Carlo (MCMC) approaches, offer an approximation of the posterior distribution through sample generation. These techniques produce samples from the posterior distribution, enabling the estimation of expectations and the execution of inference via Monte Carlo integration. A notable advantage of sampling techniques is their ability to provide precise outcomes with unlimited computational resources. Nevertheless, they often require significant computational effort and may not efficiently handle large datasets. Moreover, assessing convergence and verifying the quality of the samples can pose difficulties.
\item \textit{Deterministic Approximation}. Deterministic approximation methods approximate the posterior distribution analytically by employing parametric distribution families or specific factorizations. Techniques such as variational inference and expectation propagation are examples. These approaches are scalable and efficient, thus appropriate for handling large datasets. Nonetheless, they cannot ensure precise outcomes, even with unlimited computational resources, because of the fundamental approximations involved. Despite these constraints, deterministic approximation methods remain popular due to their computational manageability and ability to scale.
\end{enumerate}

\subsubsection{Variational Inference}

Variational inference (VI) \citep{jordan1999introduction} leverages the calculus of variations to approximate the posterior distribution $p(\boldsymbol{z} \mid \boldsymbol{x})$ by finding an approximate distribution $q(\boldsymbol{z})$ that minimizes the Kullback-Leibler (KL) divergence between the variational distribution and the true posterior. The KL divergence between $q(\boldsymbol{z})$ and $p(\boldsymbol{z} \mid \boldsymbol{x})$ is defined as:

\begin{equation}
K L[q(\boldsymbol{z}) \| p(\boldsymbol{z} \mid \boldsymbol{x})]=-\mathbb{E}_{q(\boldsymbol{z})}\left[\log \frac{p(\boldsymbol{z} \mid \boldsymbol{x})}{q(\boldsymbol{z})}\right]
\end{equation}

The goal of variational inference is to find a good approximation $q(\boldsymbol{z})$ that minimizes this KL divergence. However, the intractability of the posterior $p(\boldsymbol{z} \mid \boldsymbol{x})$ makes direct optimization challenging. To address this, we introduce the evidence lower bound (ELBO), denoted as $\mathcal{F}(q)$, which is defined as:

\begin{equation}
-\mathbb{E}_{q(\boldsymbol{z})}\left[\log \frac{p(\boldsymbol{x}, \boldsymbol{z})}{q(\boldsymbol{z})}\right]
\end{equation}

The ELBO serves as a lower bound on the marginal likelihood $\log p(\boldsymbol{x})$. By maximizing the ELBO with respect to $q(\boldsymbol{z})$, we indirectly minimize the KL divergence, as the KL divergence is equal to $\log p(\boldsymbol{x}) - \mathcal{F}(q)$.

In practice, the variational distribution $q(\boldsymbol{z})$ is often constrained to a parametric family (e.g., Gaussian distribution) to make the optimization tractable. The parameters of this distribution are then optimized to maximize the ELBO. This trade-off between flexibility and tractability ensures that the variational approximation $q(\boldsymbol{z})$ is both expressive enough to capture the posterior distribution and computationally feasible to work with.

In essence, VI offers a principled approach to approximate complex posterior distributions, facilitating efficient and scalable inference within probabilistic models. By framing the inference problem as an optimization task, VI inference reduces the complexity of inference to a simpler optimization problem.

\subsubsection{Expectation Propagation}

Unlike variational inference, which minimizes the KL divergence from a chosen approximation to the true posterior, expectation propagation (EP) \citep{minka2013expectation} minimizes the reverse KL divergence, which is defined as:

\begin{equation}
K L[p(\boldsymbol{z} \mid \boldsymbol{x}) \| q(\boldsymbol{z})] = \mathbb{E}_{p(\boldsymbol{z} \mid \boldsymbol{x})}\left[\log \frac{p(\boldsymbol{z} \mid \boldsymbol{x})}{q(\boldsymbol{z})}\right]
\end{equation}

 It is important to note that EP does not necessarily minimize the KL divergence but can be freely implemented with any divergence measure. When $q(\boldsymbol{z})$ belongs to the exponential family, which is commonly the case, the minimization of the reverse KL divergence simplifies to the alignment of natural parameters. In EP, the goal is to approximate the posterior distribution \(p(\boldsymbol{z})\) with a set of factors \(Q(\boldsymbol{z})\) such that each factor \(q_i(\boldsymbol{z})\) approximates the corresponding factor \(f_i(\boldsymbol{z})\) in the exact posterior factorization. This approximation is achieved by minimizing a sequence of local Kullback-Leibler (KL) divergences between the exact factors and the EP approximating factors. 

The EP approximation for \(p(\boldsymbol{z})\) is given by:

\begin{equation}    
Q(\boldsymbol{z}) = \prod_{i=0}^n q_i(\boldsymbol{z})
\end{equation}

where \(q_i(\boldsymbol{z})\) represents the EP approximating factors and \(Q(\boldsymbol{z})\) is the global approximation.

To find the parameters for determining the approximate factors \(q_i(\boldsymbol{z})\), EP minimizes a sequence of local KL divergences:

\begin{equation}
\begin{aligned}
q_0(\boldsymbol{z}) &= \underset{q_0(\boldsymbol{z}) \in \mathcal{Q}}{\arg \min } K L\left(\tilde{f}_0(\boldsymbol{z}) \| q_0(\boldsymbol{z}) Q^{\backslash 0}(\boldsymbol{z})\right) \\
q_1(\boldsymbol{z}) &= \underset{q_1(\boldsymbol{z}) \in \mathcal{Q}}{\arg \min } K L\left(\tilde{f}_1(\boldsymbol{z}) \| q_1(\boldsymbol{z}) Q^{\backslash 1}(\boldsymbol{z})\right) \\
& \vdots \\
q_n(\boldsymbol{z}) &= \underset{q_n(\boldsymbol{z}) \in \mathcal{Q}}{\arg \min } K L\left(\tilde{f}_n(\boldsymbol{z}) \| q_n(\boldsymbol{z}) Q^{\backslash n}(\boldsymbol{z})\right)
\end{aligned}
\end{equation}

where \(\mathcal{Q}\) denotes the space of possible approximate factors, \(\tilde{f}_i(\boldsymbol{z}) = f_i(\boldsymbol{z}) Q^{\backslash i}(\boldsymbol{z})\) is the tilted distribution, and \(Q^{\backslash i}(\boldsymbol{z})\) is the cavity distribution obtained by removing the current KL minimizer \(q_i(\boldsymbol{z})\) from \(Q(\boldsymbol{z})\). 

Each KL divergence minimization problem in the above equations is solved by exclusively optimizing \(q_i(\boldsymbol{z})\) instead of the EP global approximation. Therefore, expectation propagation is considered a local approximation algorithm as each KL divergence is minimized locally with respect to a selected EP approximating factor.

In EP, convergence is not guaranteed because the local KL divergence minimization does not necessarily ensure that the KL divergence is minimized from the exact posterior distribution to the EP global posterior approximation. Despite the absence of formal convergence guarantees, EP remains a widely used and effective approximate inference algorithm in probabilistic modeling due to its flexibility and applicability to a variety of models.

\subsection{Parameter Learning}

In parameter learning for latent variable models, we aim to estimate the optimal parameters $\theta^{\star}$ of the model given a training set comprising $N$ data points $\{\boldsymbol{x}^{i}\}_{i=1}^{N}$. The likelihood $p_{\theta}(\boldsymbol{x} \mid \boldsymbol{z})$ and the prior $p_{\theta}(\boldsymbol{z})$ are assumed to belong to families of distributions parameterized by unknown parameters $\theta$.

The optimal parameters $\theta^{\star}$ can be learned using Maximum Likelihood Estimation (MLE), which involves maximizing the log-likelihood function $\mathcal{L}(\theta)$:

\begin{equation}
\mathcal{L}(\theta) = \sum_{i=1}^{N} \log p_{\theta}(\boldsymbol{x}^{i})
\end{equation}

Given that the latent variable $\boldsymbol{z}^{i}$ is different for each data point $\boldsymbol{x}^{i}$, but the parameters $\theta$ are shared across all data points, we can express the log-likelihood as the sum of individual terms $\mathcal{L}_{i}(\theta)$:

\begin{equation}
\mathcal{L}(\theta) = \sum_{i=1}^{N} \underbrace{\log \int p_{\theta}(\boldsymbol{x}^{i}, \boldsymbol{z}^{i}) \mathrm{d} \boldsymbol{z}^{i}}_{\mathcal{L}_{i}(\theta)}
\end{equation}

In practice, the marginal density of the observations $p_{\theta}(\boldsymbol{x})$ is often intractable and needs to be approximated. One approach is to use the evidence lower bound (ELBO), denoted as $\mathcal{F}_{i}(\theta, q)$, which provides a lower bound to $\log p_{\theta}(\boldsymbol{x})$ for any distribution $q(\boldsymbol{z})$ over the latent variables:

\begin{equation}
\mathcal{L}_{i}(\theta) \geq \mathcal{F}_{i}(\theta, q) = \mathbb{E}_{q(\boldsymbol{z})}\left[\log \frac{p_{\theta}(\boldsymbol{x}, \boldsymbol{z})}{q(\boldsymbol{z})}\right]
\end{equation}

By maximizing the total ELBO $\mathcal{F}(\theta, q) = \sum_{i=1}^{N} \mathcal{F}_{i}(\theta, q)$ with respect to $\theta$ and $q(\boldsymbol{z})$, we can learn the parameters of the model. The variational distribution $q(\boldsymbol{z})$ can be interpreted as an approximation to the posterior distribution $p_{\theta}(\boldsymbol{z} \mid \boldsymbol{x})$, and the ELBO coincides with the log-likelihood only when $q(\boldsymbol{z})$ equals the true posterior distribution.

In practice, the variational distribution $q(\boldsymbol{z})$ is often constrained to a particular parametric family to make the optimization tractable, and the parameters of this distribution are optimized along with the parameters $\theta$ of the model. Therefore, by maximizing the ELBO, we indirectly maximize the log-likelihood, enabling parameter learning in latent variable models. 

\subsubsection{Expectation Maximization}

The Expectation Maximization (EM) algorithm \citep{dempster1977maximum} provides a systematic approach for maximizing the likelihood function in models with latent variables. It iteratively alternates between two steps: the E-step, where the posterior distribution over latent variables is estimated, and the M-step, where the model parameters are updated based on the estimated posteriors.

Starting with initial parameters $\theta_{0}$, the EM algorithm iterates until convergence as follows:

\begin{enumerate}
\item E-step (Expectation): Given the current parameters $\theta_{k}$, estimate the posterior distribution over latent variables, denoted as $q_{k+1}(\boldsymbol{z})$, by maximizing the ELBO $\mathcal{F}_{i}(\theta_{k}, q)$ with respect to $q(\boldsymbol{z})$. In many cases, this step involves solving a posterior inference problem, aiming to find an approximation to the true posterior distribution $p_{\theta_{k}}(\boldsymbol{z} \mid \boldsymbol{x})$. If the posterior is intractable, approximate inference methods can be employed.
\item M-step (Maximization): Fixing the estimated distribution over latent variables $q_{k+1}(\boldsymbol{z})$, update the parameters $\theta_{k+1}$ by maximizing the ELBO $\mathcal{F}_{i}(\theta, q_{k+1})$ with respect to $\theta$. This step involves optimizing the model parameters using techniques such as gradient ascent.
\end{enumerate}

For simpler classes of models where exact inference is possible, each EM iteration guarantees not to decrease the marginal likelihood after each combined step. Specifically, after the E-step, where $\theta_{k}$ is held fixed, the ELBO equals the log-likelihood. Subsequently, maximizing the ELBO in the M-step does not decrease the log-likelihood.

In summary, the EM algorithm offers a systematic and effective approach for optimizing model parameters when latent variables are involved. By iteratively refining the estimates of the latent variables and updating the model parameters, EM enables efficient learning in latent variable models even when exact inference is not feasible.
\chapter{PPG-to-ECG Signal Translation For Continuous Atrial Fibrillation Detection via Attention-based Deep State-Space Modeling}

\section{Introduction}

\label{sec:intro}

The measurement of the electrical activity generated by an individual's heart, known as an electrocardiogram (ECG), typically requires the placement of several electrodes on the body. ECG is considered the preferred method for monitoring vital signs and for the diagnosis, management, and prevention of cardiovascular diseases (CVDs), which are a leading cause of death globally, accounting for approximately 32\% of all deaths in 2017 according to Global Burden of Disease reports \citep{ref:Allen}. It has also been demonstrated that sudden cardiac arrests are becoming more prevalent in young individuals, including athletes ~\citep{ref:sudden}. Regular ECG monitoring has been found to be beneficial for the early identification of CVDs ~\citep{rosiek2016risk}. Among heart diseases, atrial fibrillation (AFib) is adults' most common rhythm disorder. Identifying AFib at an early stage is crucial for the primary and secondary prevention of cardioembolic stroke, as it is the leading risk factor for this type of stroke \citep{olier2021machine}.
Advancements in electronics, wearable technologies, and machine learning have made it possible to record ECGs more easily and accurately, and to analyze large amounts of data more efficiently. Despite these developments, there are still challenges associated with continuously collecting high-quality ECG data over an extended period, particularly in everyday life situations. 
The 12-lead ECG, considered the clinical gold standard, and simpler versions, such as the Holter ECG, can be inconvenient and bulky due to the need to place multiple electrodes on the body, which can cause discomfort. Additionally, the signals may degrade over time as the impedance between the skin and electrodes changes. Consumer-grade products such as smartwatches have developed solutions to address these issues. However, these products require users to place their fingers on the watch to form a closed circuit, making continuous monitoring impossible.

\begin{figure*}[h!]
\centering
\includegraphics[scale=.4]{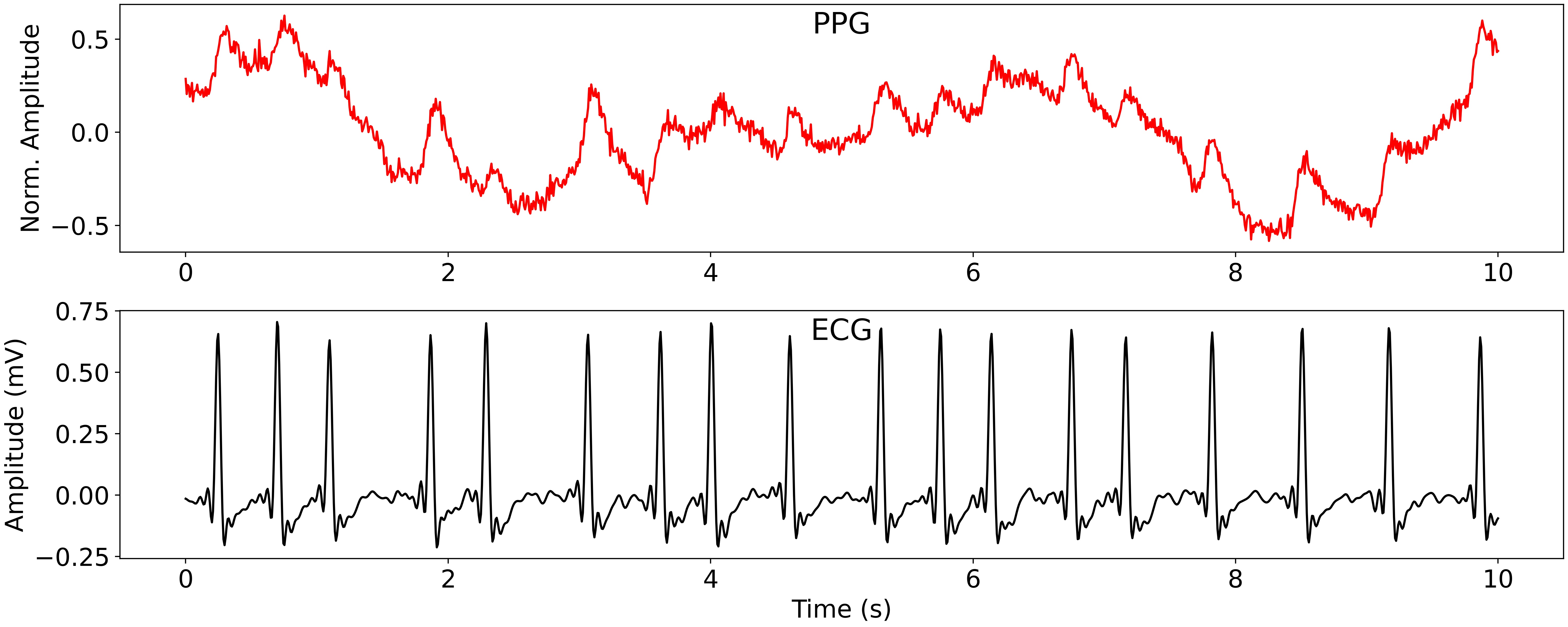}
\caption{A PPG-ECG waveform pair. PPG signals can often become contaminated by noise.}
\label{fig:preprocessed}
\end{figure*}

One potential solution to these issues is to use a mathematical method to derive ECG data from an alternative, highly correlated, non-invasive signal, such as the photoplethysmogram (PPG), which can be easily acquired using various wearable devices, including smartwatches. 
PPG is more convenient, cost-effective, and user-friendly. PPG has been increasingly adopted in consumer-grade devices. This technique involves the use of a light source, usually an LED, and a photodetector to measure the changes in light absorption or reflection as blood flows through the tissue. 
ECG and PPG signals are inherently correlated as both are influenced by the same underlying cardiac activity, namely the depolarization and repolarization of the heart. These contractions lead to changes in peripheral blood volume, which are measured by PPG.  
Figure \ref{fig:preprocessed} shows the relationship between ECG and PPG waveforms.
Although there are established standards for interpreting ECG for clinical diagnosis, the use of PPG is still mostly limited to measuring heart rate and oxygen saturation ~\citep{reisner2008utility}. By translating PPG to ECG signals, clinical diagnoses of cardiac diseases and anomalies could be made in real-time.

Few research works attempted to synthesize ECG from PPG signals. In \citep{banerjee2014photoecg}, a machine learning-based approach was proposed to estimate the ECG parameters, including the RR, PR, QRS, and QT intervals, using features from the time and frequency domain extracted from a fingertip PPG signal. Additionally, \citet{zhu2019ecg, tian2020cross} proposed models to reconstruct the entire ECG signal from PPG in the frequency domain. However, the performance of these approaches relied on cumbersome algorithms for feature crafting. With recent advances in deep learning, \citet{vo2021p2e, sarkar2021cardiogan, chiu2020reconstructing} leveraged the expressiveness and structural flexibility of neural networks to build end-to-end PPG-to-ECG algorithms. 
However, the models suffer from data-hungry problems as they do not explicitly model the underlying sequential structures of the data. 
In addition, complex deep learning models cannot run efficiently on resource-constrained devices (e.g., wearables) due to their high computational intensity, which poses a critical challenge for real-world deployment \citep{lee2020stint}.
Furthermore, deterministic models face difficulties in effectively generalizing to noisy data.

To address these challenges, we propose a deep probabilistic model to accurately estimate ECG waveforms from raw PPG. The contributions of this work are three-fold:
\begin{itemize}
    \item We present a deep generative model incorporating prior knowledge about the data structures that enable learning on small datasets. Specifically, we develop a deep latent state-space model augmented by an attention mechanism.
    \item The probabilistic nature of the model enhances its robustness to noise. We demonstrate this by evaluating the model on data corrupted with Gaussian and baseline wandering noise, replicating real-life situations.
    \item Our method is effective not only in healthy subjects but also in subjects with AFib. It is orthogonal and complementary to existing AFib detection methods \citep{hong2020opportunities} by simply providing the translated ECG to any pre-trained models. This would enhance the performance of existing models by enabling uninterrupted monitoring, thereby facilitating the early detection of cardiovascular disease.
\end{itemize}

\section{Methodology}

\subsection{Probabilistic Modeling of ECG from PPG signals}
\sloppy We are given a dataset $\mathcal{D}:=\left\{\left(\boldsymbol{x}^{1}, \boldsymbol{y}^{1}\right), \ldots,\left( \boldsymbol{x}^{N}, \boldsymbol{y}^{N}\right)\right\}$ with the $i$-th observation $\boldsymbol{y}^{i} \in$ $\mathbb{R}^{n_y}$, i.e., ECG signals of $n_y$ time samples, depending on $\boldsymbol{x}^{i} \in$ $\mathbb{R}^{n_x}$, i.e., PPG signals of $n_x$ time samples. Throughout the paper, superscript $i$ is omitted when we refer to only one sequence or when it is clear from the context.

We aim to learn a generative process with a latent-variable model 
comprising of a parametric non-linear Gaussian prior over latents $p_{\theta_z}(\boldsymbol{z} \mid \boldsymbol{x})$ and likelihood $p_{\theta_y}(\boldsymbol{y} \mid \boldsymbol{z}, \boldsymbol{x})$. The learning process minimizes a divergence between the true data-generating distribution and the model w.r.t $\theta$:
\begin{equation}
\begin{aligned}
& \underset{\theta}{\arg \min } K L\left(p_{\mathcal{D}}(\boldsymbol{y} \mid \boldsymbol{x}) \| p_\theta(\boldsymbol{y} \mid \boldsymbol{x})\right) \\ 
&=\underset{\theta}{\arg \max } \mathbb{E}_{p_{\mathcal{D}}(\boldsymbol{y} \mid \boldsymbol{x})}\left[\log p_\theta(\boldsymbol{y} \mid \boldsymbol{x})\right]
\end{aligned}\
\end{equation}
where $p_\theta\left(\boldsymbol{y} \mid \boldsymbol{x}\right)=\int p_{\theta_y}\left(\boldsymbol{y} \mid \boldsymbol{z}, \boldsymbol{x}\right) p_{\theta_z}\left(\boldsymbol{z} \mid \boldsymbol{x}\right) d   \boldsymbol{z}$ is the conditional likelihood/evidence of data point $\boldsymbol{y}$ given condition $\boldsymbol{x}$, approximated by averaging over the latent $\boldsymbol{z}$.

Nevertheless, estimating $p_\theta(\boldsymbol{y} \mid \boldsymbol{x})$ is typically intractable. This issue can be mitigated by introducing a parametric inference model $q_\phi(\boldsymbol{z} \mid \boldsymbol{x}, \boldsymbol{y})$ to construct a conditional variational evidence lower bound on the conditional log-likelihood $\log p_\theta(\boldsymbol{y} \mid \boldsymbol{x})$ as follows
\begin{equation} 
\begin{aligned}
& \mathcal{L}(\boldsymbol{x}, \boldsymbol{y}; \theta, \phi) \\ 
& \triangleq \log p_\theta(\boldsymbol{y} \mid \boldsymbol{x})-K L\left(q_\phi(\boldsymbol{z} \mid \boldsymbol{x}, \boldsymbol{y}) \| p_\theta(\boldsymbol{z} \mid \boldsymbol{x}, \boldsymbol{y})\right) \\
& =\mathbb{E}_{q_\phi(\boldsymbol{z} \mid \boldsymbol{x}, \boldsymbol{y})}\left[\log p_{\theta_y}(\boldsymbol{y} \mid \boldsymbol{z}, \boldsymbol{x})\right]-K L\left(q_\phi(\boldsymbol{z} \mid \boldsymbol{x}, \boldsymbol{y}) \| p_{\theta_z}(\boldsymbol{z} \mid \boldsymbol{x})\right)
\end{aligned}
\end{equation}
Taking the likelihood model $p_{\theta_y}(\boldsymbol{y} \mid \boldsymbol{z}, \boldsymbol{x})$ to be a decoder, the latent inference model $q_\phi(\boldsymbol{z} \mid \boldsymbol{x}, \boldsymbol{y})$ to be an encoder, and the prior model $p_{\theta_z}(\boldsymbol{z} \mid \boldsymbol{x})$, a conditional variational autoencoder (CVAE) \citep{kingma2013auto, sohn2015learning} considers this objective from a deep probabilistic autoencoder perspective. Here $\theta$ and $\phi$ are neural network parameters, and learning takes place via stochastic gradient ascent using unbiased estimates of $\nabla_{\theta, \phi} \frac{1}{n} \sum_{i=1}^n \mathcal{L}\left(\boldsymbol{x}^{i}, \boldsymbol{y}^{i}; \theta_z, \theta_y, \phi\right)$.

\subsection{State-Space Modeling of ECG from PPG Signals}
In the previous section, we consider the networks that process the entire time series as a whole, which do not explicitly model the underlying sequential natures of the data. This may lead to resource-inefficient learning.  Here, propose to address the problems by leveraging the \textit{quasi-periodic nature} of the physiological signals.

\subsubsection{ECG Generative (Decoding) Process from PPG}
We consider nonlinear dynamical systems with observations $\boldsymbol{y}_{t} \in \mathbb{R}^{n_{rr}}$, i.e., RR intervals or the time elapsed between two successive R peaks on the ECG, depending on control inputs $\boldsymbol{x}_{t} \in \mathbb{R}^{n_{pp}}$, i.e., PP intervals or the time elapsed between two successive systolic peaks on the PPG. We choose the peaks to segment the signals as they are the most robust features. Corresponding discrete-time sequences of length $T$ are denoted as $\boldsymbol{y}_{1: T}=\left(\boldsymbol{y}_{1}, \boldsymbol{y}_{2}, \ldots, \boldsymbol{y}_{T}\right)$ and $\boldsymbol{x}_{1: T}=\left(\boldsymbol{x}_{1}, \boldsymbol{x}_{2}, \ldots, \boldsymbol{x}_{T}\right)$.

Given an input PPG $\boldsymbol{x}_{1: T}$, we are interested in a probabilistic model $p\left(\boldsymbol{y}_{1: T} \mid \boldsymbol{x}_{1: T}\right)$. Formally, we consider
\begin{equation}
p\left(\boldsymbol{y}_{1: T} \mid \boldsymbol{x}_{1: T}\right)=\int p\left(\boldsymbol{y}_{1: T} \mid \boldsymbol{z}_{1: T}, \boldsymbol{x}_{1: T}\right) p\left(\boldsymbol{z}_{1: T} \mid \boldsymbol{x}_{1: T}\right) \mathrm{d} \boldsymbol{z}_{1: T}
\end{equation}
where $\boldsymbol{z}_{1: T}$ represents the latent sequence associated with the given model. This implies that we are considering a generative model that incorporates a latent dynamical system with an emission model $p\left(\boldsymbol{y}_{1: T} \mid \boldsymbol{z}_{1: T}, \boldsymbol{x}_{1: T}\right)$ and transition model $p\left(\boldsymbol{z}_{1: T} \mid \boldsymbol{x}_{1: T}\right)$.

\begin{figure}[h!]
\centering
\includegraphics[scale=.625]{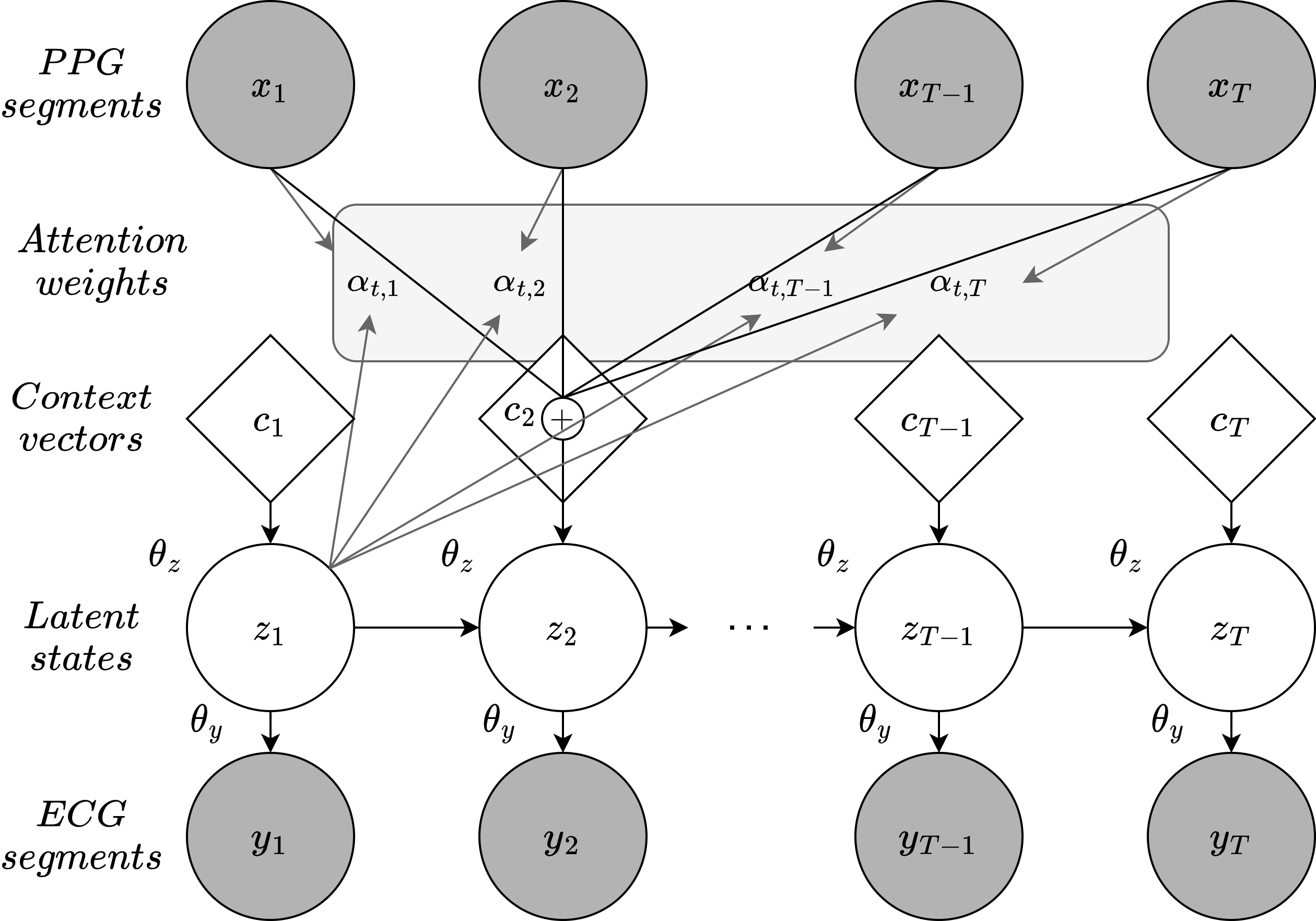}
\caption{The graphical model for ECG translation from PPG. Shaded nodes represent observed variables. Clear nodes represent latent variables. Diamond nodes denote deterministic variables. Variables $\boldsymbol{x}_t, \boldsymbol{y}_t$, and $\boldsymbol{c}_t$ represent PP intervals, RR intervals, and context vectors, respectively. $\alpha_{t,i}$ are attention weights defines how well two intervals $\boldsymbol{x}_i$ and $\boldsymbol{y}_t$ are aligned. The attention mechanism is shown only at time step 2.}
\label{fig:gen}
\end{figure}

To derive state-space models, we make certain assumptions regarding the state transition and emission models, as shown in Figure \ref{fig:gen}:
\begin{align}
p\left(\boldsymbol{z}_{1: T} \mid \boldsymbol{x}_{1: T}\right) & =\prod_{t=0}^{T-1} p\left(\boldsymbol{z}_{t+1} \mid \boldsymbol{z}_{t}, \boldsymbol{x}_{1: T}\right) \\
p\left(\boldsymbol{y}_{1: T} \mid \boldsymbol{z}_{1: T}, \boldsymbol{x}_{1: T}\right) & =\prod_{t=1}^{T} p\left(\boldsymbol{y}_{t} \mid \boldsymbol{z}_{t}\right)
\end{align}
Equations 3.4 and 3.5 make the assumption that the current state $\boldsymbol{z}_{t}$ includes all the relevant information about both the current observation $\boldsymbol{y}_{t}$ and the next state $\boldsymbol{z}_{t+1}$, given the current control input $\boldsymbol{x}_{t}$.


In contrast to the DKF model of \citep{krishnan2015deep, krishnan2017structured}, our model takes into account the entire input signal $\boldsymbol{x}_{1: T}$ for each output $\boldsymbol{y}_{t}$ via an attention mechanism \citep{bahdanau2014neural}.
Note that there are usually misalignments between the PPG and ECG cycles. Therefore, it is difficult to construct optimal and exact sample pairs. This attention mechanism not only helps to add more context to generate ECG segments, but also helps to address the problem of misalignment. 
 
Let us define $\boldsymbol{c}_t$ a sum of features of the input sequence (PP intervals), weighted by the alignment scores:
\begin{align}
\boldsymbol{c}_t & =\sum_{i=1}^T \alpha_{t, i} \boldsymbol{x}_i \\
\alpha_{t, i} 
& =\frac{\exp \left(\boldsymbol{s}\left(\boldsymbol{z}_{t-1}, \boldsymbol{x}_i\right)\right)}{\sum_{i^{\prime}=1}^n \exp \left(\boldsymbol{s}\left(\boldsymbol{z}_{t-1}, \boldsymbol{x}_{i^{\prime}}\right)\right)}
\end{align}
The alignment function $\boldsymbol{s}$ assigns a score $\alpha_{t,i}$ to the pair of input at position $i$ and output at position $t$, ($\boldsymbol{x}_i, \boldsymbol{y}_t$), based on how well they match. The set of $\alpha_{t,i}$ are weights defining how much of each source segment should be considered for each output interval. 

Both state transition (prior) and emission models are non-linear Gaussian transformations parametrized by neural networks $\theta_z$ and $\theta_y$ :
\begin{align}
 p_{\theta_z}(\boldsymbol{z}_{t+1} \mid \boldsymbol{z}_t, \boldsymbol{x}_{1: T}) & = \mathcal{N}(\boldsymbol{z}_{t+1} \mid \boldsymbol{\mu}_{\theta_z}(\boldsymbol{z}_t, \boldsymbol{c}_{t+1}), \boldsymbol{\sigma}^2_{\theta_z}(\boldsymbol{z}_t, \boldsymbol{c}_{t+1})); \\
 p_{\theta_y}(\boldsymbol{y}_t \mid \boldsymbol{z}_t) & =\mathcal{N}(\boldsymbol{y}_t \mid \boldsymbol{\mu}_{\theta_y}(\boldsymbol{z}_t), \boldsymbol{I})
\end{align}

where $\boldsymbol{\mu}$ and $\boldsymbol{\sigma}^2$ are the means and diagonal covariance matrices of the normal distributions $\mathcal{N}$, $\boldsymbol{I}$ is the identity covariance matrix.

\subsubsection{Latent State Inference (Posterior Encoding) Process}

\begin{figure}[h!]
\centering
\includegraphics[scale=.625]{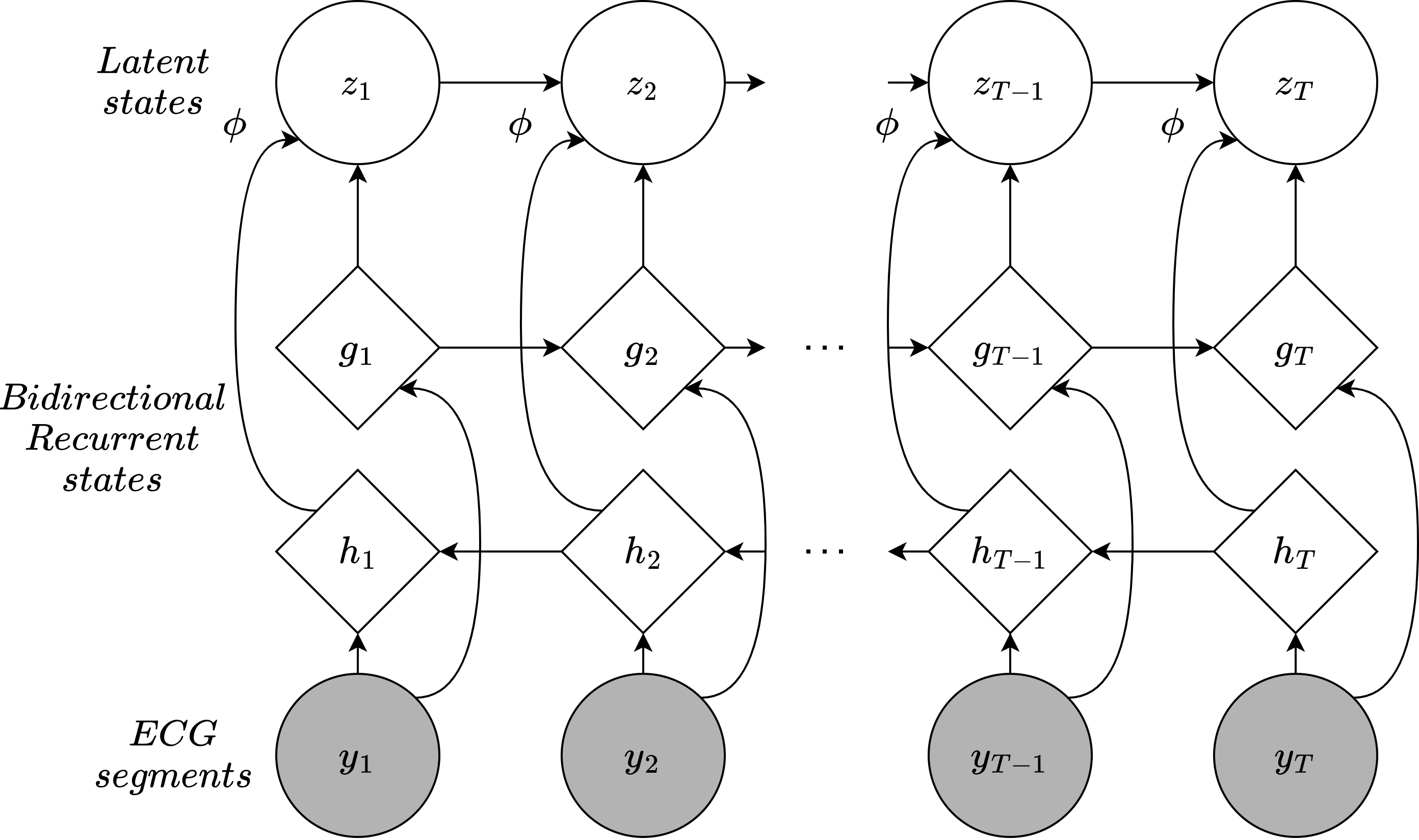}
\caption{The graphical model at latent state inference time. Variables $\boldsymbol{y}_t, \boldsymbol{h}_t, \boldsymbol{g}_t$, and $\boldsymbol{z}_t$ represent respectively RR intervals, backward, forward recurrent states, and latent states.}
\label{fig:infer}
\end{figure}

Unlike a deterministic translation model, the process needs to find meaningful probabilistic embeddings of ECG segments in the latent space. We want to identify the structure of the parametetrized posterior distribution $q_\phi\left(\boldsymbol{z}_{1: T} \mid \boldsymbol{y}_{1: T}\right)$. Notice that we made a design choice to perform inference using only $\boldsymbol{y}_{1: T}$. We chose this with the conditional independence assumption that PPG segments do not provide more information than ECG segments alone. The graphical model in Figure \ref{fig:gen} shows that the $\boldsymbol{z}_{t}$ node blocks all information coming from the past and flowing to $\boldsymbol{z}_{t+1}$ (i.e., $\boldsymbol{z}_{1: t-1}$ and $\left.\boldsymbol{y}_{1: t}\right)$, leading to the following structure as in Figure \ref{fig:infer}:
\begin{equation}
q_\phi\left(\boldsymbol{z}_{1: T} \mid \boldsymbol{y}_{1: T}\right)=q_\phi\left(\boldsymbol{z}_{1} \mid \boldsymbol{y}_{1: T}\right)\prod_{t=1}^{T-1} q_\phi\left(\boldsymbol{z}_{t+1} \mid \boldsymbol{z}_{t}, \boldsymbol{y}_{t+1: T}\right)    
\end{equation}
where
\begin{equation}
 q_{\phi}(\boldsymbol{z}_{t+1} \mid \boldsymbol{z}_t, \boldsymbol{y}_{t+1: T}) = \mathcal{N}(\boldsymbol{z}_{t+1} \mid \boldsymbol{\mu}_{\phi}(\boldsymbol{z}_t, \boldsymbol{y}_{t+1: T}), \boldsymbol{\sigma}^2_{\phi}(\boldsymbol{z}_t, \boldsymbol{y}_{t+1: T}))
\end{equation}

\subsubsection{Training Process}
The objective function becomes a timestep-wise conditional variational lower bound \citep{kingma2013auto, sohn2015learning, krishnan2017structured}: 
\begin{equation}
\begin{aligned}
& \log p_\theta(\boldsymbol{y} \mid \boldsymbol{x}) \geq \mathcal{L}(\boldsymbol{x}, \boldsymbol{y} ;\theta_y, \theta_z, \phi) \triangleq \\
& \sum_{t=1}^T \underset{q_\phi\left(\boldsymbol{z}_t \mid \boldsymbol{y}_{t:T}\right)}{\mathbb{E}}[\overbrace{\log \underbrace{p_{\theta_y}\left(\boldsymbol{y}_t \mid \boldsymbol{z}_t\right)}_{\text {emission model}}}^{\text {reconstruction}}] \\ 
& - \beta\overbrace{K L\left(q_\phi\left(\boldsymbol{z}_1 \mid \boldsymbol{y}_{1:T}\right) \| p_{\theta_z}\left(\boldsymbol{z}_1 \mid \boldsymbol{x}_{1:T}\right)\right)}^{\text {regularization}} \\
& -\beta\sum_{t=1}^{T-1} \underset{q_\phi\left(\boldsymbol{z}_{t} \mid \boldsymbol{y}_{t:T}\right)}{\mathbb{E}}[\overbrace{K L(\underbrace{q_\phi\left(\boldsymbol{z}_{t+1} \mid \boldsymbol{z}_{t}, \boldsymbol{y}_{t:T}\right)}_{\text {posterior inference model}}|| \underbrace{p_{\theta_z}\left(\boldsymbol{z}_{t+1} \mid \boldsymbol{z}_t, \boldsymbol{x}_{1:T}\right)}_{\text {prior transition model}})}^{\text {regularization}}]
\end{aligned}
\end{equation}
where $\beta$ controls the regularization strength. During training, the Kullback–Leibler (KL) losses in the regularization terms "pull" the posterior distributions (which encode EEG segments) and the prior distributions (which embed PPG segments) towards each other. We learn the generative and inference models jointly by maximizing the conditional variational lower bound with respect to their parameters.
\subsection{Neural Network Architectures}
Let us denote $\boldsymbol{W}$, $\boldsymbol{v}$, and $\boldsymbol{b}$ the weight matrices.

\textbf{Score Model}: The alignment score $\alpha$ in Equation 3.7 is parametrized by a feedforward network with a single hidden layer, and this network is jointly trained with other parts of the model. 
The score function $\boldsymbol{s}$ is in the following form:
\begin{equation}
    \boldsymbol{s}\left(\boldsymbol{z}_{t-1}, \boldsymbol{x}_i\right)=\boldsymbol{v}_s^{\top} \tanh \left(\boldsymbol{W}_{s}\left[\boldsymbol{z}_{t-1}; \boldsymbol{W}_{x}\boldsymbol{x}_i\right] + \boldsymbol{b}_s \right)
\end{equation}
\textbf{Prior Transition Model}: We parametrize the transition function in Equation 3.8 from $z_t$ to $z_{t+1}$ using a Gated Transition Function as in \citep{krishnan2017structured}. The model is flexible in choosing a non-linear transition for some dimensions while having linear transitions for others. The function is parametrized as follows:
\begin{equation}
\begin{aligned}
& \boldsymbol{g}_t= \operatorname{sigmoid}( \boldsymbol{W}_{g_3}\operatorname{ReLU} \left(\boldsymbol{W}_{g_2}\operatorname{ReLU} \left(\boldsymbol{W}_{g_1}\left[\boldsymbol{z}_{t}; \boldsymbol{c}_{t+1}\right] + \boldsymbol{b}_{g_1}\right) + \boldsymbol{b}_{g_2}\right) + \boldsymbol{b}_{g_3})\\
&  \boldsymbol{d}_t=  \boldsymbol{W}_{d_3}\operatorname{ReLU}\left(\boldsymbol{W}_{d_2}\operatorname{ReLU} \left(\boldsymbol{W}_{d_1}\left[\boldsymbol{z}_{t}; \boldsymbol{c}_{t+1}\right] + \boldsymbol{b}_{d_1}\right) + \boldsymbol{b}_{d_2}\right) + \boldsymbol{b}_{d_3}\\
& \boldsymbol{\mu}_{\theta_z}(\boldsymbol{z}_t, \boldsymbol{c}_{t+1})=\left(1-\boldsymbol{g}_t\right) \odot\left(\boldsymbol{W}_{\mu_z} \left[\boldsymbol{z}_{t}; \boldsymbol{c}_{t+1}\right]+\boldsymbol{b}_{\mu_z}\right)+\boldsymbol{g}_t \odot \boldsymbol{d}_t \\
& \boldsymbol{\sigma}^2_{\theta_z}(\boldsymbol{z}_t, \boldsymbol{c}_{t+1})=\operatorname{softplus}\left(\boldsymbol{W}_{\sigma_z^2} \operatorname{ReLU}\left(\boldsymbol{d}_t\right)+\boldsymbol{b}_{\sigma_z^2}\right)
\end{aligned}
\end{equation}
where $\mathbb{I}$ denotes the identity function, and $\odot$ denotes element-wise multiplication.

\textbf{Emission Model}: We parameterize the emission function in Equation 3.9 using a two-hidden layer network as:
\begin{equation}
\begin{aligned} 
& \boldsymbol{\mu}_{\theta_y}\left(\boldsymbol{z}_t\right)= \boldsymbol{W}_{e_3} \operatorname{ReLU} \left(\boldsymbol{W}_{e_2}\operatorname{ReLU} \left(\boldsymbol{W}_{e_1}\boldsymbol{z}_t + \boldsymbol{b}_{e_1}\right) + \boldsymbol{b}_{e_2}\right) + \boldsymbol{b}_{e_3}
\end{aligned}
\end{equation}

\textbf{Posterior Inference Model}: We use a Bi-directional Gated Recurrent Unit network \citep{chung2014empirical} (GRU) to process the sequential order of RR intervals backward from $\boldsymbol{y}_T$ to $\boldsymbol{y}_{t+1}$ and forward from $\boldsymbol{y}_{t+1}$ to $\boldsymbol{y}_{T}$. The GRUs are denoted here as $\boldsymbol{h}_{t}=\operatorname{GRU}\left(\boldsymbol{W}_y\boldsymbol{y}_{T}, \ldots, \boldsymbol{W}_y\boldsymbol{y}_{t+1}\right)$ and $\boldsymbol{g}_{t}=\operatorname{GRU}\left(\boldsymbol{W}_y\boldsymbol{y}_{t+1}, \ldots, \boldsymbol{W}_y\boldsymbol{y}_{T}\right)$, respectively. The hidden states of the GRUs parametrize the variational distribution, which are combined with the previous latent states for the inference in Equation 3.11 as follows:
\begin{equation}
\begin{aligned}
\boldsymbol{\tilde{h}}_{t} & =\frac{1}{3}\left(\tanh \left(\boldsymbol{W}_{h} \boldsymbol{z}_{t}+\boldsymbol{b}_{h}\right)+\boldsymbol{h}_t +  \boldsymbol{g}_t\right) \\
\boldsymbol{\mu}_{\phi}(\boldsymbol{z}_t, \boldsymbol{y}_{t+1: T}) & =\boldsymbol{W}_\mu \boldsymbol{\tilde{h}}_{t}+\boldsymbol{b}_\mu \\
\boldsymbol{\sigma}^2_{\phi}(\boldsymbol{z}_t, \boldsymbol{y}_{t+1: T}) & =\operatorname{softplus}\left(\boldsymbol{W}_{\sigma^2} \boldsymbol{\tilde{h}}_{t}+\boldsymbol{b}_{\sigma^2}\right) 
\end{aligned}
\end{equation}

All the hidden layer sizes are 256, and the latent space sizes are 128. Input and output segments at each timestep are of size 90. We use Adam \citep{kingma2014adam} for optimization, with a learning rate of 0.0008, exponential decay rates $\beta_1$ = 0.9, and $\beta_2$ = 0.999. We train the models for 5000 epochs, with a minibatch size 128. We set the regularization hyperparameter $\beta = 0$ at the beginning of training and gradually increase it until $\beta = 1$ is reached at epoch 1250.

\section{Experiments}
\subsection{Dataset}
The MIMIC-III Waveform Database Matched Subset \citep{moody2020mimic, johnson2016mimic} was used for the experiments. The database contains recordings collected from patients at various hospitals. Each session has multiple physiological signals, including PPG and ECG signals, sampled at a frequency of 125 Hz. We used the records of 43 healthy subjects and 12 subjects having AFib, including 30 males and 25 females, 23-84 years old. The dataset is made publicly available \footnote{https://github.com/khuongav/dvae\_ppg\_ecg}. Each record duration is 5 minutes. The first 48 s of each record were used as the training set, the next 12 s as the validation set, and the remaining 228 s as the test set. The preprocessing steps, including filtering, alignment, and normalization, were performed as described in \citep{tang2022robust}. We applied HeartPy \citep{van2019heartpy, van2019analysing} to identify peaks in PPG signals. Each long signal is split into 4-s chunks.
Each peak-to-peak interval was linearly interpolated to a length of 90 during training, which is the mean length of the intervals in the training set. The original interval length information can be preserved by making it an additional feature along with each normalized interval. Alternatively, we can apply padding instead of interpolation. However, we found that these did not contribute to improving the performance under the experimental setting. Original PP interval lengths were used as RR interval lengths in translated ECG signals during testing. This can be justified, as PPG recordings are used to analyze heart rate variability as an alternative to ECG \citep{lu2009comparison, aschbacher2020atrial}. Noise was added to the signals for robustness evaluation. The amplitudes of the baseline noise signals are 0.3, 0.4, and 0.1, and the frequencies are 0.3, 0.2 and 0.9 Hz, respectively. Gaussian noise of standard deviation 0.3.

\subsection{Evaluation Metrics}
\subsubsection{ECG Translation from PPG}
\indent\indent {Pearson's correlation coefficient} \(\left(\rho\right)\) measures how much an original ECG signal \( \boldsymbol{y}_{1:T} \) and its reconstruction \( \hat{\boldsymbol{y}}_{1:T} \) co-vary:
\begin{equation}
\rho=\frac{\left(\boldsymbol{y}_{1:T}-\bar{y}_{1:T}\right)^{\top}(\hat{\boldsymbol{y}}_{1:T}-\bar{\hat{y}}_{1:T})}{\left\|\boldsymbol{y}_{1:T}-\bar{y}_{1:T}\right\|_2\|\hat{\boldsymbol{y}}_{1:T}-\bar{\hat{y}}_{1:T}\|_2}
\end{equation}

{Root Mean Squared Error} (RMSE) measures the differences between the values of the original signal and its reconstruction:
\begin{equation}
    \text{RMSE}=\frac{\left\|\boldsymbol{y}_{1:T}-\hat{\boldsymbol{y}}_{1:T}\right\|_2}{\sqrt{n_y}}
\end{equation}

{Signal-to-Noise Ratio} (SNR) compares the level of the desired signal to the level of undesired noise:

\begin{equation}
    \text{SNR}=20\log\frac{\left\|\boldsymbol{y}_{1:T}\right\|_2^2}{\left\|\boldsymbol{y}_{1:T} - \hat{\boldsymbol{y}}_{1:T}\right\|_2^2}
\end{equation}

\subsubsection{AFib Detection}
Performance was measured by the Area under the Receiver Operating Characteristic (ROC-AUC), the Area under the Precision-Recall Curve (PR-AUC), and the F1 score. The PR-AUC is considered a better measure for imbalanced data.

\subsection{Implementation and Results}
\subsubsection{ECG Translation from PPG}

\begin{table}[h!]
\centering
\caption{ECG translation performance of different models. The top three rows show models' performance on healthy subjects, while the fourth row shows the performance on both the healthy and AFib subjects. If not specified, healthy subjects and clean signals is the default setting. The LSTM model \citep{tang2022robust} is subject-dependent, while the P2E-WGAN \citep{vo2021p2e} and our model are subject-independent.}
\label{tab:perf-tab}
\begin{tabular}{lccc}
 & Correlation & RMSE (mV) & SNR (dB) \\
\midrule
ADSSM &
  0.858 $\pm$ 0.174 &
  0.07 $\pm$ 0.047 &
  15.365 $\pm$ 11.053 \\
\midrule
ADSSM \\ w/o attention &
  0.823 $\pm$ 0.194 &
  0.08 $\pm$ 0.047 &
  13.013 $\pm$ 10.537 \\
\midrule
ADSSM \\ (healthy sub., \\ noisy sig.) &
  0.847 $\pm$ 0.174 &
  0.076 $\pm$ 0.049 &
  13.887 $\pm$ 10.58 \\
\midrule
ADSSM \\ (healthy \& \\AFib sub.) &
  0.804 $\pm$ 0.22 &
  0.078 $\pm$ 0.05 &
  12.261 $\pm$ 11.328 \\
\midrule
P2E-WGAN &
  0.773 $\pm$ 0.242 &
  0.091 $\pm$ 0.052 &
  9.616 $\pm$ 9.252 \\
\midrule
LSTM \\ (sub. dependent) &
  0.766 $\pm$ 0.234 &
  0.093 $\pm$ 0.053 &
  8.189 $\pm$ 9.560 \\
\bottomrule
\end{tabular}

\end{table}

Table \ref{tab:perf-tab} shows the performance of our model and compares it with other models in terms of means and standard deviations of $\rho$, RMSE and SNR. 
The correlation between the signals generated by our model and the reference signals is statistically strong, with a value $\rho$ of 0.858. Also, low values of RMSE (0.07) and high SNR (15.365) show strong similarities between them and reference ECG signals. When the attention mechanism is not applied on the input PPG, there is a notable decline in performance, with $\rho$ falling to 0.823, RMSE increasing to 0.08, and SNR decreasing to 13.013. This underscores the importance of the mechanism in providing relevant contexts for translation. The third row shows our model's performance on the noisy dataset. The negligible drop in metrics from 0.858 to 0.847 ($\rho$), 0.07 to 0.76 (RMSE), and 15.365 to 13.887 (SNR) demonstrates the robustness of our model. We attribute this to the probabilistic nature of the model, which better handles the measurement noise. As expected, the model performed worse on subjects with AFib due to the erratic patterns of the AFib signals (no visible P waves and an irregularly irregular QRS complex). In the next section, we show that the synthetic AFib signals are beneficial to the downstream detection task.

The P2E-WGAN model \citep{vo2021p2e}, a 1D deep convolutional generative adversarial network (4,064,769 parameters) for signal-to-signal translation, was recently proposed to translate PPG into ECG signals from a large number of subjects. P2E-WGAN achieved significantly lower performance than our model (645,466), requiring almost six times the parameters. Our model is less affected when data is scarce, which is common in healthcare. On the other hand, the LSTM model \citep{tang2022robust} is a deep recurrent neural network that was also recently proposed and built separately for each subject. The performance of our model, trained in a cross-subject setting, surpassed that of the LSTM model trained separately for each subject. These results prove the effectiveness and efficiency of our proposed sequential data structure. Further work with a larger number of subjects having AFib is needed to demonstrate that we can extend the model to new individuals. In addition, exploring strategies to manage the class imbalance problem \citep{johnson2019survey}, which arises from the fewer AFib records compared to the healthy ones, would be beneficial.

\begin{figure*}[htb!] 
  \centering
  \begin{subfigure}[b]{0.45\textwidth}
    \includegraphics[width=\textwidth]{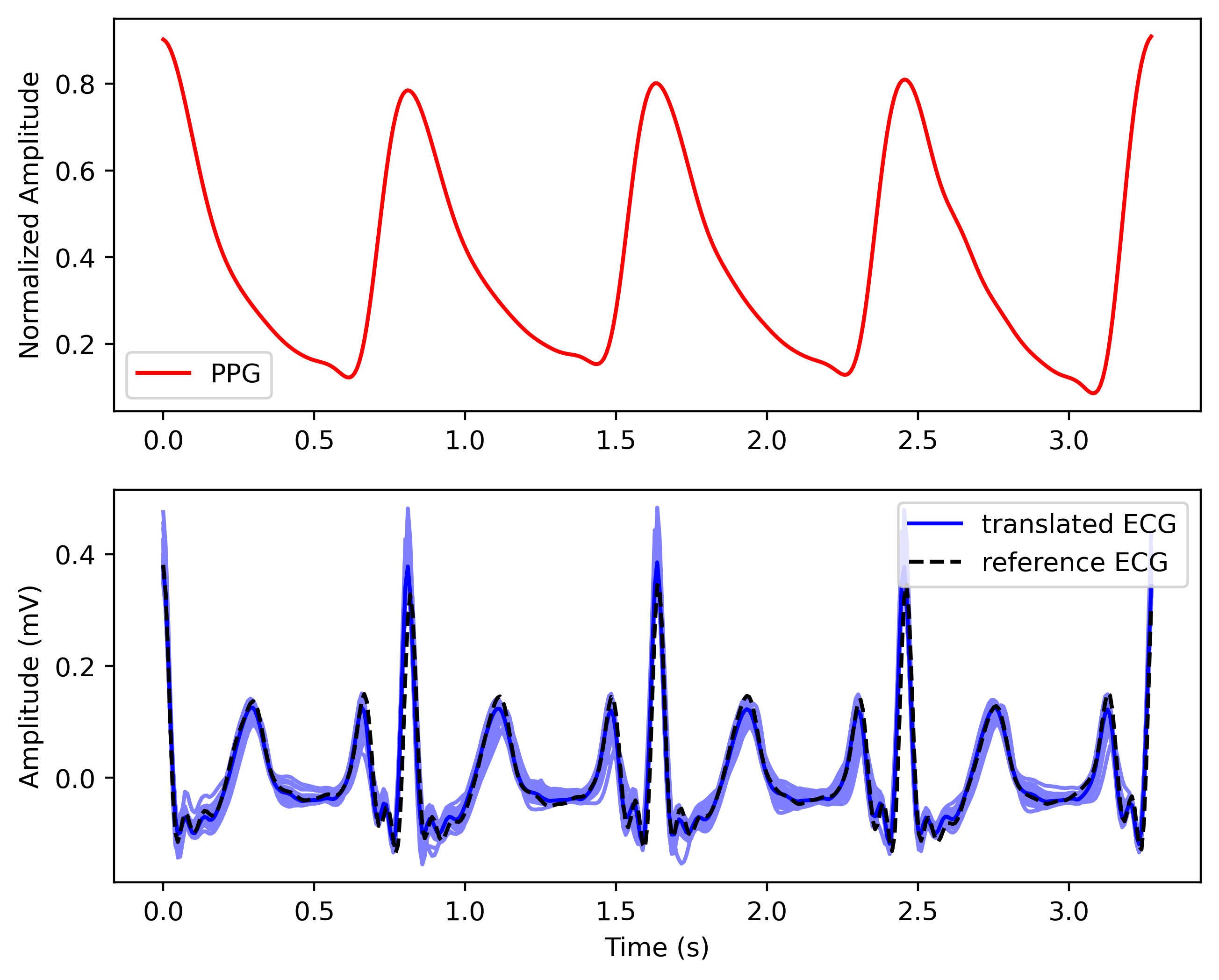}
    \caption{Clean input PPG}
    \label{fig:clean1}
  \end{subfigure}\hfil
  \begin{subfigure}[b]{0.45\textwidth}
    \includegraphics[width=\textwidth]{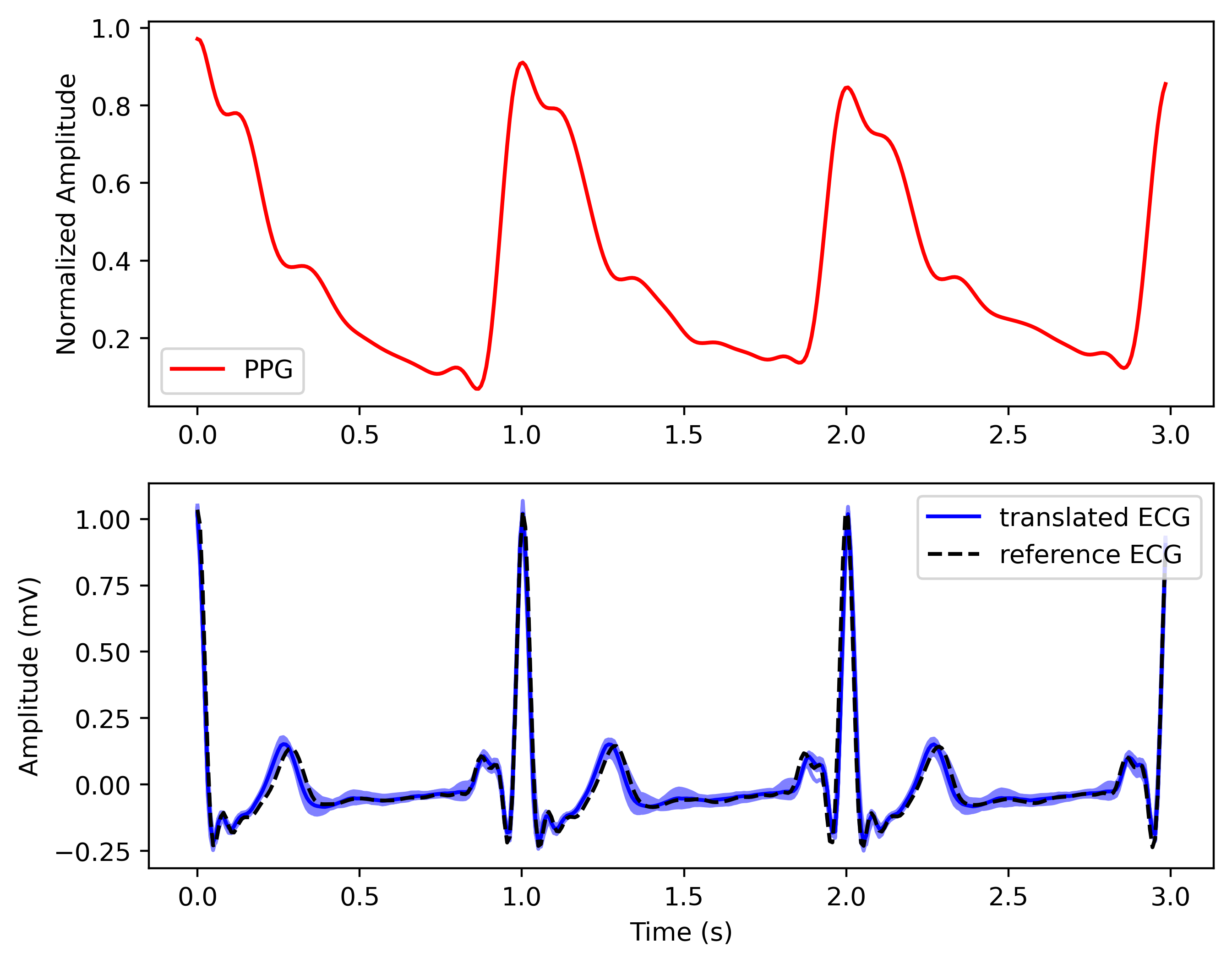}
    \caption{Clean input PPG}
    \label{fig:clean2}
  \end{subfigure}\par\medskip
  \begin{subfigure}[b]{0.45\textwidth}
    \includegraphics[width=\textwidth]{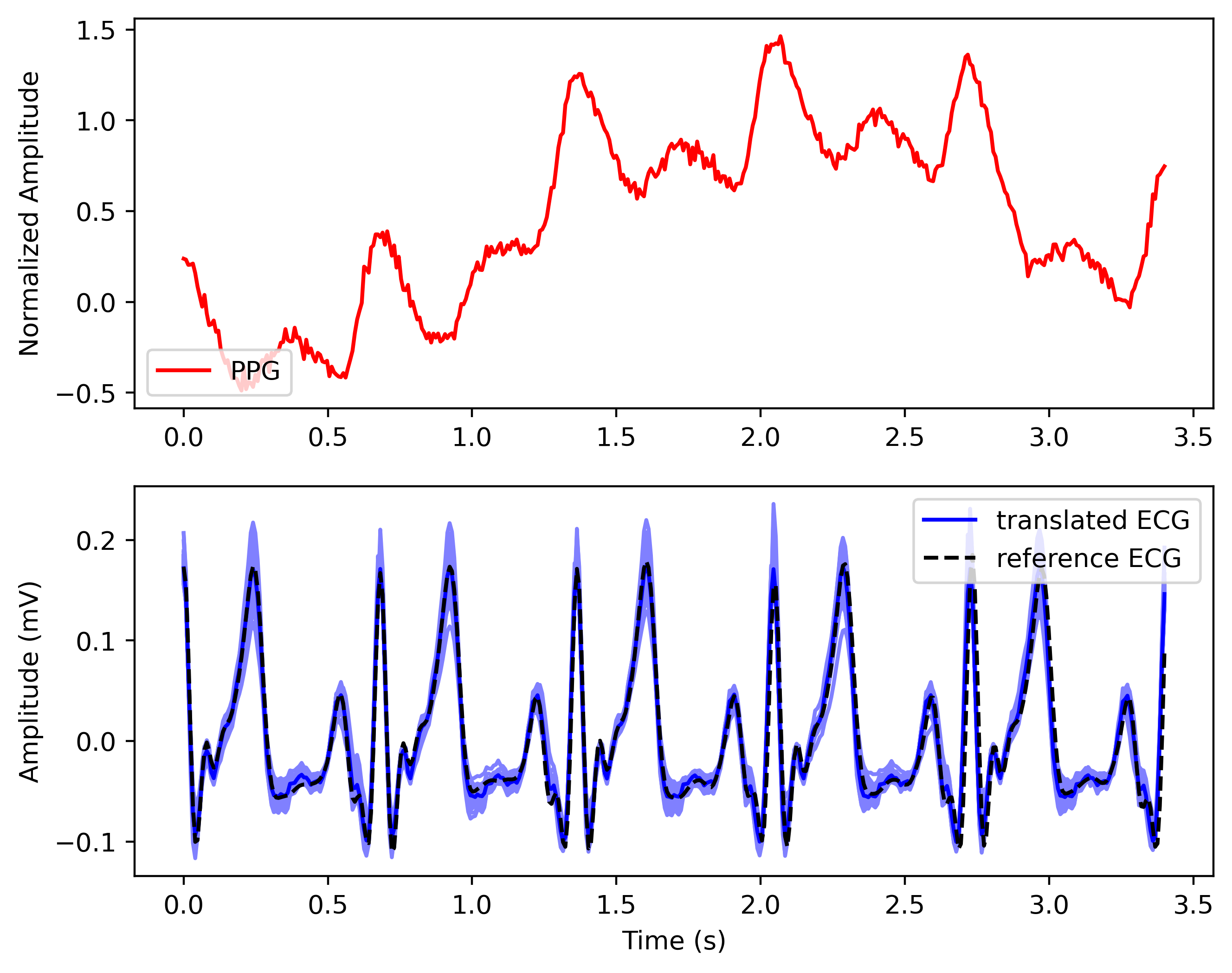}
    \caption{Noisy input PPG}
    \label{fig:noisy1}
  \end{subfigure}\hfil
  \begin{subfigure}[b]{0.45\textwidth}
    \includegraphics[width=\textwidth]{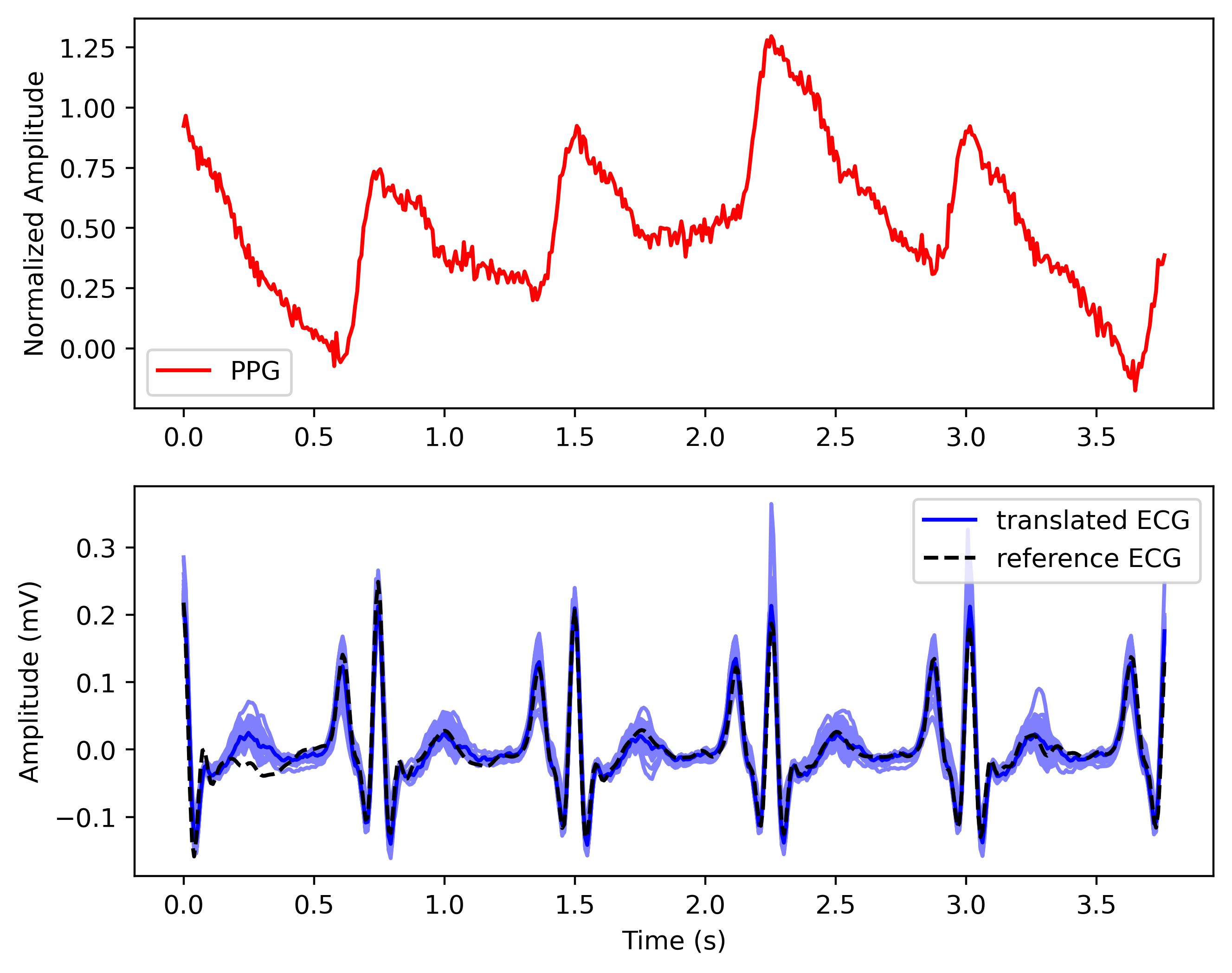}
    \caption{Noisy input PPG}
    \label{fig:noisy2}
    \end{subfigure}\par\medskip
  \begin{subfigure}[b]{0.45\textwidth}
    \includegraphics[width=\textwidth]{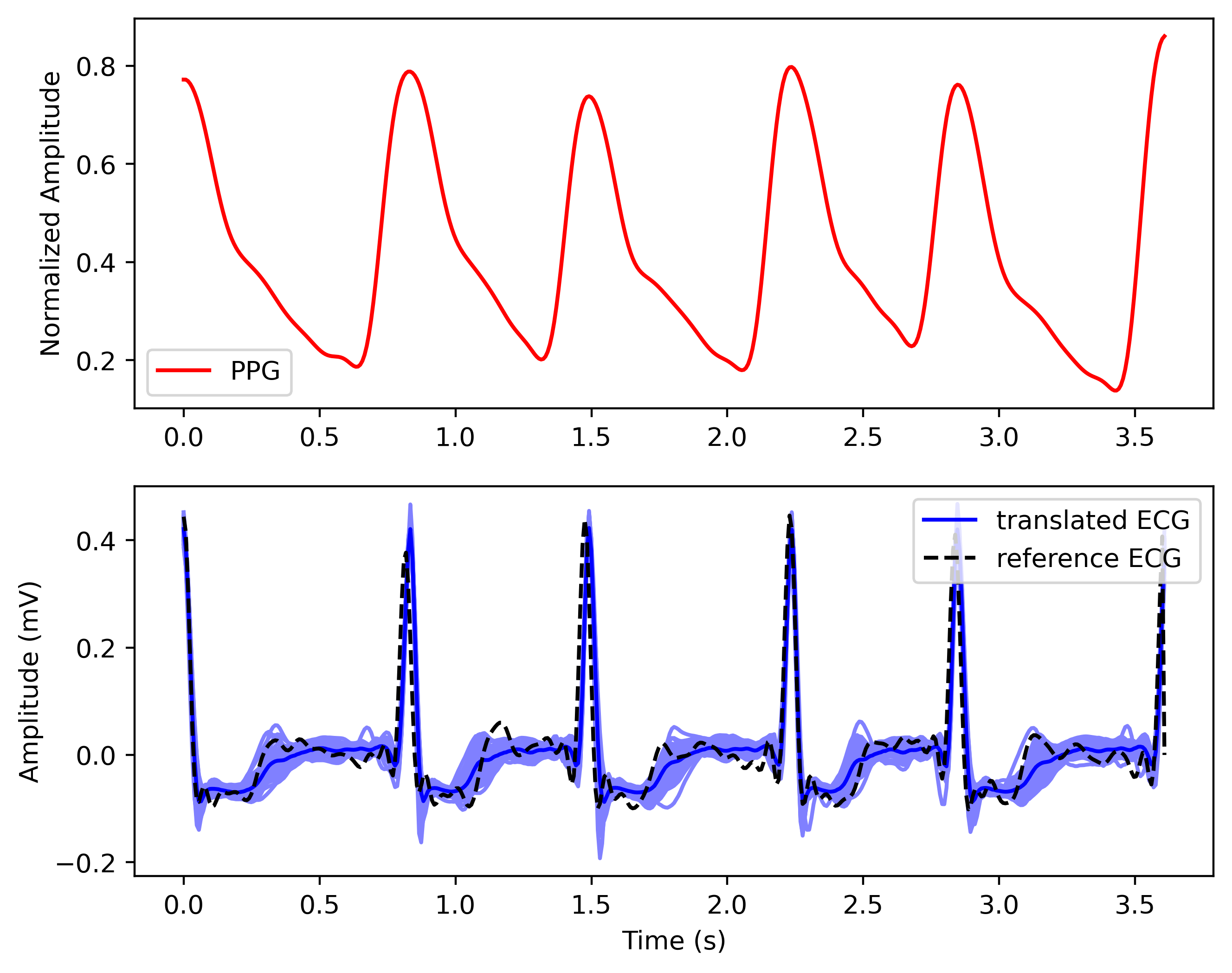}
    \caption{AFib input PPG}
    \label{fig:afib1}
  \end{subfigure}\hfil
  \begin{subfigure}[b]{0.45\textwidth}
    \includegraphics[width=\textwidth]{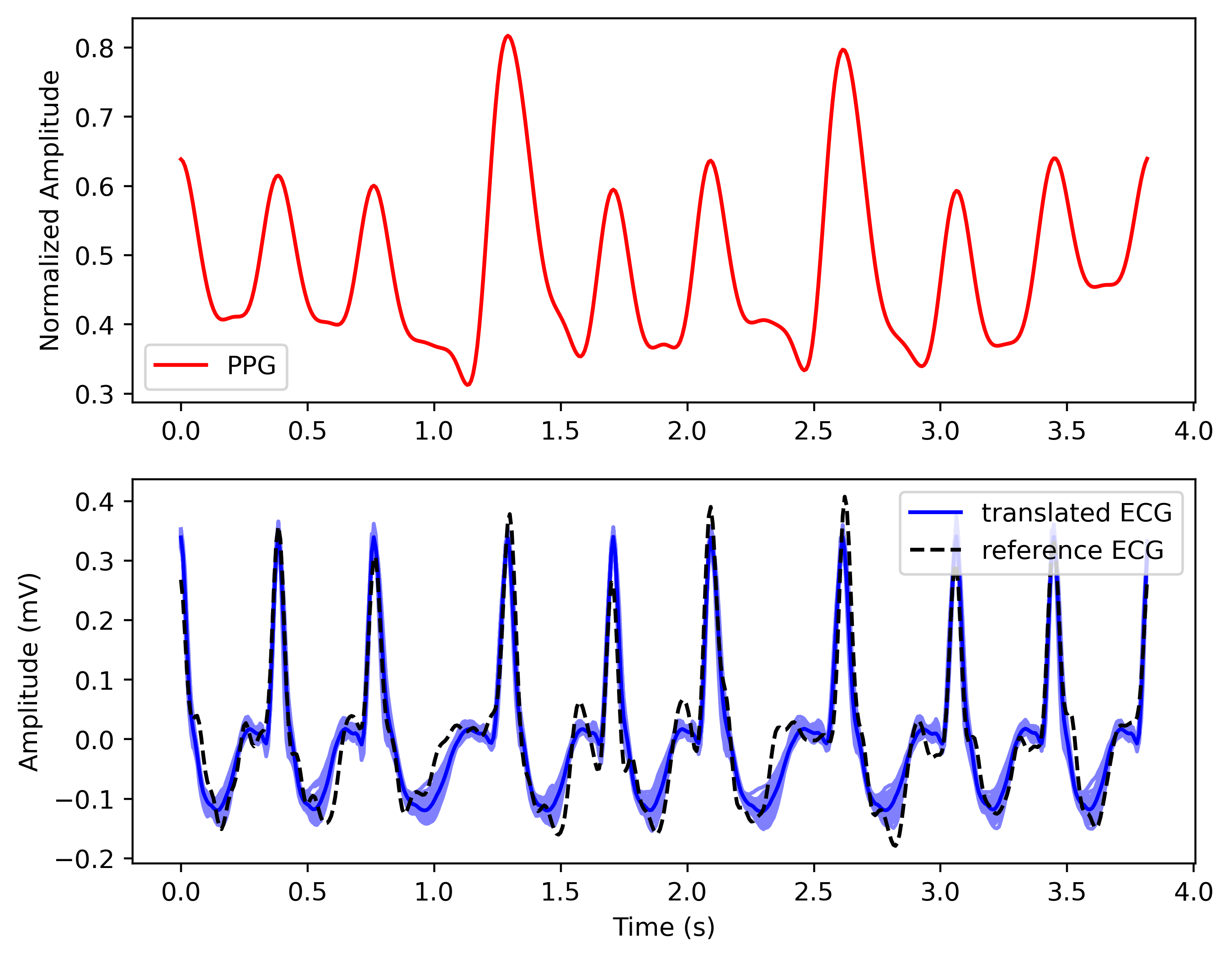}
    \caption{AFib input PPG}
    \label{fig:afib2}
  \end{subfigure}
  \caption{Examples of the translated ECG signals. In each subfigure: the top panel shows the input PPG waveform and the bottom panel shows the reconstructed ECG waveform compared with the reference waveform. The average ECG waveform (dark blue) of all possible pulses overlaid on each individual pulse (light blue).}
  \label{fig:examples}
\end{figure*}

In Figure \ref{fig:examples}, translated ECG waveforms are plotted with respect to the reference ECG waveforms of different heart rates. We can see that the model closely reconstructed the waveforms and maintained their essential properties, such as the missing P waves of the AFib ECG. In addition, we can be informed of the translation uncertainty by using a posterior on the latent embedding to propagate uncertainty from the embedding to the data. More specifically, with a distribution $p(\boldsymbol{z})$ on the latent feature our predictions will be $p_\theta\left(\boldsymbol{y} \mid \boldsymbol{x}\right)=\int p_{\theta_y}\left(\boldsymbol{y} \mid \boldsymbol{z}\right) p_{\theta_z}\left(\boldsymbol{z} \mid \boldsymbol{x}\right) d\boldsymbol{z}$. This would make the model more trustworthy and give patients and clinicians greater confidence in using it for medical diagnosis \citep{begoli2019need}. Future studies are expected to investigate methods to develop a fully Bayesian model and introduce a more flexible latent space \citep{tran2023fully, bendekgey2024unbiased}. Such advancements are advantageous in the medical field, particularly when data availability is limited or when uncertainty quantification and learning interpretable representations are essential.

\subsubsection{AFib Detection}

\begin{table}[h!]
\centering
\caption{AFib detection performance. The performance on the translated ECG is evaluated when the MINA model \citep{hong2019mina} is trained on real ECG but tested on synthetic ECG. The fusion performance is when the MINA model is extended to receive both real ECG and synthetic ECG inputs. x\% random time samples are omitted, simulating intermittent ECG recording, while synthetic ECG is always available.}
\label{tab:afib-tab}
\begin{tabular}{cccc}
& Real ECG & Translated ECG \\
\midrule
ROC-AUC & 0.995 $\pm$ 0.006 & 0.99 $\pm$ 0.004\\
PR-AUC & 0.987 $\pm$ 0.013 & 0.986 $\pm$ 0.007\\
F1 & 0.985 $\pm$ 0.009 & 0.944 $\pm$ 0.014\\
\midrule
Fusion & 30\% missing & 50\% missing & 70\% missing \\
\midrule
ROC-AUC & 0.992 $\pm$ 0.006 & 0.99 $\pm$ 0.006 & 0.99 $\pm$ 0.009 \\
PR-AUC & 0.986 $\pm$ 0.011 & 0.982 $\pm$ 0.012 & 0.981 $\pm$ 0.016 \\
F1 & 0.971 $\pm$ 0.01 & 0.969 $\pm$ 0.012 & 0.956 $\pm$ 0.046 \\ 
\bottomrule
\end{tabular}
\end{table}

We evaluated the performance of our model on the benefits of the translated ECG for the AFib detection task. To do so, we used a state-of-the-art AFib detection model, Multilevel Knowledge-Guided Attention (MINA) \citep{hong2019mina}, trained on real ECG signals, each of 10 s, and tested against synthetic. It should be noted that any pre-trained AFib detection model can be used in our pipeline. Table \ref{tab:afib-tab} reveals the mean detection performance of the model in the translated ECG that is close to that of the real ECG, ROC-AUC of 0.99 vs. 0.995, PR-AUC of 0.986 vs. 0.987, and F1 of 0.944 vs. 0.985. This implies that our model allows for the combined advantages of ECG's rich knowledge base and PPG's continuous measurement.

Furthermore, we extended the ability of the MINA model to receive real and translated ECG signals by incorporating the translated frequency channels into the model. In this scenario, both ECG and PPG signals can be measured simultaneously. This setting requires retraining of the MINA model on the fused real and synthetic ECG signal data set. To simulate the real-life setting where ECG measurement is intermittent while PPG input is continuous, we randomly zeroed out time samples with different probabilities: 30\%, 50\%, and 70\%. As shown in the bottom results of Table \ref{tab:afib-tab}, the performance remains almost unchanged in the fusion mode across the omission thresholds. Additionally, the model learns to utilize the sparse real ECG to marginally improve performance against only the translated ECG.

\section{Conclusion}

In this work, we present a novel attention-based deep state-space model to generate ECG waveforms with PPG signals as input. The results demonstrate that our model has the potential to provide a paradigm shift in telemedicine by bringing about ECG-based clinical diagnoses of heart disease via simple PPG assessment through wearable devices. 
Our model, trained on a small and noisy dataset, achieves an average Pearson's correlation of 0.847, RMSE of 0.076 mV, and SNR of 13.887 dB, demonstrating the efficacy of our approach. Significantly, our model enables the AFib monitoring capability in a continuous setting, assisting a state-of-the-art AFib detection model to achieve a PR-AUC of 0.986. Being a lightweight method also facilitates its deployment on resource-constrained devices.
In our future work, we aim to validate the generalizability of the model with other pairs of physiological signals. Our method allows for the screening and early detection of cardiovascular diseases in the home environment, saving money and labor, while supporting society in unusual pandemic situations.


\chapter{Composing Graphical Models with Generative Adversarial Networks for EEG Signal Modeling}

\section{Introduction}
\label{sec:intro}
Electroencephalogram (EEG) is a non-invasive technique that measures the spontaneous electrical activity of the brain. EEG has been a driver of studies from basic neurological research to clinical applications. EEG modeling is essential to understanding the underlying mechanisms that generate brain signals and serve to design experiments and test hypotheses \textit{in silico}. There exist extensive prior works on EEG computational models \citep{glomb2020computational} that derived principled neuroscience laws, empirically validated rules, or other domain expertise. Those are often in the form of general time-dependent and nonlinear partial differential equations. Nevertheless, they rely on strong assumptions which are not always generalizable. 
Further, those are slow to simulate and often suffer from model misspecifications. 

Generative Adversarial Networks (GANs) \citep{goodfellow2014generative} provide a powerful framework and tools for machine learning, especially for deep representation learning and generative models. Over the past few years, GANs have witnessed tremendous advancements and achieved state-of-the-art performance in a variety of prominent tasks, including photo editing, video prediction, text generation, and signal synthesis \citep{jabbar2020survey, vo2021p2e}. As a data-driven method, GANs are flexible and do not depend on rigid assumptions. Therefore, GANs hold great potential in modeling the inherent stochasticity and extrinsic uncertainty of EEG signals. 

Recent work \citep{hartmann2018eeg, aznan2019simulating, pascual2019synthetic} applying GANs in EEG synthesis tend to simply characterize the spatio-temporal characteristics of EEG data subject to latent spaces of basic distributions, e.g., Gaussian or uniform distributions. Such assumptions impose limitations in capturing the intrinsic dependence among latent variables. Also, the GANs require deeper networks to synthesize longer sequences, which are computationally expensive and challenging to train, e.g., vanishing or exploding gradient problems. Moreover,
the lack of inference capability in 
vanilla GANs hinder insight into structural information of EEG signals. 
On the other hand,
probabilistic graphical models \citep{koller2009probabilistic, wu2015bayesian} 
enable inference 
through
structured representations but often lack the capability to model arbitrarily complex distributions.

To address these challenges, we propose a novel GAN-based approach for EEG signal modeling that couples deep implicit likelihoods \citep{mohamed2016learning} with structured latent variable representations to combine their complementary strengths. Our method uses graphical models for representing underlying structures of the signals, and applies ideas from the Graphical-GAN \citep{li2018graphical} for effectively learning not only a generative model mapping from latent distributions to complex high-dimensional EEG data space but also an inverse inference model mapping from the data space to the latent space. Our study paves the way for leveraging implicit probabilistic models to comprehensively investigate the mechanisms that generate brain waves. 

\section{Methodology}
\subsection{EEG Signal Synthesis with GANs}
A GAN is a generative model trained by a pair of neural networks in a game-theoretic approach \citep{goodfellow2014generative}. In GANs, a discriminator neural network $D$ is trained to distinguish real from synthetic EEG signals, while a neural generator network $G$ is trained to generate EEG signals from a latent space to make them indistinguishable by the discriminator. With EEG signal $x$ drawn from data generating distribution $q(x)$, $z$ drawn from noise prior $p_z$, and $p(x)$ is the generator’s distribution over synthetic data, $G$ and $D$ jointly optimize the following objective:
\begin{equation}
\begin{aligned}
\mathcal{L}_{G A N}(G, D) & = \mathbb{E}_{x \sim q(x)}[\log D(x)] + \mathbb{E}_{z \sim p_{z}(z)}[\log (1-D(G(z))] \\ & = \mathbb{E}_{x \sim q(x)}[\log D(x)]  + \mathbb{E}_{x \sim p(x)}[\log (1-D(x))]
\end{aligned}
\end{equation}
The discriminator is expected to output a high probability for a valid EEG signal and a low probability for a synthesized one, corresponding to the values of $\log D(x)$ and $\log (1-D(G(z))$, respectively. $G$ and $D$ are trained simultaneously until $G$ is able to successfully fool $D$.

Following the proofs in \citep{goodfellow2014generative}, given a fixed generator G, the optimal discriminator is given by $D^{*}(x)=\frac{q(x)}{q(x)+p(x)}$


Under an optimal discriminator $D^{*}$, the generator minimizes the Jensen-Shannon (JS) divergence, which attains its minimum if and only if $p(x) = q(x)$.

\subsection{Conjoining GANs with Bayesian Networks}

\subsubsection{Generative and Inverse Inference Process}

\begin{figure}[h!]
    \begin{subfigure}{0.75\columnwidth}
    \includegraphics[width=\columnwidth]{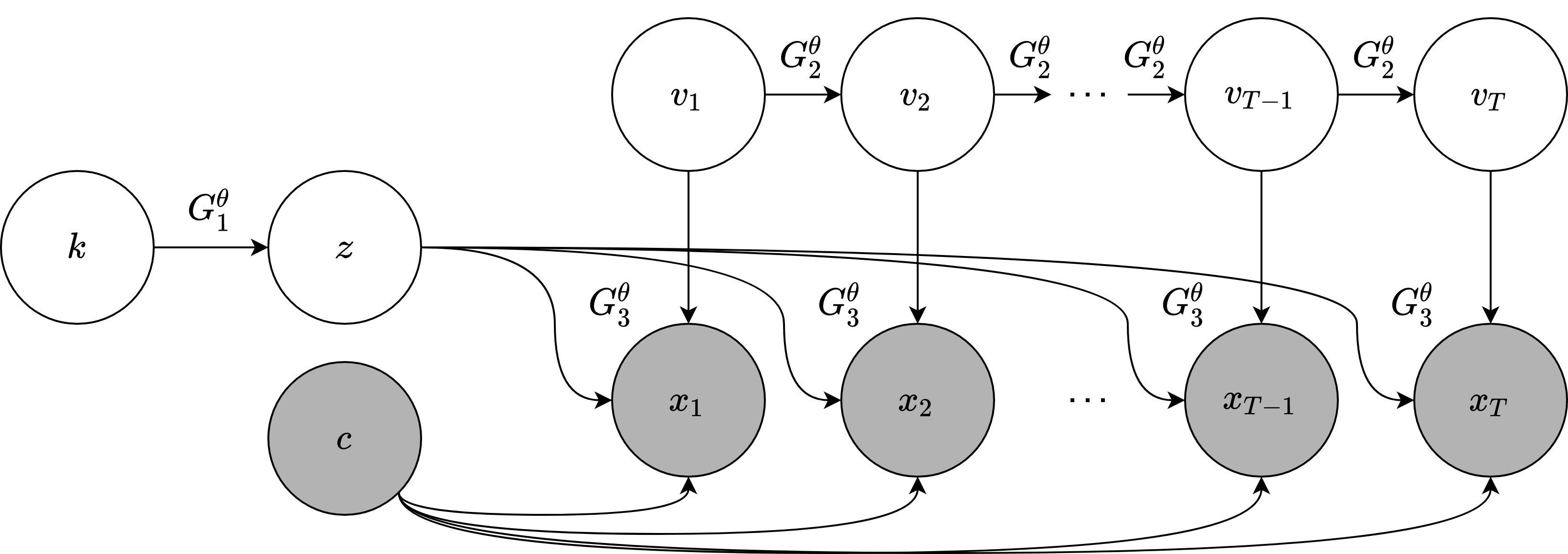}
     \caption{Generative model - $p$}
    \end{subfigure} 
  
  \vspace{10mm}

  \begin{subfigure}{0.75\columnwidth}
    \includegraphics[width=\columnwidth]{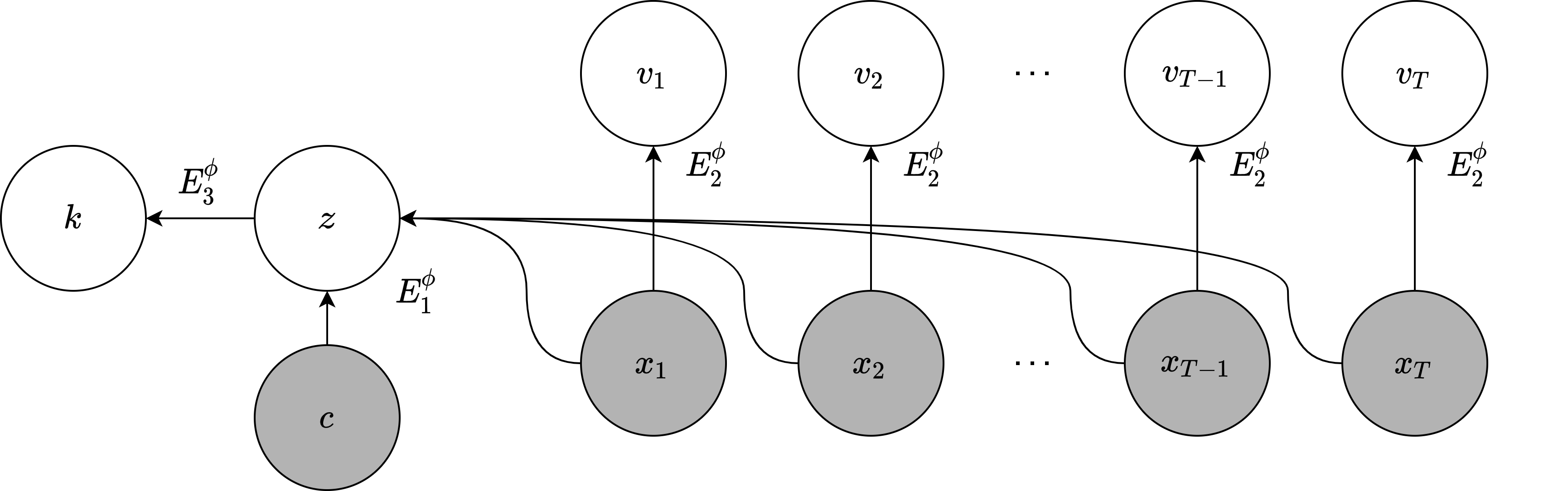}
    \caption{Inverse inference model - $q$}
  \end{subfigure}
  
  \caption{Directed graphical models for EEG signal modeling. Each time step corresponds to a $\delta$-second multi-channel signal. Shaded nodes represent observed variables. Clear nodes represent latent variables. Directed edges indicate statistical dependencies between variables.}
  \label{fig:ggans}
\end{figure}

As shown in Figure \ref{fig:ggans}, we model the generative process and the inverse inference process by a generative model and an inverse inference model in the Bayesian network. 
The framework exploits a Gaussian mixture model (GMM) to characterize the static latent variable structure with its capability to approximate arbitrary distributions, and a Markov model for the dynamic latent characterization.
We use the notations $p$ and $q$ to denote the generative and inverse inference models. 

The joint distribution of the generative model $p$ is
\begin{equation}
\begin{aligned}
&p(x_{1:T}, v_{1:T}, z, k, c) \\ & =p(k)p(z \mid k)p(c) \prod_{t=1}^{T} p(v_t \mid v_{t-1}) p(x_t \mid z, v_t, c) 
\end{aligned}
\end{equation} where $p(k)$ and $p(c)$ are simple prior distributions for Gaussian mixture indicator $k$ and condition $c$, e.g., a categorical distribution and a uniform distribution, $p(z \mid k)$ models a component selecting procedure for sampling noise $z$ which encodes the temporal-spatial relationships invariant across time, $v_t$'s form a first-order Markov chain, 
with $p(v_1|v_0) \sim \mathcal{N}(0, I)$, to encodes the temporal relationships variant across time, $p(x_t \mid z, v_t, c)$ specifies the conditional probability of the data at each time step $t$ given noise $z$, state $v_t$, and condition $c$, and is of interest for the final generation.

The distribution function $p(x_{1:T}, v_{1:T}, z, k, c)$ is parametrized as generator neural networks. It consists of three parts: $z^p = G_1(k^p)$, $v_{t+1}^p = G_2(v_t^p, \epsilon_t)$, $\epsilon_t \sim \mathcal{N}(0, I)$, and $x_t^p = G_3(z, v_t^p, c)$. $G_1$ is responsible for a mapping from the input prior to a mixed Gaussian distribution with respect to $k_p$. $G_2$ transitions to a new state $v_t^p$ given the previous state. $G_3$ uses noise $z^p$, state $v_t^p$, and condition $c$ to generate the synthetic $\delta$-second EEG signal $x_t^p$.

The joint distribution of the inverse inference model $q$ is
\begin{equation}
\begin{aligned}
& q(x_{1:T}, v_{1:T}, z, k, c)  \\ & = q(x_{1:T}) q(z \mid x_{1:T}, c) q(k \mid z) \prod_{t=1}^T q(v_t \mid x_t)
\end{aligned}
\end{equation} where each latent variable of the Markov structure is assumed to be independent using the mean-field approximation \citep{jordan1999introduction}. 
$q(x_{1:T})$ is the empirical data distribution, $q(z \mid x_{1:T}, c)$, $q(v_t \mid x_t)$, and $q(k \mid z)$ are of interest for the inference. Contrary to $p(v_{t+1} \mid v_{t})$, $q(v_t \mid x_t)$ models a dynamic tracing procedure for reconstructing the hidden features $v_t$. In contrast to $p(z \mid k), q(k \mid z)$ models a component tracing procedure for reconstructing the Gaussian mixture indicator $k$.

The distribution function $q(x_{1:T}, v_{1:T}, z, k, c)$ is parametrized as extractor neural networks. It consists of three parts: $z^q = E_1(x_{1:T}^q, c)$, $v_t^q = E_2(x_t^q)$, and $k^q = E_3(z^q)$. $E_1$ and $E_2$ are responsible for a mapping from original signals to noise $z^q$ and state $v_t^q$, respectively. $E_3$ infers within the latent space from $z^q$ to $k^q$.

\subsubsection{Learning Process}

Our goal is to learn the parameters of the generative model $p$ and the inverse inference model $q$ by jointly minimizing the Jensen-Shannon (JS) divergence
\begin{equation}
\textit{JS}(q(x_{1:T}, v_{1:T}, z, k, c) \| p(x_{1:T}, v_{1:T}, z, k, c))
\end{equation}
Expectation Propagation (EP) \citep{minka2013expectation}, a deterministic approximation algorithm, is proposed to utilize the locally structured data following \citep{li2018graphical}. The joint distributions can be factorized in terms of a set of factors $F_{\mathcal{G}} = \{\left(\mathrm{k}, \mathrm{z}\right),
\left(\mathrm{v}_{t}, \mathrm{v}_{t-1}\right),
\left(\mathrm{x}_{t}, \mathrm{v}_{t}, \mathrm{z}, c\right)\}$. For a factor $a$, the divergence of interest is
\begin{equation}
\mathit{JS}(q(a) \prod_{b \neq a} q(b) \| p(a) \prod_{b \neq a} p(b))
\end{equation}
EP iteratively minimize a local divergence in terms of each factor individually with the assumption that $\prod_{b \neq a} q(b) \approx \prod_{b \neq a} p(b)$. The divergence becomes
\begin{equation}
\mathit{JS}(q(a) \prod_{b \neq a} q(b) \| p(a) \prod_{b \neq a} q(b))
\end{equation}
Using the same proof sketch as in \citep{li2018graphical}, the divergence for factor $a$ is approximated as
\begin{equation}
\begin{aligned}
& \mathit{JS}(q(x_{1:T}, v_{1:T}, z, k, c) \| p(x_{1:T}, v_{1:T}, z, k, c)) \\ & \approx \mathbb{E}_{q}\left[\log \frac{2q(a)}{p(a)+q(a)}\right] + \mathbb{E}_{p}\left[\log \frac{2p(a)}{p(a)+q(a)}\right]
\end{aligned}
\end{equation}
The divergences are further averaged over all local factors as
\begin{equation}
\frac{1}{\left|F_{\mathcal{G}}\right|}\left[\mathbb{E}_{q}\left[\sum_{a \in F_{\mathcal{G}}} \log \frac{2q(a)}{p(a)+q(a)}\right]+\mathbb{E}_{p}\left[\sum_{a \in F_{\mathcal{G}}} \log \frac{2p(a)}{p(a)+q(a)}\right]\right]
\end{equation}

Individual parametric discriminators $D_a$ can be employed to estimate the local divergences as follows
\begin{equation}
\max _{\psi} \frac{1}{\left|F_{\mathcal{G}}\right|} \mathbb{E}_{q}\left[\sum_{a \in F_{\mathcal{G}}} \log \left(D_{a}(a)\right)\right]+\frac{1}{\left|F_{\mathcal{G}}\right|} \mathbb{E}_{p}\left[\sum_{a \in F_{\mathcal{G}}} \log \left(1-D_{a}(a)\right)\right]
\end{equation} where $\psi$ denotes the parameters in all discriminators. The discriminative models distinguish between the variables from the generative model $p$ and those from the inverse inference model $q$ as synthetic and original, respectively. 

\subsubsection{Optimization Objective}
Three discriminators $D_{3}$, $D_{2}$ and $D_{1}$ receive local variable pairs, i.e., $(k, z)$, $(v_t, v_{t-1})$, $(x_t, v_t, z, c)$, from either the generative model $p$ or the inverse inference model $q$, separately. The adversarial loss is as follows

\begin{equation}
\label{eqn:objective}
\begin{aligned}
& \mathcal{L}_{GAN}(G_*, E_*, D_*) \\ & =\mathbb{E}_{q}\left[\log D_{3}\left(\mathrm{k}^{q}, \mathrm{z}^{q}\right) + \log D_{2}\left(\mathrm{v}^{q}_{t}, \mathrm{v}^{q}_{t-1}\right) +\log D_{1}\left(\mathrm{x}_{t}^{q}, \mathrm{v}_{t}^{q}, \mathrm{z}^{q}, c\right)\right] \\
& \quad +\mathbb{E}_{p}\left[\log \left(1-D_{3}\left(\mathrm{k}^{p}, \mathrm{z}^{p}\right)\right) + \log \left(1-D_{2}\left(\mathrm{v}_{t}^{p}, \mathrm{v}_{t-1}^{p}\right)\right)\right. \\& \quad \quad +\left.\log \left(1-D_{1}\left(\mathrm{x}_{t}^{p}, \mathrm{v}_{t}^{p}, \mathrm{z}^{p}, c\right)\right)\right]
\end{aligned}
\end{equation}
All components are trained simultaneously in an adversarial process. Let $\theta$ and $\phi$ denote the parameters of $G_*$ and $E_*$, respectively. Iteratively, $D_*$ learn to maximize Equation \ref{eqn:objective} by updating $\psi$, while $G_*$ and $E_*$ learn to minimize Equation \ref{eqn:objective} by updating corresponding parameters $\theta$ and $\phi$, respectively.

In order to ensure the global consistency of an entire signal across time steps, a frequency domain loss is added as
\begin{equation}
\begin{aligned} 
\mathcal{L}_{f}(G_*)= \|\bar{r}({x}_{i, 1:T}^{q})-\bar{r}({x}_{i, 1:T}^{p})\|_{1} + \|\bar{\varphi}({x}_{i, 1:T}^{q})-\bar{\varphi}({x}_{i, 1:T}^{p})\|_{1}
\end{aligned}
\end{equation}
where $\bar{r}$ and $\bar{\varphi}$ refer to the average magnitude and phase across signals $i$ in a batch, respectively. They are computed by a fast Fourier transform (FFT). Hence, the total objective is
\begin{equation}
\min _{G_*, E_*} \max _{D_*} \mathcal{L}_{G A N} + \lambda \mathcal{L}_{f}
\end{equation}
\subsection{Network Architectures and Training Hyperparameters}

\begin{table}[h!]
\centering
\captionsetup{justification=centering}
\caption{Network architectures. Models having similar architectures are grouped together.}
\label{tab:net-table}
\begin{tabular}{ccl}
\multicolumn{1}{c|}{$\mathbf{G_2, D_3, D_2}$}                       & $\mathbf{G_3}$                              &     \\ \cline{1-2}
\multicolumn{1}{c|}{Linear 512, (SN), lReLU}                        & Linear 1536, lReLU                          &     \\
\multicolumn{1}{c|}{Linear 512, (SN), lReLU}                        & Reshape 96x16                               &     \\ \cline{2-2}
\multicolumn{1}{c|}{Linear 256, (SN), lReLU}                        & Upsample                                    &     \\
\multicolumn{1}{c|}{$G_2$ Linear 32}                                & Conv 6, BN, lReLU                           & X 4 \\
\multicolumn{1}{c|}{$D_{3}$ Linear 1, SN, Sigmoid}                  & Conv 6, BN, lReLU                           &     \\ \cline{2-2}
\multicolumn{1}{l|}{$D_{2}$ Linear 1, SN, Sigmoid}                  & \multicolumn{1}{l}{Conv 1, Tanh}            &     \\ \cline{1-2}
\multicolumn{1}{l}{}                                                & \multicolumn{1}{l}{}                        &     \\
\multicolumn{1}{c|}{$\mathbf{D_1}$}                                 & $\mathbf{E_2, E_1}$                         &     \\ \cline{1-2}
\multicolumn{2}{c}{\begin{tabular}[c]{@{}c@{}}Get  $x_t$ or $x_{[1,T]}$ (concatenated along channels) \\ Conv 1, lReLU - 96x256\end{tabular}}   &     \\ \cline{1-2}
\multicolumn{2}{c}{\begin{tabular}[c]{@{}c@{}}Conv 6, BN/SN, lReLU\\ Conv 6, Stride 2, BN/SN, lReLU\end{tabular}} & X 4 \\ \cline{1-2}
\multicolumn{2}{c}{Reshape 1536}                                                                                  &     \\ \cline{1-2}
\multicolumn{1}{c|}{\begin{tabular}[c]{@{}c@{}}Get $v_t$, $z$, $c$\\ Linear 256, SN, lReLU\end{tabular}} &
  \begin{tabular}[c]{@{}c@{}}$E_2$ Linear 32 \\ $E_1$ Linear 128\end{tabular} &
   \\
\multicolumn{1}{c|}{Join features of $x_t$, $v_t$, $z$, $c$}             &                                             &     \\
\multicolumn{1}{c|}{Linear 512, SN, lReLU}                          &                                             &     \\
\multicolumn{1}{c|}{Linear 1, SN, Sigmoid}                          &                                             &     \\ \cline{1-2}
\end{tabular}
\end{table}

Table \ref{tab:net-table} presents the architectures of the deep neural networks. Each time step corresponds to a 1-second EEG signal ($\delta$ = 1). All the feature maps have 96 channels. Leaky ReLU activation functions are applied to all layers, with the slope 0.1 to stimulate easier gradient flow. Batch normalizations (BN) \citep{ioffe2015batch} are used at each convolutional layer of the generators and extractors. Spectral normalizations (SN) \citep{miyato2018spectral} are applied to the discriminators to constrain their Lipschitz constants.  $c$ are subject embeddings as one-hot vectors. The sizes of $z$, $k$, and $v_t$, and $\epsilon_t$  are set at 128, 6, 32, and 16 respectively.

$G_1$ and $E_2$ are single-layer neural networks.
We use the reparameterization trick \citep{kingma2013auto} to estimate the gradients with the continuous variable $z$, and the Gumbel-Softmax trick  \citep{jang2016categorical} (the temperature of 0.1) to estimate the gradients with the discrete variable $k$.

$\lambda$ is set at 0.1 to have the training process driven mainly by the adversarial loss. In order to mitigate the issue of slow learning in regularized discriminators, a higher learning rate is provided to the discriminators than the generators and extractors by the Two Time-scale Update Rule (TTUR) \citep{heusel2017gans}. The models are trained with the Adam optimizer with the initial learning rate of 0.0004 for $D_*$, the learning rate of 0.0001 for $G_*$ and $E_*$, and the exponential decay rates $\beta_1$ = 0.5 and $\beta_2$ = 0.999. All weights are initialized using a zero-centered Gaussian distribution with a standard deviation of 0.02. We make the implementation publicly available \footnote{https://github.com/khuongav/Graphical-Adversarial-Modeling-of-EEG}. 

\section{Experiments}
\subsection{Dataset}
The 23-channel interictal EEG recordings from the CHB-MIT epilepsy dataset \citep{shoeb2009application} are used for the experiments. The dataset consists of scalp EEG from pediatric subjects with intractable seizures. We select a subset of 6 patients (chb01-03, chb05-06, chb10) having the same measurement setup, including males and females, 1.5-14 years old. Interictal periods are extracted at least 4-hour away before a seizure onset and after the seizure ends. The signals are low-pass filtered with a cut-off frequency at 50 Hz and scaled to the range $[-1,1]$. Overall, the dataset contains 43593 signals, from which 70\% are used for training and validation, and the other 30\% are used as the test set. 
Each signal is 10-second long (T=10), at a sampling rate of 256 Hz. Additionally, 339 ictal EEG signals are extracted for evaluating epilepsy seizure detection performance.

\subsection{Evaluation Metrics}

\textit{Sliced 2-Wasserstein distance} (SWD) \citep{bonneel2015sliced, flamary2021pot} quantifies the cost of transforming one distribution to another. It is an approximation to the 2-Wasserstein distance using 1D projections for a closed-form solution and is defined as
\begin{equation}
S \mathcal{W} \mathcal{D}_{2}(\mu, \nu)=\underset{\theta \sim \mathcal{U}\left(\mathbb{S}^{d-1}\right)}{\mathbb{E}}\left[\mathcal{W}_{2}^{2}\left(\theta_{\#} \mu, \theta_{\#} \nu\right)\right]^{\frac{1}{2}}
\end{equation}
where $\mu$ and $\nu$ are two probability measures, $\theta_{\#} \mu$ stands for the pushforwards of the projection $\mathbb{R}^{d} \ni X \mapsto\langle\theta, X\rangle$, and $\mathcal{U}\left(\mathbb{S}^{d-1}\right)$ is the uniform distribution on the hypersphere of $d$ dimensions. 

\textit{Spectral entropy} (SEN) measures the uniformity the of signal energy distribution in the frequency-domain. It is given by
\begin{equation}
\mathbb{H}(x)=-\sum_{f=0}^{f_{s} / 2} P(f) \log _{2}[P(f)]
\end{equation}
where $P$ is the normalised power spectral density, and $f_s$ is the sampling frequency of signal $x$.

\textit{Reconstruction error} (REC) measures the differences between the values of an original signal and its reconstruction $\tilde{x}$ as
\begin{equation}
REC = \|x_{1:T}^{q} -  \tilde{x}_{1:T}^{q}\|_{1}
\end{equation}


\subsection{Results and Discussion}

\begin{table}[h!]
\centering
\caption{Performances of different GAN models in interictal EEG signal synthesis and reconstruction tasks.}
\label{tab:res-table}
\begin{tabular}{llll}
                        & \multicolumn{1}{c}{SWD} & \multicolumn{1}{c}{REC}      & \multicolumn{1}{c}{SEN}    \\ \hline
Original data &        &                     & 0.620 $\pm$ 0.070 \\ \hline
\textbf{GMMarkov-GAN} & \textbf{1.16e-2}         & \textbf{0.0474 $\pm$ 0.0392} & \textbf{0.608 $\pm$ 0.063} \\ \hline
Markov-GAN  & 1.34e-2 & 0.0494 $\pm$ 0.0413 & 0.636 $\pm$ 0.070 \\ \hline
GMMarkov-GAN (w/o FFT)  & 1.70e-2 & 0.0519 $\pm$ 0.0438 & 0.585 $\pm$ 0.074 \\ \hline
Markov-GAN (w/o FFT)    & 1.78e-2 & 0.0530 $\pm$ 0.0391 & 0.583 $\pm$ 0.069 \\ \hline
C-BiGAN/ALI     & 2.13e-2 & 0.0562 $\pm$ 0.0415 & 0.539 $\pm$ 0.066 \\ \hline
\end{tabular}
\end{table}

\begin{figure}[h!]
    \centering
    \includegraphics[width=0.75\linewidth]{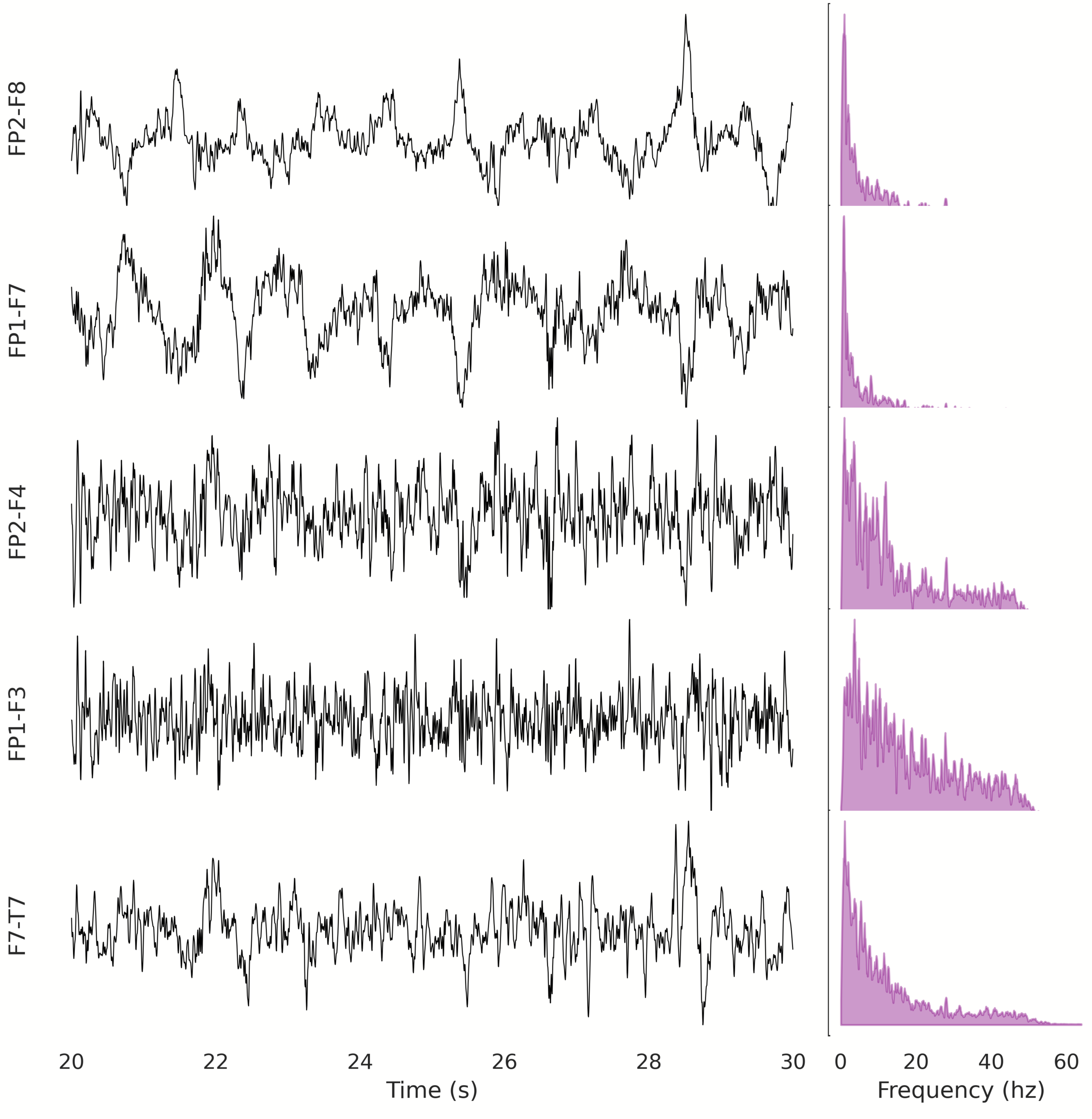}
    \caption{Last 10-second of a 30-second synthetic 23-channel EEG signal by the GMMarkov-GAN model, conditioned on patient 3. 5 channels with the highest standard deviations are shown.}
    \label{fig:syn}
\end{figure}

Table \ref{tab:res-table} presents the performance of our proposed approaches and the comparison with the BiGAN/ALI model \citep{dumoulin2016adversarially, donahue2016adversarial}. We denote its conditional version as C-BiGAN/ALI. GMMarkov-GAN is our model characterized by Gaussian mixture and Markov latent structures, while Markov-GAN is only with the Markov structure. C-BiGAN/ALI is the GAN with an inference capability but without a latent variable structure, in which the latent space is a simple Gaussian, and data at each timestep are generated independently.

\begin{figure}[h!]
\centering

  \begin{subfigure}[b]{0.65\textwidth}
    \includegraphics[width=\textwidth]{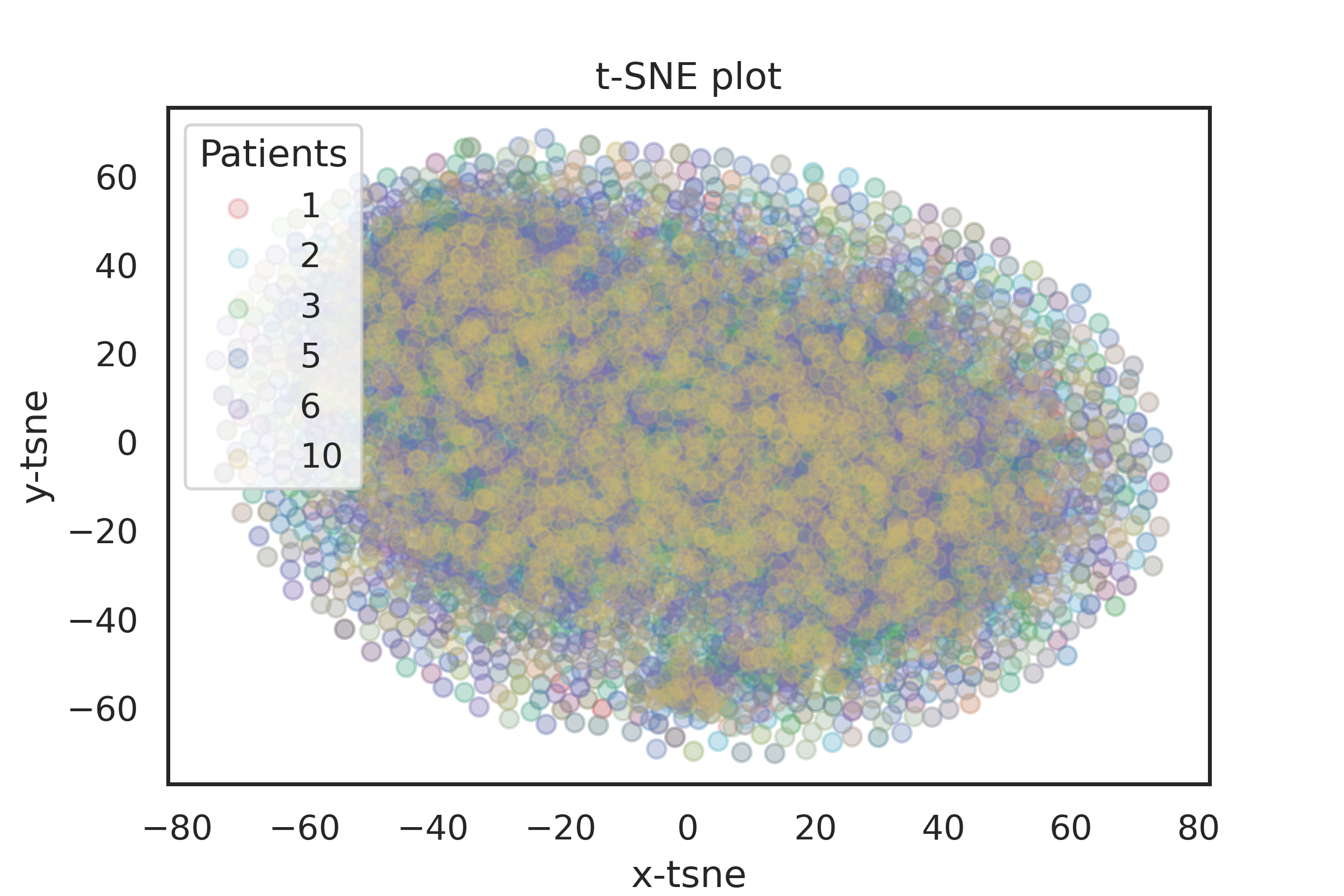}
    \caption{C-BiGAN/ALI}
  \end{subfigure}
  
  \begin{subfigure}[b]{0.65\textwidth}
    \includegraphics[width=\textwidth]{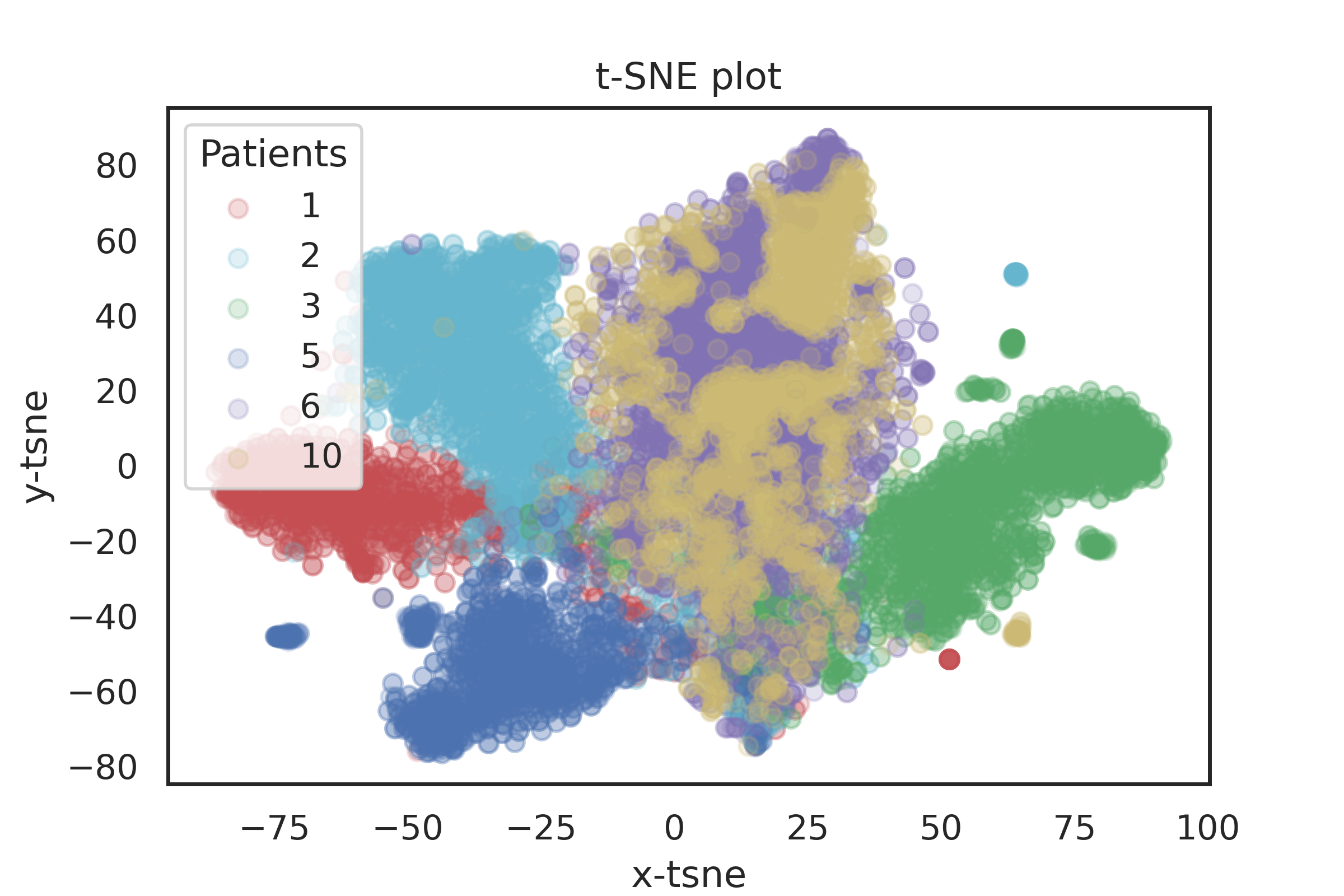}
    \caption{GMMarkov-GAN}

  \end{subfigure}

  \caption{t-SNE visualization of the static latent spaces.}
  \label{fig:tsne}
\end{figure}

Both the graphical GANs achieve significantly lower SWD, REC, and SEN differences than C-BiGAN/ALI, indicating that they are better at capturing the characteristics of EEG in both time and frequency domains. Besides, by encoding the invariant spatial-temporal features of EEG signals subject to the flexibility of a Gaussian mixture, GMMarkov-GAN enjoys better performance (SWD of 0.0173, REC of 0.0519, and SEN difference of 0.035) than the Markov-GAN. We attribute this to GMMarkov-GAN being able to learn a structured clustering of the latent space as shown in Figure \ref{fig:tsne}. 
These results prove the effectiveness of our proposed data structures and confirm our inverse inference strategy.

By training with the additional FFT loss, GMMarkov-GAN enjoys the highest performance (SWD of 0.0116, REC of 0.0474, and SEN difference of 0.012). It should be noted that the frequency-domain loss added little time for training, yet it noticeably improved the results.

\begin{figure}[h!]
    \centering
    \includegraphics[width=0.65\linewidth]{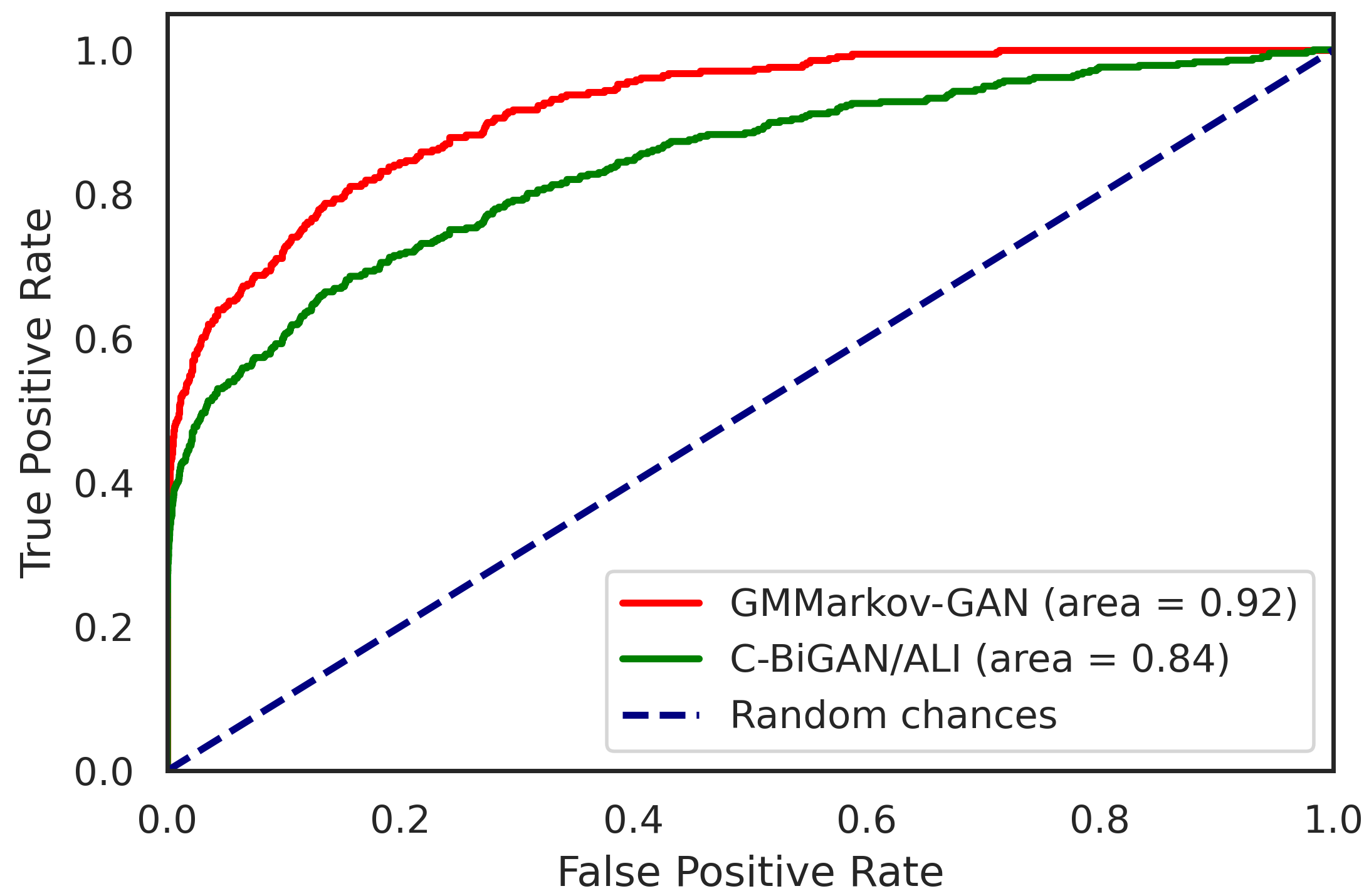}
    \caption{ROC curve for epilepsy seisure detection.}
    \label{fig:roc}
\end{figure}


In Figure \ref{fig:syn}, synthetic multi-channel EEG signals are plotted. The signals are naturally realistic across channels and show good fits in different frequency bands. Although our model is trained on 10 second-long signals, it can generate much longer sequences of 30 seconds, thanks to the Markov structure.

To demonstrate the efficacy of our generative and inverse mapping approach for auxiliary tasks, we further evaluated our approach in epilepsy seizure detection. As the model is trained on the interictal EEG signals, seizure segments are detected with reconstruction error thresholds in an anomaly detection framework. Figure \ref{fig:roc} shows a high detection performance from our model by the ROC curve with the area under the curve of 0.92, competitive with contemporary approaches in supervised learning \citep{siddiqui2020review}.
We plan to build on these results in our future work for interpreting more encoded features in the low-dimensional manifolds and further investigate the partial mode collapse issue of GANs \citep{bau2019seeing}.

\section{Conclusion}

In this work, we proposed an EEG modeling scheme that combines the strengths of probabilistic graphical models and generative adversarial networks.  
Our experimental results demonstrate that our method effectively characterized EEG latent variable structure via a Gaussian mixture and a Markov model. 
The structured representations can provide interpretability and encode inductive biases to reduce the data complexity of neural oscillations.
Our approach holds promise to new generative applications in neuroscience and neurology. 
Future directions include generalizing learning and inference algorithms with more complicated structures to truly model the underlying relationships at different scales spanning from the single cell spike train up to macroscopic oscillations.

\chapter{Deep Latent Variable Joint Cognitive Modeling of Neural Signals and Human Behavior}

\section{Introduction}

Current approaches to understanding brain function emphasize the search for statistical relationships between human behavior and individual physiological measures \citep[EEG, fMRI, fNIRS, etc.; e.g.][]{itthipuripat2019functional}. Behavioral measures, such as accuracy and speed of responses, reflect latent cognitive processes that underlie decision making that are not observed directly and must be inferred by cognitive models \citep{lee2014bayesian}. An ongoing challenge in computational cognitive neuroscience research is formulating the link between brain activity and latent cognitive processes. Here, we present a novel approach that allows a theoretical account of the cognitive process of decision-making, and artificial neural networks to estimate a joint latent space to link cognitive parameters to both neural signals and behavioral measures. This joint latent space model is a valuable new framework for computational cognitive neuroscience, allowing for new forms of inference and hypothesis generation.           

Previous work has focused on neurocognitive relationships between human neural data and behavioral data in decision-making tasks \citep{nunez2015individual, nunez2017attention, nunez2019latency, lui2021timing, turner2013bayesian, turner2016more}. The hierarchical Bayesian models used in these projects make strong predictions about the relationships between brain activity and the speed of decision-making.  These models typically make use of the drift-diffusion model \citep[DDM;][]{ratcliff2008diffusion}, a widely-used cognitive model in decision-making, as their generative model of choice and reaction time data. To integrate neural signals, these models require knowledge of previously discovered features of the neural data (e.g., known functional signals in the cognitive neuroscience literature) that are then linked by prescribed (usually linear) relationships to the latent cognitive variables in a Bayesian hierarchical model. The resulting \emph{neurocognitive} models test the relationship between neural signals and cognitive variables, and enhance the accuracy of predictions of behavior directly from brain signals \citep{turner2016more,nunez2017attention}. This can be thought of as one domain of the larger field of \textit{model-based cognitive neuroscience} \citep{Forstmann2015}. 

A limitation of this approach is that we must know in advance which brain signals are possibly linked to cognitive functions. However, advances in frameworks and tools for neuroscience allow for the discovery of previously unknown neural features that we could use to explain latent cognitive variables. Ideally, such frameworks operate across observations, experimental manipulations, and individual differences.  Deterministic models that leverage deep learning have been proposed for learning feature representation of EEG data to analyze and decode brain activity \citep{roy2019deep}. As a notable example, \citet{sun2022decision} have proposed a SincNet-based neural network that made use of EEG signals to learn the latent cognitive variables of the DDM on individual decisions. This approach identifies time windows of information processing and frequency bands that can be used to predict latent processes directly from EEG data as a trial-level association between neural features, choice, and response time. 


This work aims to develop a deep probabilistic method for linking neural data from EEG to the latent parameters of a cognitive model.
The innovation of our work lies in the use of a theoretical account of the cognitive process.  This theoretical account drives the analysis of neural and behavioral measures. The framework allows for one-step, joint inference on integrative neurocognitive models that map EEG and behavior into a joint latent space. Uniquely, this new approach has the potential to allow us to generate task-relevant EEG signals from behavioral data, and \emph{predict modulation of EEG signals by cognitive model parameters}.  By combining the exploratory potential of modern latent variable methods with the theoretical appeal of human-interpretable cognitive model parameters, the proposed technique can be used to make predictions of brain signals and cognitive parameters in future experiments to test neurocognitive theories.

\begin{figure*}[!htb]
\vspace*{-10pt}
  \centering

  \begin{subfigure}{\textwidth}
    \centering
    \includegraphics[width=1.09\textwidth]{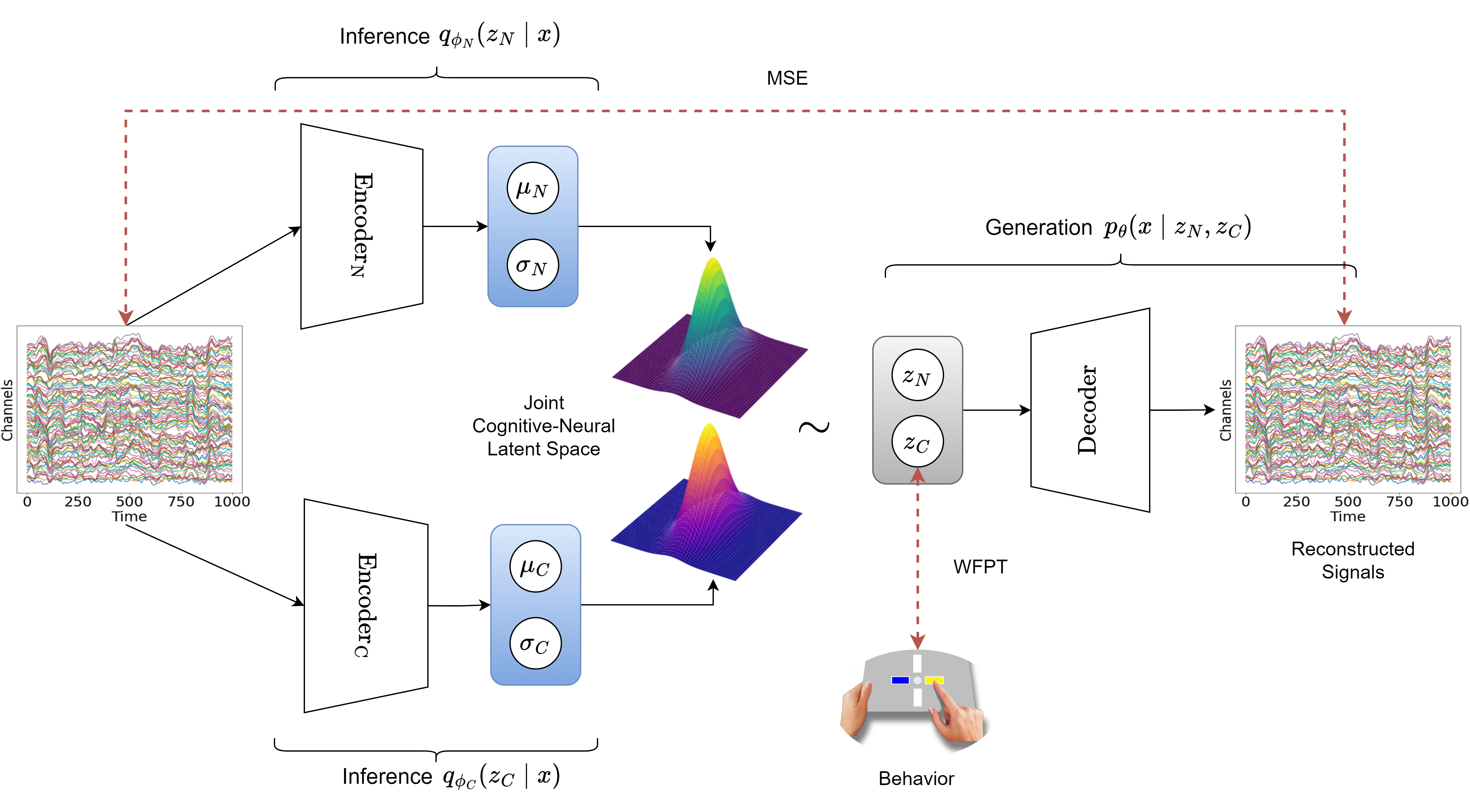}
    \vspace{-4mm}
    \subcaption{Generative Process}
  \end{subfigure}
  
  \vspace{0.5cm}
  
  \begin{subfigure}{\textwidth}
    \centering
    \includegraphics[width=1.0\textwidth]{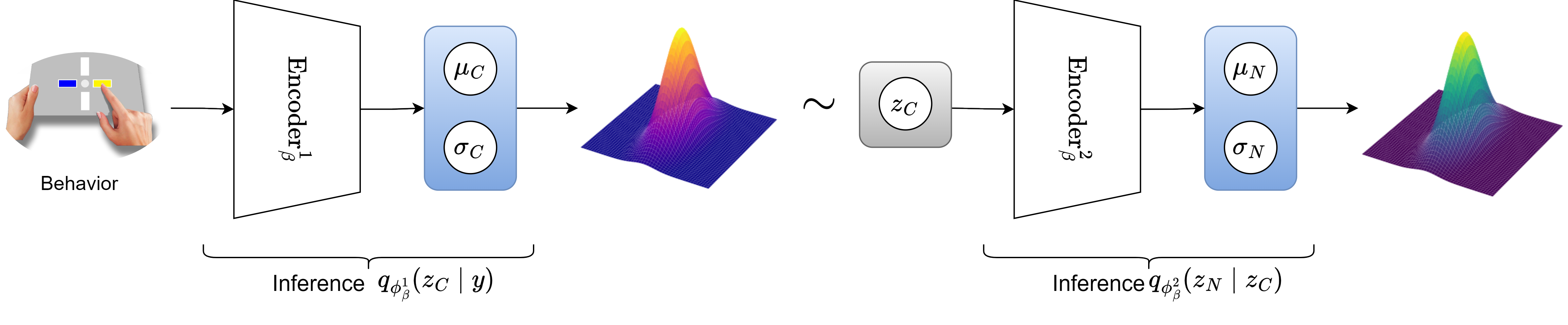}
    \vspace{-4mm}
    \subcaption{Regularized Discriminative Process}
  \end{subfigure}

  \caption{The Neurocognitive VAE. After the generative process (a) learns the joint latent neurocognitive variables (Section \ref{gen}), the regularized discriminative process (b) retrofits its hierarchical latent space to the joint latent space (Section \ref{disc}). Inference networks $q$ and Generation networks $p$ contain neural network parameters $\theta$ and $\phi$. Black arrows: flows of operations. Red arrows: loss functions. MSE and WFPT stand for Mean Squared Error and Wiener First Passage Time, respectively. The heatmaps represent the probability distributions in the latent spaces. Plasma color maps are for the drift-diffusion variables ($z_C \in \mathbb{R}^{3}$), while greenery color maps are for residual neural variables ($z_N \in \mathbb{R}^{32}$). Blue blocks contain $\mu$ and $\sigma$, which are the parameters of the multivariate Gaussian latent spaces. Gray blocks contain $z$ sampled ($\sim$) from the distributions. The variables $x$ and $y$ represent EEG signals and choice-RTs, respectively. Each trapezoid represents a different convolutional neural network (see Table \ref{tab:architecture} for detailed architectures).}
  \label{fig:architecture}
\vspace*{-20pt}
\end{figure*}

\section{Neurocognitive Variational Autoencoders}

\subsection{Generative EEG Modeling with VAEs}
\sloppy Consider first a data set $\mathcal{P} \defeq \left\{ \mathcal{D}_{1}, \ldots, \mathcal{D}_{M} \right\}$ containing $M$ subjects, where each subject $\mathcal{D}_{m} \defeq \left\{\boldsymbol{x}_{1}, \ldots, \boldsymbol{x}_{I}\right\}$ consists of $I$ trials $\boldsymbol{x}_{i} \in$ $\mathbb{R}^{C \times T}$ that are EEG signals of $C$ channels by $T$ time samples. Throughout the paper, the subscript $m$ is omitted when we refer to only one subject or when it is clear from the context. 

For each subject $m$, we aim to learn an EEG generative process with a latent-variable model 
comprising of a fixed Gaussian prior over latent variables $p(\boldsymbol{z}) = \mathcal{N}(\boldsymbol{z} \mid \boldsymbol{0}, \boldsymbol{I})$, where $\boldsymbol{I}$ is the identity covariance matrix, and a parametric non-linear Gaussian likelihood $p_\theta(\boldsymbol{x}  \mid  \boldsymbol{z})$. The learning process finds $\theta$ such that the Kullback-Leibler (KL) divergence is minimized between the true data generating distribution $p_{\mathcal{D}}$ and the model $p_\theta$:
\begin{equation}
\begin{aligned}
& \underset{\theta}{\arg \min }\, K L\left(p_{\mathcal{D}}(\boldsymbol{x} ) \| p_\theta(\boldsymbol{x} )\right) \\ 
&=\underset{\theta}{\arg \max }\, \mathbb{E}_{p_{\mathcal{D}}(\boldsymbol{x} )}\left[\log p_\theta(\boldsymbol{x} )\right]
\end{aligned}
\end{equation}
where $p_\theta(\boldsymbol{x} )=\int_{\mathcal{Z}} p_\theta(\boldsymbol{x} \mid \boldsymbol{z}) p(\boldsymbol{z}) d \boldsymbol{z}$ is the likelihood of data point $\boldsymbol{x}$, approximated by averaging over the latent $\boldsymbol{z}$.

Nevertheless, estimating $p_\theta(\boldsymbol{x} )$ is typically intractable. This issue can be mitigated by introducing a parametric inference model $q_\phi(\boldsymbol{z} \mid \boldsymbol{x})$ to construct a variational evidence lower bound on the log-likelihood $\log p_\theta(\boldsymbol{x} )$ as follows:
\begin{equation} 
\begin{aligned}
& \mathcal{L}(\boldsymbol{x}; \theta, \phi) \\ & \qquad \defeq  \log p_\theta(\boldsymbol{x})-K L\left(q_\phi(\boldsymbol{z} \mid \boldsymbol{x}) \| p_\theta(\boldsymbol{z} \mid \boldsymbol{x})\right) \\
& \qquad=\mathbb{E}_{q_\phi(\boldsymbol{z} \mid \boldsymbol{x})}\left[\log p_\theta(\boldsymbol{x} \mid \boldsymbol{z})\right]-K L\left(q_\phi(\boldsymbol{z} \mid \boldsymbol{x}) \| p(\boldsymbol{z})\right)
\end{aligned}
\end{equation}
Taking the likelihood model $p_\theta(\boldsymbol{x} \mid \boldsymbol{z})$ to be a decoder and the inference model $q_\phi(\boldsymbol{z} \mid \boldsymbol{x})$ to be an encoder, a variational autoencoder \citep[VAE;][]{kingma2013auto, sohn2015learning} considers this objective from a deep probabilistic autoencoder perspective. Here, $\theta$ and $\phi$ are neural network parameters, and learning takes place via stochastic gradient ascent using unbiased estimates of $\nabla_{\theta, \phi} \frac{1}{n} \sum_{i=1}^n \mathcal{L}\left(\boldsymbol{x}_i; \theta, \phi\right)$.

In the following sections, we extend the traditional VAE to create the Neurocognitive VAE (NCVA) (Figure \ref{fig:architecture}). This model allows us to model a joint distribution of neural and behavioral data. Instead of a training technique that encourages disentanglement, as in $\beta$-VAE \citep{higgins2016beta}, NCVA imposes restrictions on latent space by using a cognitive model that provides interpretability and controllable generation.

\subsection{Disentangled Cognitive Latent Space of EEG} \label{gen}

Now consider the data $\mathcal{D}_{m} \defeq \left\{\left(\boldsymbol{x}_{1}, \boldsymbol{y}_{1}\right), \ldots,\left( \boldsymbol{x}_{I}, \boldsymbol{y}_{I}\right)\right\}$, consisting, on the one hand, of $N$ trials of the EEG data $\boldsymbol{x}_{i}$ and, on the other hand, of the corresponding choice response times (choice-RT) $\boldsymbol{y}_{i}$. Both $\boldsymbol{x}_{i}$ and $\boldsymbol{y}_{i}$ are associated with a context vector $\boldsymbol{c}_i$ (where the applicable context might be an experimental condition; say, noise conditions $\boldsymbol{c}_{i}$). For mathematical simplicity, the context vector $\boldsymbol{c}$ is not mentioned when we refer to one of the data modalities.  

Crucially, we propose a generative model with two sources of variation: $\boldsymbol{z}_C$, which is cognitively specific, and $\boldsymbol{z}_N$, which captures any residual neural variations left in $\boldsymbol{x}$. We assume the approximate posterior $q_\phi(\boldsymbol{z}_N, \boldsymbol{z}_C \mid \boldsymbol{x})$ has the following fully factorized form:
\begin{equation} 
\begin{aligned}
 q_\phi\left(\boldsymbol{z}_N, \boldsymbol{z}_C \mid \boldsymbol{x}\right) &= q_{\phi_N}\left(\boldsymbol{z}_N \mid \boldsymbol{x}\right)q_{\phi_C}\left(\boldsymbol{z}_C \mid \boldsymbol{x}\right) \\ 
 q_{\phi_N}\left(\boldsymbol{z}_N \mid \boldsymbol{x}\right) &= \mathcal{N}\left(\boldsymbol{z}_N \mid \boldsymbol{\mu}_{\phi_N}(\boldsymbol{x}), \text{diag}\left(\boldsymbol{\sigma}^2_{\phi_N}(\boldsymbol{x})\right)\right) \\
 q_{\phi_C}\left(\boldsymbol{z}_C \mid \boldsymbol{x}\right) &= \mathcal{N}\left(\boldsymbol{z}_C \mid \boldsymbol{\mu}_{\phi_D}(\boldsymbol{x}), \text{diag}\left(\boldsymbol{\sigma}^2_{\phi_D}(\boldsymbol{x})\right)\right)
\end{aligned}
\end{equation}
A Gaussian prior over latent variables $p(\boldsymbol{z}_C)$ can be chosen for each subject. We use subject priors obtained from a Bayesian hierarchical fitting of a DDM using the Markov chain Monte Carlo (MCMC) \citep{nunez2019latency}.

We learn the generative model by maximizing the lower bound on $\log p_\theta(\boldsymbol{x}, \boldsymbol{y})$ as:
\begin{equation}
\label{eqn:elbo}
\begin{aligned}
& \mathcal{L}(\boldsymbol{x}, \boldsymbol{y} ; \theta, \phi_N, \phi_C) \\ 
& = \mathbb{E}_{q_\phi(\boldsymbol{z}_N, \boldsymbol{z}_C \mid \boldsymbol{x})}\left[\log p_{\theta}(\boldsymbol{x} \mid \boldsymbol{z}_N, \boldsymbol{z}_C) + \log p(\boldsymbol{y} \mid \boldsymbol{z}_C)\right] \\ 
& \quad - K L\left(q_{\phi_N}(\boldsymbol{z}_N \mid \boldsymbol{x}) \| p(\boldsymbol{z}_N)\right) \\ 
& \quad - K L\left(q_{\phi_C}(\boldsymbol{z}_C \mid \boldsymbol{x}) \| p(\boldsymbol{\boldsymbol{z}_C})\right)
\end{aligned}
\end{equation}
where 
$p_{\theta}(\boldsymbol{x} \mid \boldsymbol{z}_N, \boldsymbol{z}_C)=\mathcal{N}(\boldsymbol{x} \mid \boldsymbol{\mu}_{\theta}(\boldsymbol{z}_N, \boldsymbol{z}_C), \boldsymbol{I})$ 
and $p(\boldsymbol{y} | \boldsymbol{z}_C)$ can be any neurocognitive likelihood. This work applies the Wiener First Passage Time distribution \citep[WFPT;][]{navarro2009fast} corresponding to the lower boundary: 
\begin{equation}
\begin{aligned}
& p(\boldsymbol{y} | \boldsymbol{z}_C) \\ 
& = \text { Wiener }(\text{RT} \mid \alpha, \tau, \delta) \\ 
& = \frac{\pi}{\alpha^2} e^{-\frac{1}{2}\left(\alpha \delta+\delta^2(RT-\tau)\right)} \\
& \quad \times \sum_{k=1}^{+\infty}\left[k \sin\left(\frac{\pi k}{2}\right) e^{- \frac{k^2 \pi^2}{2\alpha^2}(RT-\tau)}\right] 
\end{aligned}
\end{equation}
The probability at the upper boundary is obtained by setting $\delta' = - \delta$. $\boldsymbol{z}_C$ comprises of three parameters including drift rate $\delta$, boundary $\alpha$, non-decision time (ndt) $\tau$. The bias towards correct or incorrect responses is fixed at 0.5, that is, the starting point is always unbiased.

The joint inference is performed using only EEG $\boldsymbol{x}$ to ensure that encoder $\theta_C$ would learn to extract neural features that are tailored to cognitive parameters, without relying on choice-RT $\boldsymbol{y}$. This has the advantage of providing more accurate trial-level parameter estimates that are associated with the EEG data.

Note that the dimension of the cognitive space is significantly lower than that of the residual neural space. This facilitates the representation of the variation in neural signals only through flexible $\boldsymbol{z}_N$. Maximizing the likelihood of observing neural signals does not guarantee decoder $\theta$ utilizing $\boldsymbol{z}_C$ to output $\boldsymbol{x}$. 
In the next section, we present an approach to capture the correlation between behavior and cognition, as well as the mapping of the variability of behavior and cognition to neural signals.

\subsection{Structured EEG Modeling from Behavior} \label{disc}
Here, we propose a discriminative model regularized by the generative model learned in the previous section. 
We aim to discriminatively learn the distribution of the cognitive parameters conditioned on behaviors, and the distribution of the neural latent variables conditioned on cognitive parameters.
The joint latent space inferred from the behavior can be factorized into the two-level latent space as follows:
\begin{equation}
q_{{\phi}_{B}}\left(\boldsymbol{z}_N, \boldsymbol{z}_C \mid \boldsymbol{y}_{i}\right) = q_{\phi^2_B}\left(\boldsymbol{z}_N \mid \boldsymbol{z}_C\right)q_{\phi^1_B}\left(\boldsymbol{z}_C \mid \boldsymbol{y}_{i}\right)
\end{equation}
Inspired by \citet{suzuki2016joint}, we learn the following approximations, w.r.t parameter $\phi_B^1$:
\begin{equation}
\begin{aligned}
& \mathbb{E}_{p_\mathcal{D}}\left[K L\left(q_{\phi_C}(\boldsymbol{z}_C \mid \boldsymbol{x}) \mid q_{{\phi}_{B}^1}(\boldsymbol{z}_C \mid \boldsymbol{y})\right)\right]
\end{aligned}
\end{equation}
and w.r.t parameter $\phi_B^2$:
\begin{equation}
\begin{aligned}
& \mathbb{E}_{p_\mathcal{D}}\left[K L\left(q_{\phi_N}(\boldsymbol{z}_N \mid \boldsymbol{x}) \mid q_{{\phi}_{B}^2}(\boldsymbol{z}_N \mid \boldsymbol{z}_C)\right)\right]
\end{aligned}
\end{equation}
{\color{black}{By decomposing the KL divergences as in \citet{hoffman2016elbo, vedantam2017generative}, we effectively minimize $K L\left(q_{\phi_C}^{\text{avg}}\left(\boldsymbol{z}_C \mid \boldsymbol{x}\right) \mid q_{{\phi}_{B}^1}(\boldsymbol{z}_C \mid \boldsymbol{y})\right)$ and $K L\left( q_{\phi_N}^{\text{avg}}\left(\boldsymbol{z}_N \mid \boldsymbol{x}\right) \mid q_{{\phi}_{B}^2}(\boldsymbol{z}_N \mid \boldsymbol{z}_C)\right)$, where $q_{{\phi}}^{\operatorname{avg}}(\boldsymbol{z} \mid \boldsymbol{x})=\mathbb{E}_{{p}(\boldsymbol{x} \mid \boldsymbol{y})}\left[q_{{\phi}}(\boldsymbol{z} \mid \boldsymbol{x})\right]$.}}
As there is little posterior uncertainty once conditioned on an EEG signal $\boldsymbol{x}_{i}$, the approximations are close to the average posterior induced by each of the EEG $\boldsymbol{x}_i$ associated with similar $\boldsymbol{y}$. 

Having fit both the generative and discriminative models, we can now explore the three-way relationship between behavior, brain activity, and cognitive processes.

\section{Experiments}

\subsection{EEG and Behavioral Dataset}

We used behavioral and EEG data collected while participants performed a two-alternative forced-choice task where they had to decide whether a Gabor patch presented with added dynamic noise is higher or lower spatial frequency \citep[for details, see Experiment 2 by][]{nunez2019latency}. Task difficulty was manipulated by adding spatial white noise to manipulate the quality of the perceptual evidence available to make the discrimination. The signal and the noise flickered at 30 and 40 Hz frequencies, respectively. 4 participants performed the task in blocks of trials at 3 added noise levels (low, medium, and high). Each subject performed approximately 3000 trials over 7 experimental sessions, while 128 channels of EEG and behavioral data were recorded. The independent component analysis (ICA)-based artifact rejection method was used on EEG data to remove eyeblinks, electrical noise, and muscle artifacts. A subset of 98 EEG channels were selected, excluding channels located in the outer ring. EEG data were bandpass filtered to 1 to 45 Hz in the frequency domain and then downsampled from 1000 Hz to 250 Hz in the time domain prior to data analysis. The data for each subject were divided into 80\% for training and validation and the remaining 20\% for testing.

\subsection{Results}

\begin{table}[h!]
    \centering
    \caption{Comparison of the sum of Wiener negative log-likelihood ($- \sum \log \text { Wiener }(\text{RT}_i \mid \boldsymbol{\omega}_i)$) of four subjects on the test sets.  $\boldsymbol{\bar\omega}$ represents the median fitted cognitive parameters from the training set.}
    \label{tab:llh}
    \begin{tabular}{crl}
        \hline
        Subjects & $\boldsymbol{\omega}_{i}^\text{test}$ & $\boldsymbol{\bar\omega}^{\text{train}}$ \\ 
        \hline
        s1 & $-0.018$ & $0.212$ \\
        s2 & $-0.244$ & $0.159$ \\
        s3 & $0.264$ & $0.735$ \\
        s4 & $0.031$ & $0.230$ \\
        \hline
    \end{tabular}
\end{table}

\begin{figure*}[!htb]
  \centering

  \begin{subfigure}{\linewidth}
    \centering
    \includegraphics[trim=0 0 0 0.7cm, clip, width=0.73\linewidth]{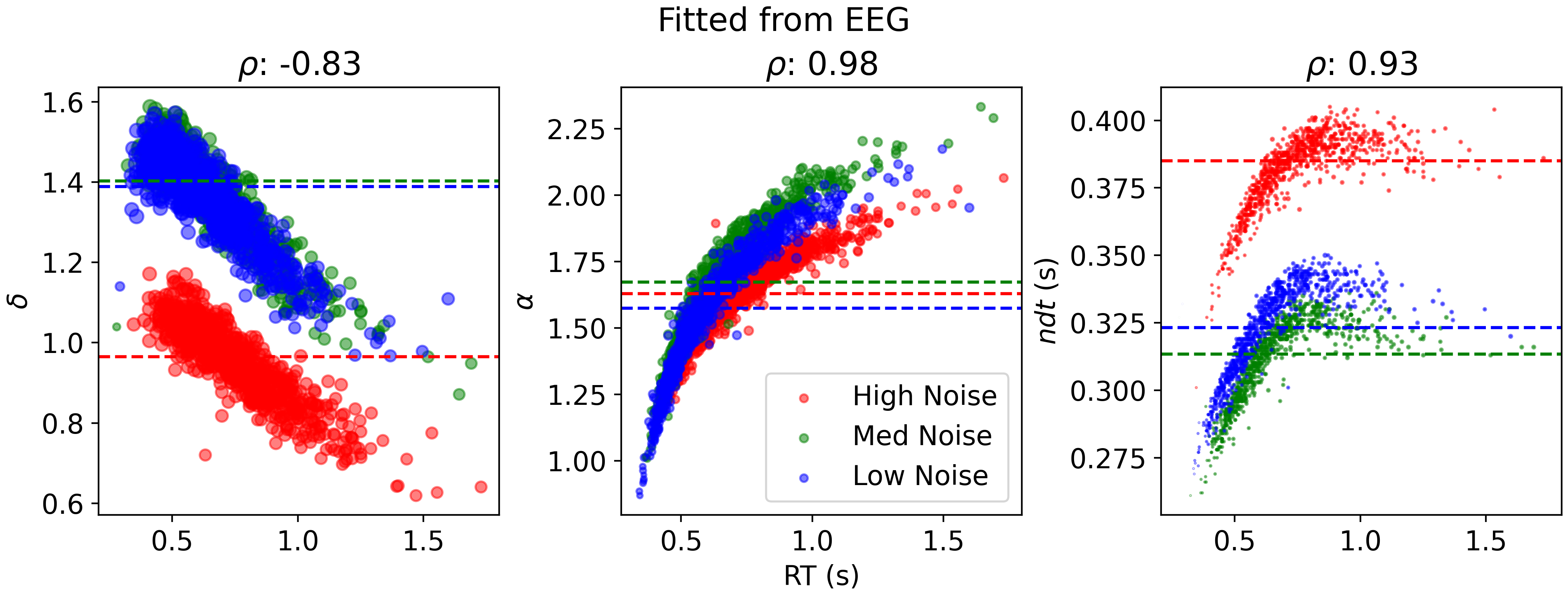}
    \vspace{-2mm}
    \subcaption{Fitted from EEG (training data)}
    \label{fig:ddm_params1}
  \end{subfigure}
  
  \vspace{0.4cm}
  
  \begin{subfigure}{\linewidth}
    \centering
    \includegraphics[trim=0 0 0 0.7cm, clip, width=0.73\linewidth]{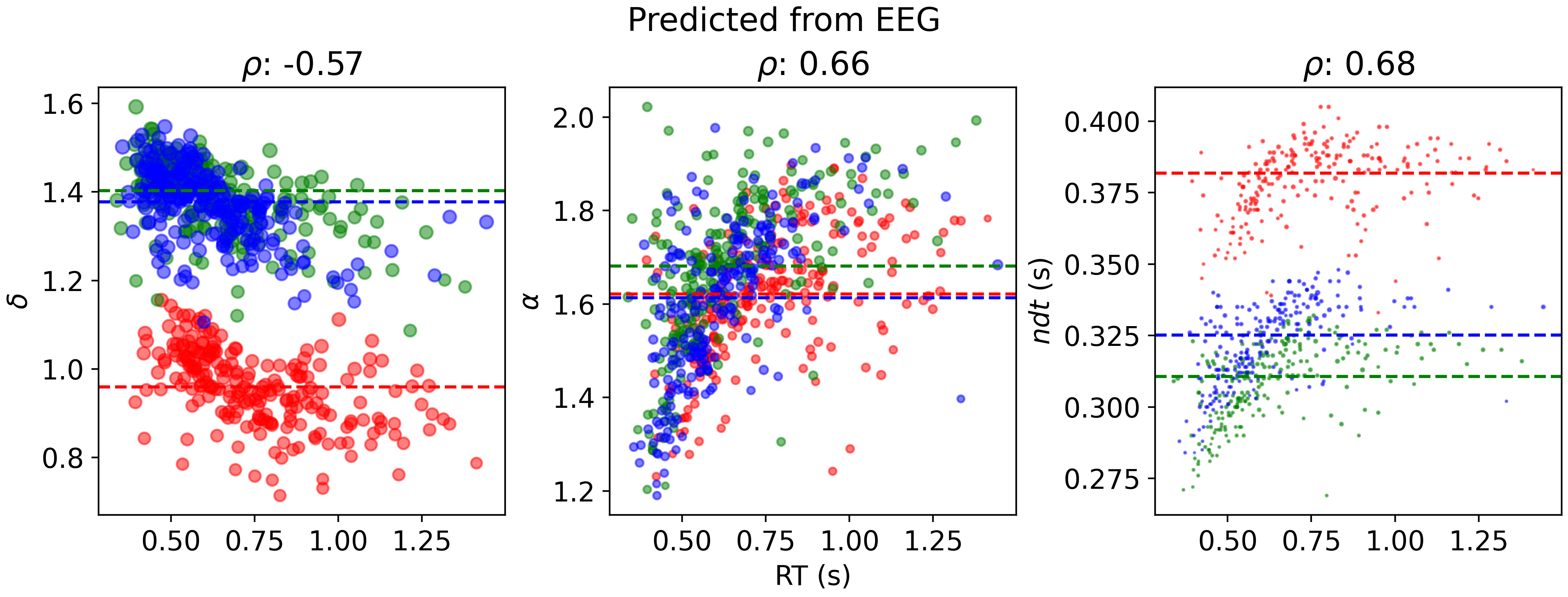}
    \vspace{-2mm}
    \subcaption{Predicted from EEG (test data)}
    \label{fig:ddm_params2}
  \end{subfigure}
    
  \vspace{0.4cm}
    
  \begin{subfigure}{\linewidth}
    \centering
    \includegraphics[trim=0 0 0 0.7cm, clip, width=0.73\linewidth]{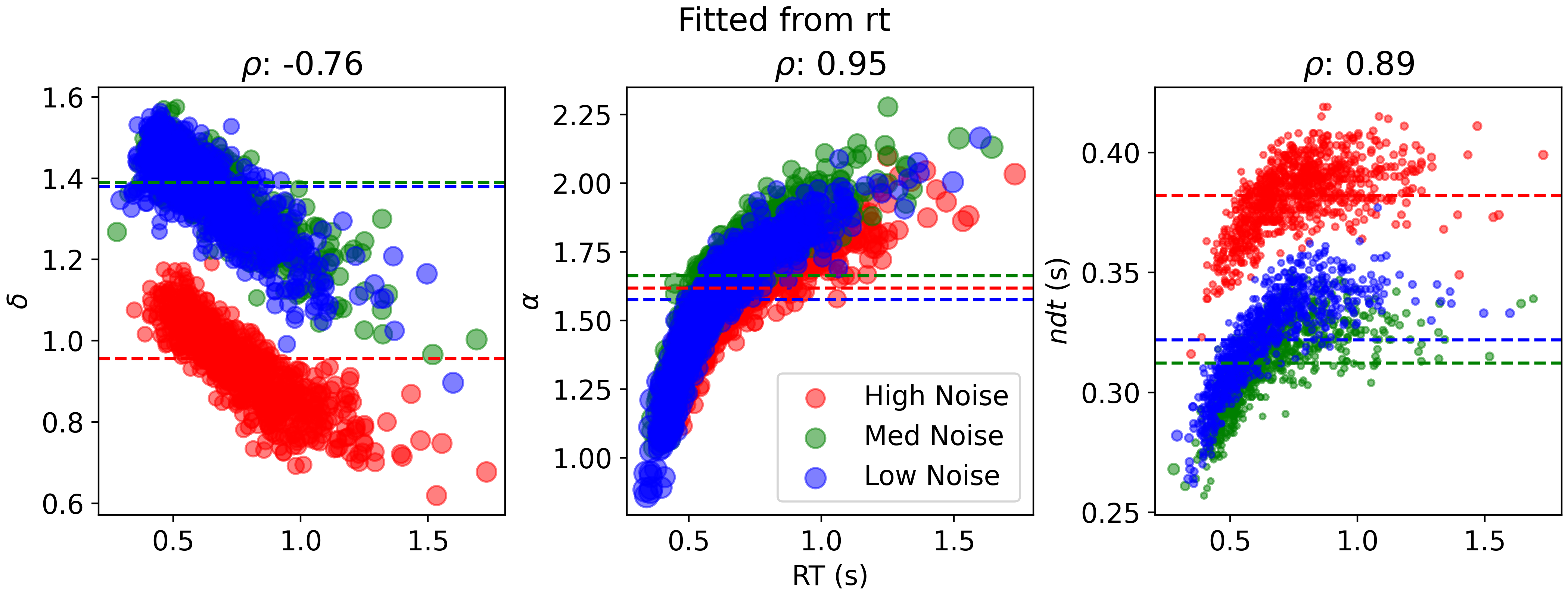}
    \vspace{-2mm}
    \subcaption{Fitted from choice-RTs (training data)}
    \label{fig:ddm_params3}
  \end{subfigure}

  \vspace{0.4cm}

  \begin{subfigure}{\linewidth}
    \centering
    \includegraphics[trim=0 0 0 0.7cm, clip, width=0.73\linewidth]{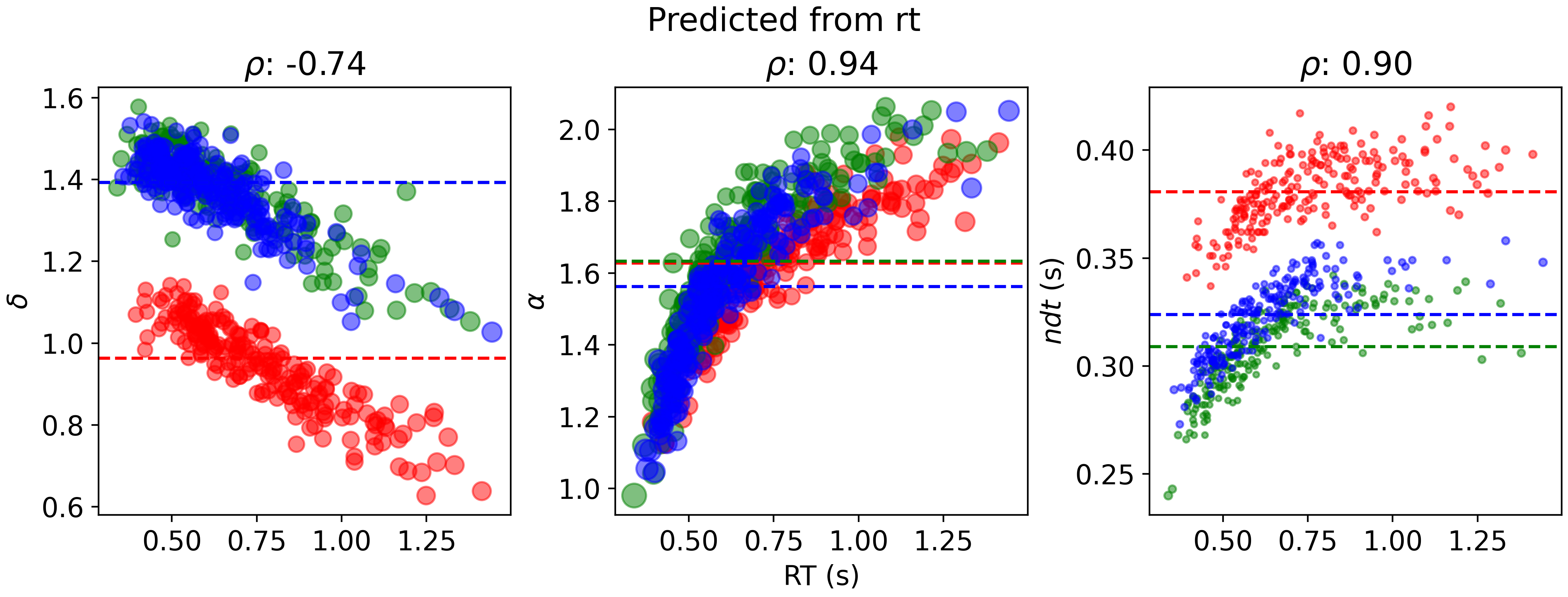}
    \vspace{-2mm}
    \subcaption{Predicted from choice-RTs (test data)}
    \label{fig:ddm_params4}
  \end{subfigure}

  \end{figure*}

To validate the neurocognitive modeling approach, we first examine the trial-by-trial variability of the parameters within each subject and the generalization of the model to unseen data. Figures \ref{fig:ddm_params1} and \ref{fig:ddm_params3} show the trial-by-trial correlations between estimated DDM posteriors and observed choice-RTs in the training data from neural signals and behavior, respectively. Spearman correlations between fitted drift rates ($\delta$) and choice-RTs are negatively strong. At the same time, there are strong positive correlations between boundaries ($\alpha$) and choice-RTs, as well as between non-decision time and choice-RTs. The estimates in NCVA are regularized by the subject priors obtained from a Bayesian hierarchical fitting of a DDM using MCMC \cite{nunez2019latency}. The model was individually fitted for each subject using choice-RT and accuracy only and accounted for between-condition variability within subjects. Clear clusters of drift rates and non-decision-time estimates depending on the noise conditions can be seen, though boundary estimates are highly overlapped. It is worth noting that uncertainties in the estimates can be inspected from the figures through the posterior covariance. Understandably, the uncertainties in the estimations from choice-RTs are significantly higher than from EEG signals, which agree with the theoretical derivations in Section \ref{disc}. Figures \ref{fig:ddm_params2} and \ref{fig:ddm_params4} also demonstrate a satisfactory generalization to unseen data. The drift rates positively correlate with choice-RTs, whereas the boundaries and non-decision time negatively correlate with choice-RTs. The model successfully learns to extract the neural features that account for the choice-RT variability at each trial. 
To evaluate whether obtaining trial estimates of cognitive parameters improved the model of choice and choice-RT data, Table \ref{tab:llh} presents the Wiener likelihood test for the neurocognitive generalization ability to unseen data. The results show that the use of single-trial predictions of cognitive parameters $\boldsymbol{\omega}_{i}$ provides higher likelihood than the median estimates $\boldsymbol{\bar\omega}$ fitted from the training data. This implies that single-trial estimates better account for new data compared to median estimates. 

\begin{figure*}[!htb]
  \centering

  \begin{subfigure}{\linewidth}
    \centering
    \includegraphics[width=0.9\linewidth]{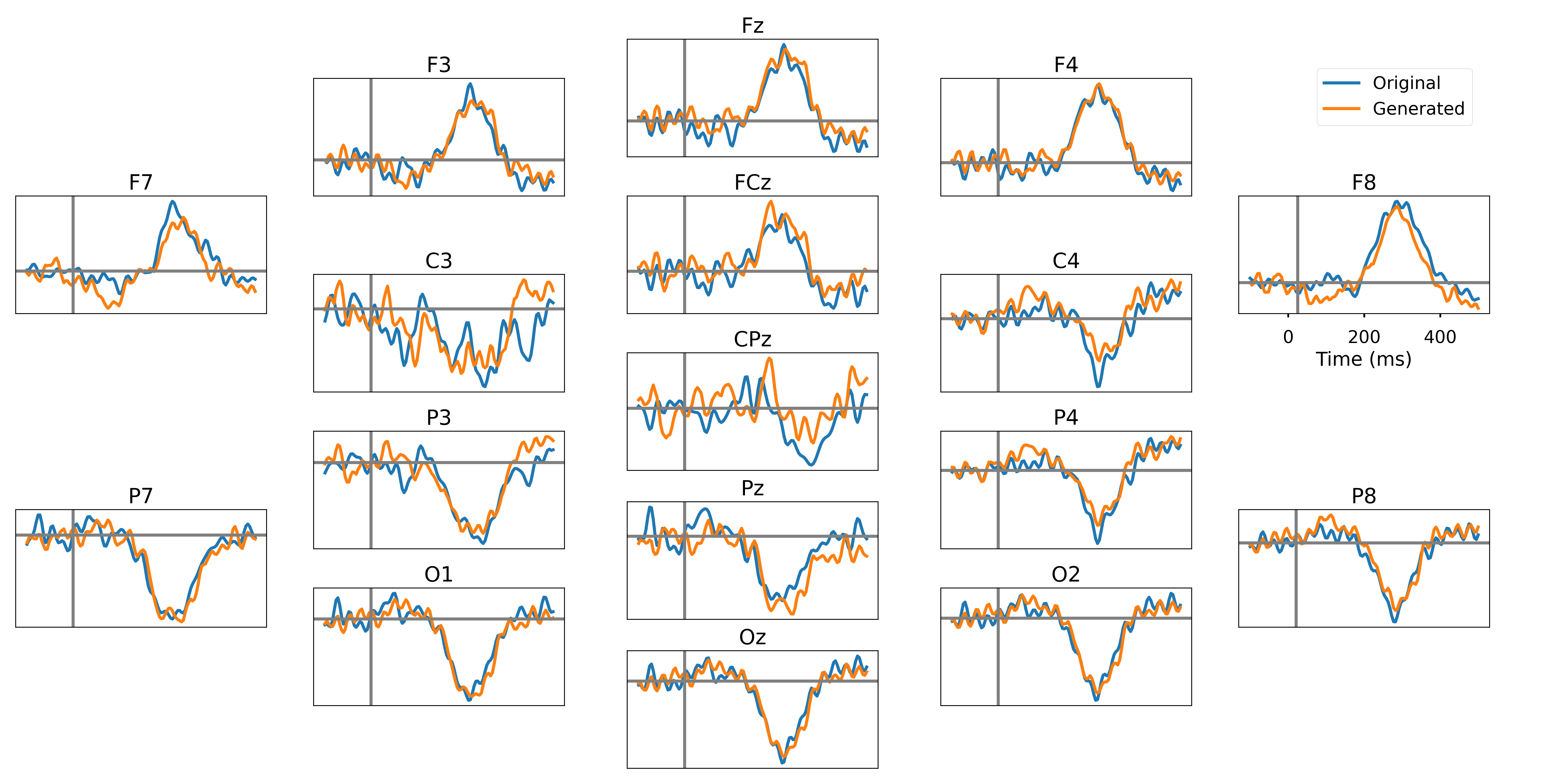}
    \vspace{-2mm}
    \subcaption{EEG data at the selected electrodes}
    \label{fig:trial_avg_erp1}
  \end{subfigure}

  \vspace{0.2cm}
  
  \begin{subfigure}{\linewidth}
    \centering
    \includegraphics[width=0.9\linewidth]{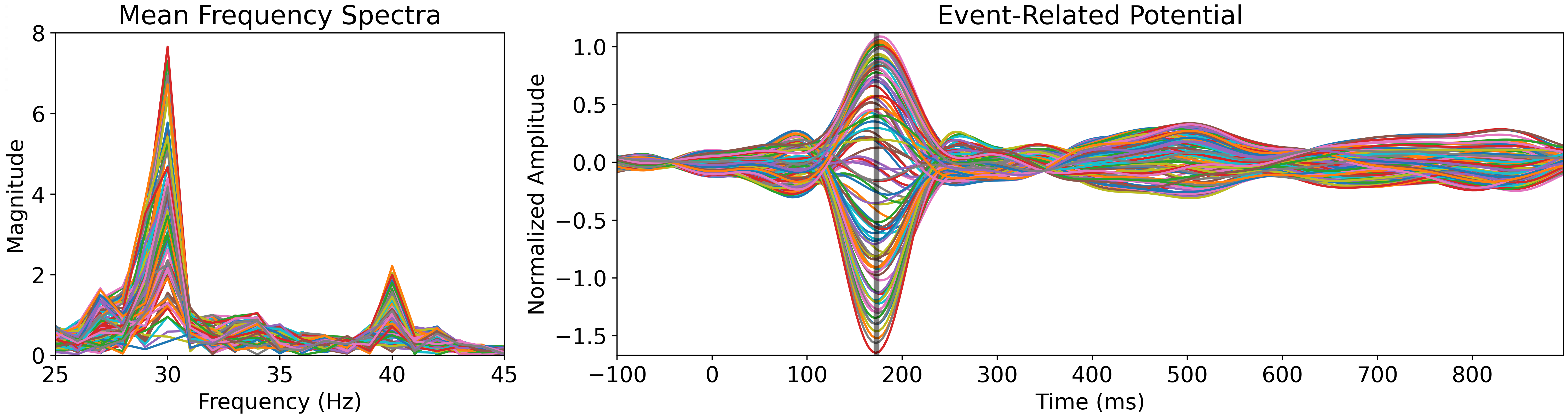}
    \vspace{-2mm}
    \subcaption{Low noise condition}
    \label{fig:trial_avg_erp2}
  \end{subfigure}

  \vspace{0.2cm}

  \begin{subfigure}{\linewidth}
    \centering
    \includegraphics[width=0.9\linewidth]{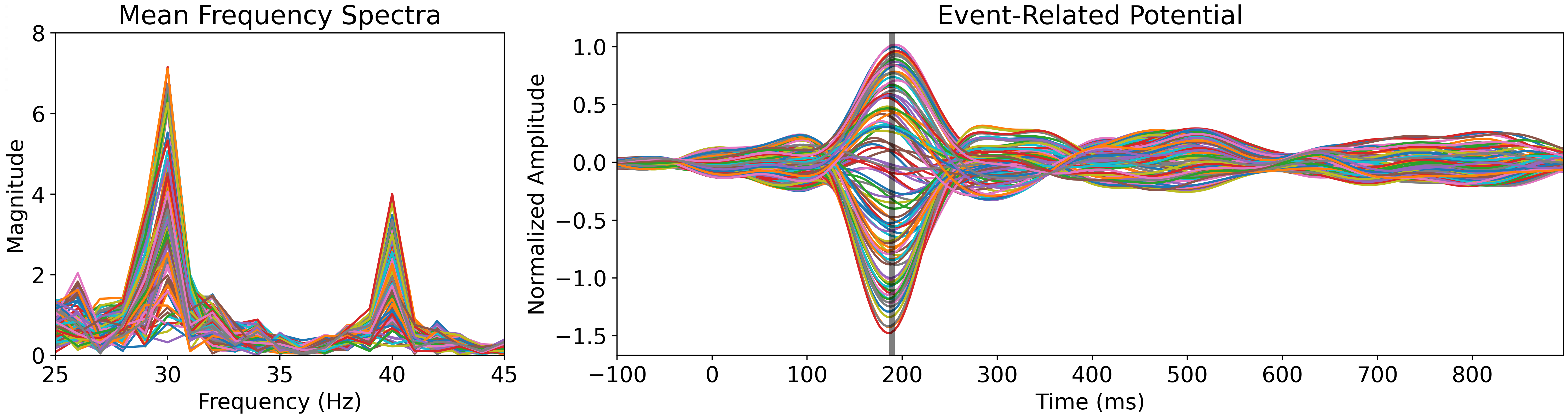}
    \vspace{-2mm}
    \subcaption{Medium noise condition}
    \label{fig:trial_avg_erp3}
  \end{subfigure}

  \vspace{0.2cm}

  \begin{subfigure}{\linewidth}
    \centering
    \includegraphics[width=0.9\linewidth]{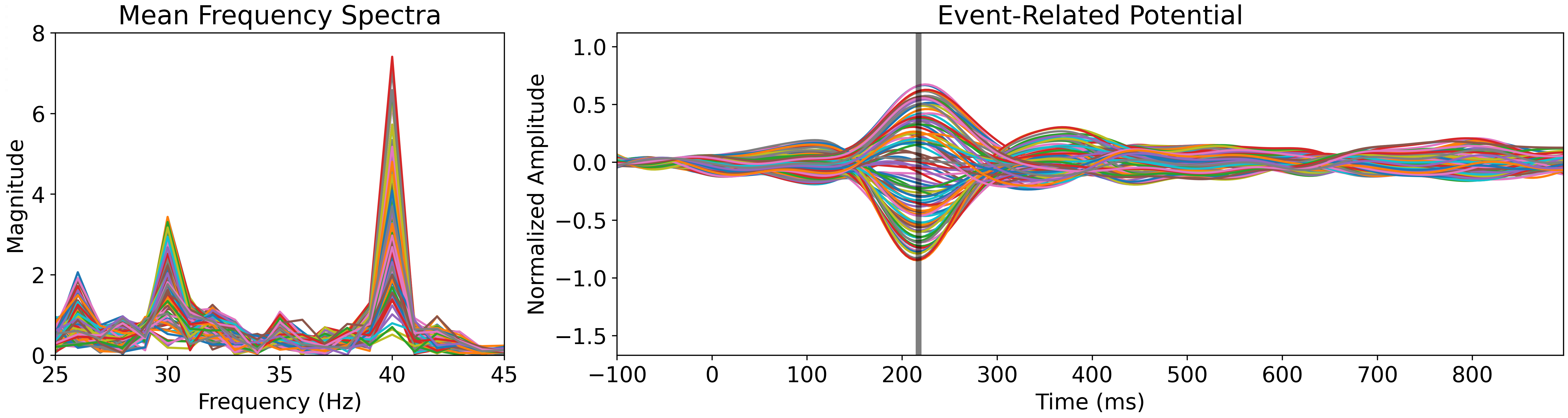}
    \vspace{-2mm}
    \subcaption{High noise condition}
    \label{fig:trial_avg_erp4}
  \end{subfigure}
  \end{figure*}

Figure \ref{fig:trial_avg_erp1} shows the average of signals generated by the neurocognitive autoencoder when given a set of approximately 800 test choice-RTs compared to the average of actual signals associated with the same choice-RTs. At the selected electrodes, the window of interest is 100 ms pre-stimulus to 500 ms post-stimulus, which captures the N200 waveform. The generated and original signals appear visually similar in the timing and amplitudes of the peaks and troughs. 
Figures \ref{fig:trial_avg_erp2}, \ref{fig:trial_avg_erp3}, and \ref{fig:trial_avg_erp4} depict the trial-averaged frequency spectra and corresponding ERP waveforms of the reconstructed signals. Regarding the frequency spectra, the most important features are the 30 and 40 Hz peaks, which correspond to the flicker frequency of the signal (Gabor patch) and spatial white noise, respectively. Interestingly, the generative model learns to structure output the steady-state visually evoked potentials (SSVEPs) that occur in response to a visual stimulus flickering at different frequencies, even though it was never explicitly encoded in the model. Moreover, in the low noise condition (b), the 30 Hz peak is large and the 40 Hz is small, while in the high noise condition (d), the 30 Hz peak is reduced and the 40 Hz peak is enhanced.
In terms of ERP waveforms, the model captures the relationships of the N200 peak latencies with respect to the additive noise conditions. Higher additive white noise in the stimulus effectively increases the latency and decreases the amplitude of the N200. We focus on the N200 signal because the original study \citep{nunez2019latency} found strong relationships between N200 latency and choice-RT, and thus the N200 is a good validation of our model.
These prove the convergence of the model in optimizing the lower bound of the conditional likelihood mapping from behavioral data to EEG features, which effectively encodes differences in the stimuli presented to the subjects in the latent variable space.

\begin{figure*}[!htb]
  \centering
  
  \begin{subfigure}{\linewidth}
    \centering
    \includegraphics[width=0.7\linewidth]{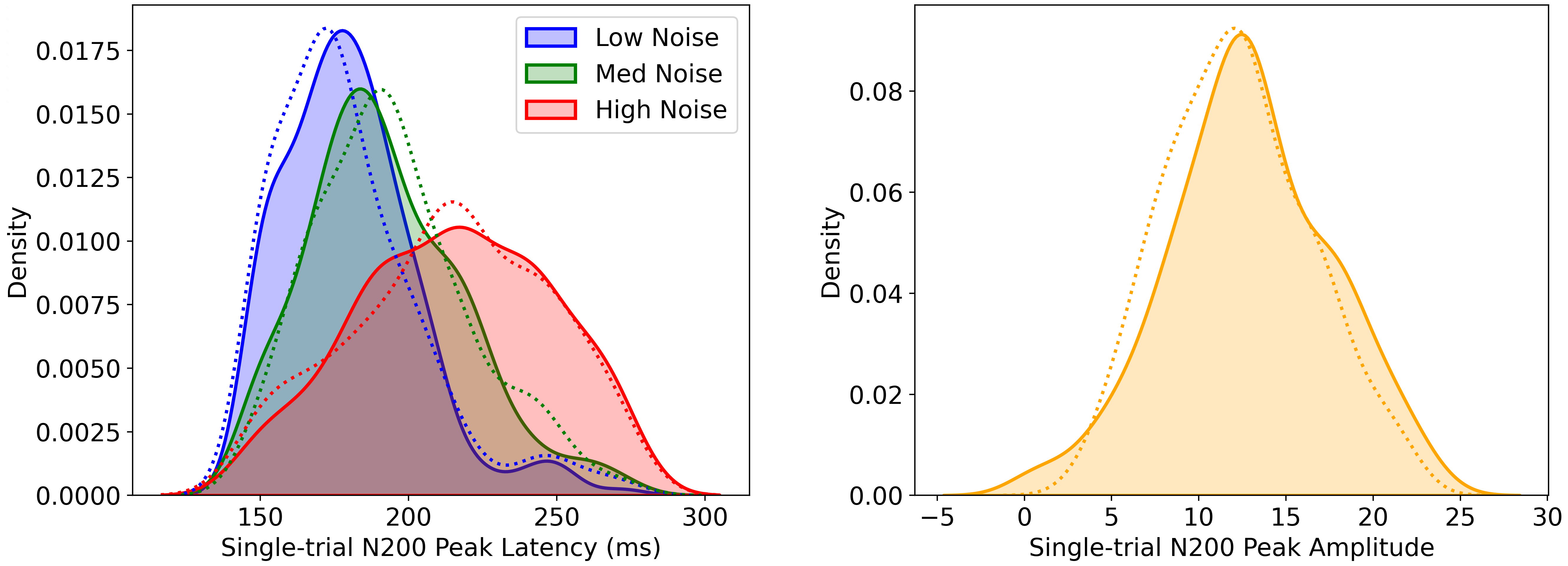}
    \vspace{-2mm}
    \subcaption{Subject s1}
    \label{fig:n2001}
  \end{subfigure}

  \vspace{0.7cm}

  \begin{subfigure}{\linewidth}
    \centering
    \includegraphics[width=0.7\linewidth]{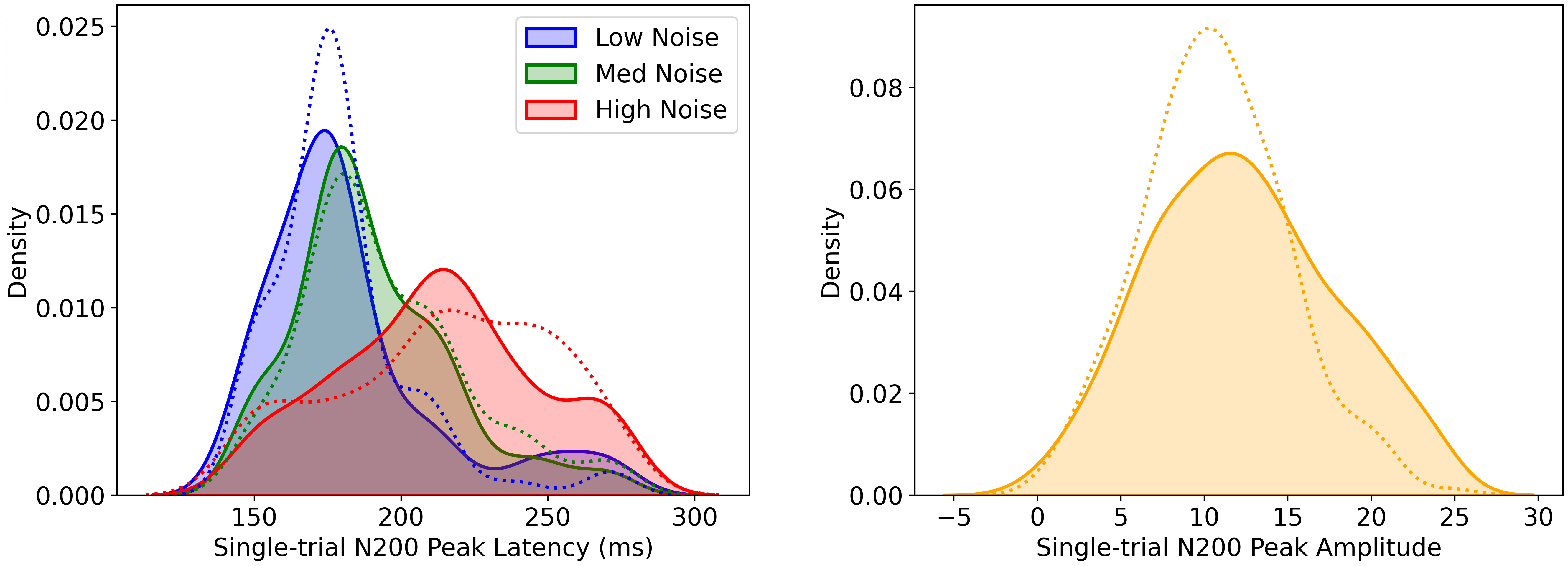}
    \vspace{-2mm}
    \subcaption{Subject s2}
    \label{fig:n2002}
  \end{subfigure}

  \vspace{0.7cm}

  \begin{subfigure}{\linewidth}
    \centering
    \includegraphics[width=0.7\linewidth]{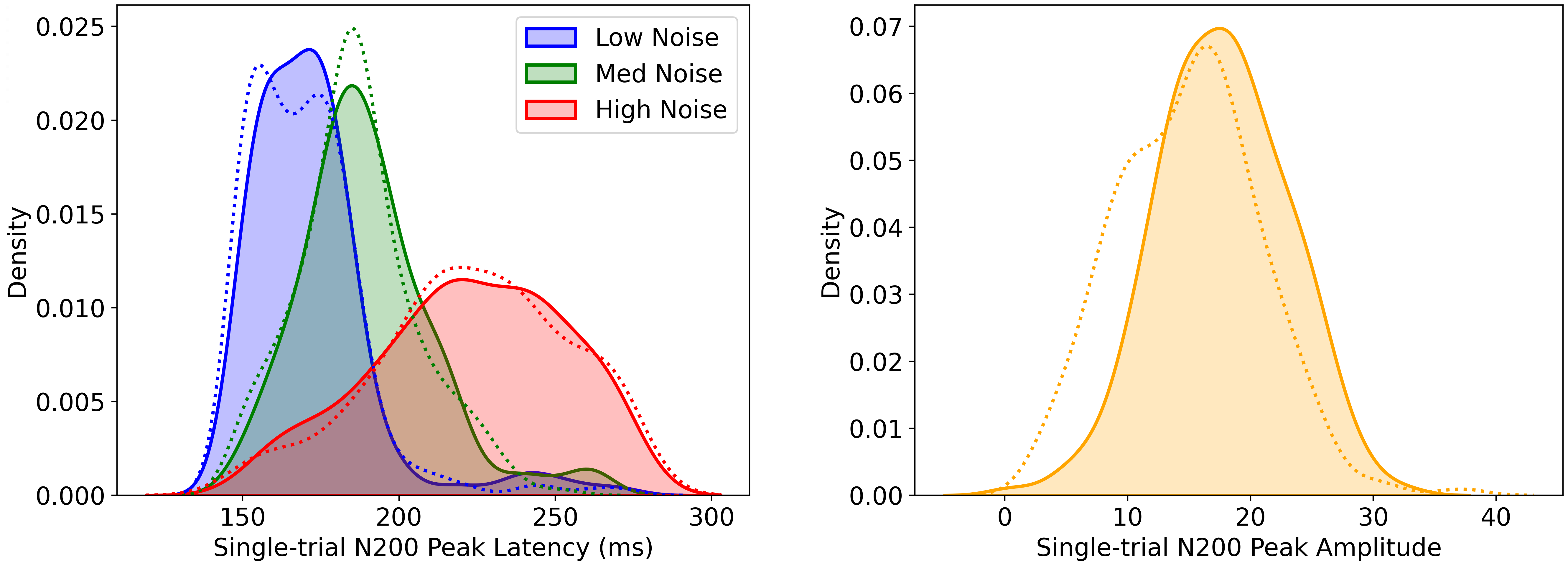}
    \vspace{-2mm}
    \subcaption{Subject s3}
    \label{fig:n2003}
  \end{subfigure}

  \vspace{0.7cm}

  \begin{subfigure}{\linewidth}
    \centering
    \includegraphics[width=0.7\linewidth]{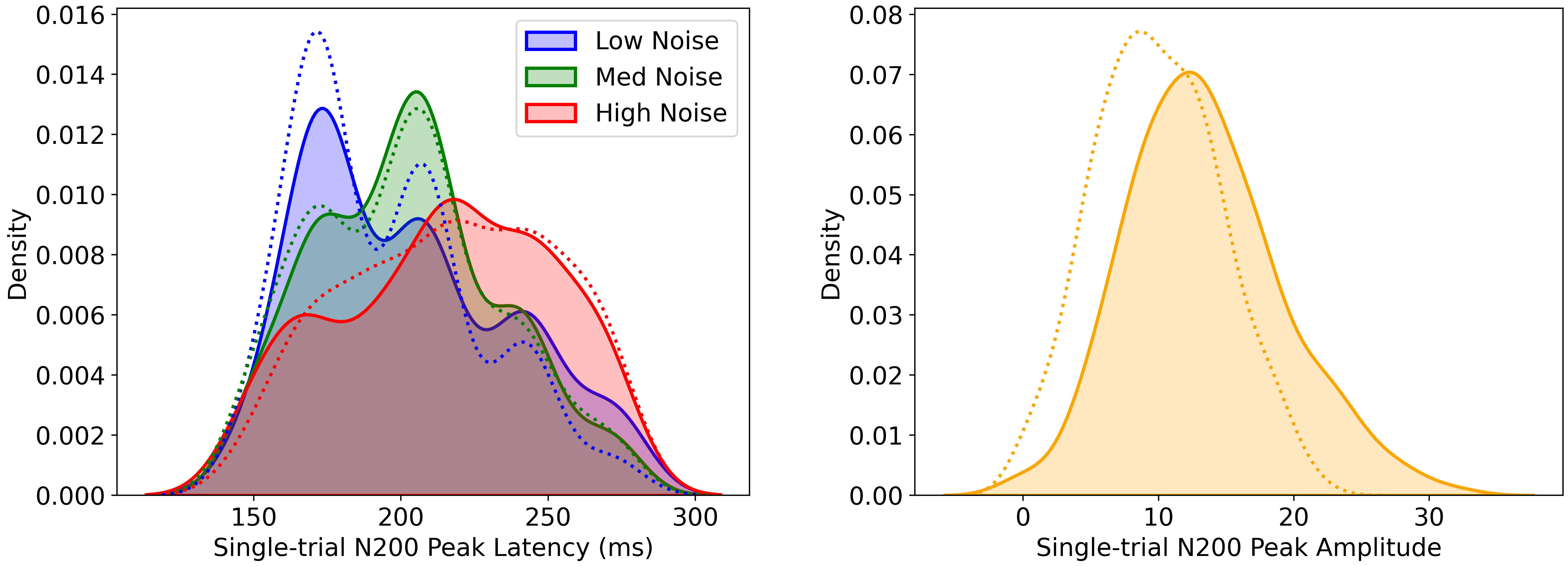}
    \vspace{-2mm}
    \subcaption{Subject s4}
    \label{fig:n2004}
  \end{subfigure}

  \caption{Performance of the model in reconstructing single-trial N200 peaks from choice-RTs in four subjects. The dotted lines are references to the original data. The distributions of (left) single-trial N200 peak latencies across three noise conditions and (right) the N200 peak amplitude statistics are shown. Single-trial observations of the peak latency of N200 are found using the SVD method \citep{nunez2019latency} for each subject and noise condition.}
  \label{fig:n200}
\end{figure*}

In addition to evaluating traditional ERP estimates (trial-averaged), we also assess the single-trial ERP estimate (channel-averaged). To increase the signal-to-noise ratio to better detect the N200, the first singular-value decomposition (SVD) component obtained from the ERP response is taken as a channel weighting function. More details of the SVD method can be seen at \citep{nunez2019latency}. Figure \ref{fig:n200} shows the performance of the model in learning the N200 feature in each trial.
As shown in Figure \ref{fig:n200}, the distributions of the single-trial N200 peak latencies, as well as the amplitudes calculated from the generated signals, closely match those of the original signals at three different noise levels. The peak amplitude distribution is somewhat broader than the original data's generated distribution. Importantly, the model can generate the variability of the N200 latency with the experimental manipulation of low, medium, and high noise, systematically increasing the N200 latency in the generated signals. 

Figure \ref{fig:sensitivity} represents the sensitivity analysis of the choice-RT and drift-diffusion parameters regardless of the noise conditions. In the left column, we examine the sensitivity of the neural signals generated by the choice-RTs.  We can see similar patterns across subjects where the increases in choice-RTs lead to significant declines in the 30 Hz and the rises of the N200 latencies. This confirms the minimization approach of the KL divergence between the latent spaces inferred from the behavioral data and the neural signals. Power at 40 Hz reflecting the neural response to the noise also changes according to the choice-RTs, though the pattern is not as strong as the subjects suppressed the noise signal in all conditions.  

One of the powerful tools for exploring the relationship between cognitive processes is to examine the sensitivity of neural signals to cognitive parameters. The middle and right columns of Figure \ref{fig:sensitivity} depict the effect of \emph{hypothetical} modulations of drift rates and non-decision time on the generated neural signals. The results show that our model reveals the intricate interactions between cognitive parameters and neural signals, which is consistent with prior discoveries in the cognitive modeling literature. As the non-decision time is faster, the N200 latencies are shorter, and the 30 Hz peaks are larger. Accordingly, the amplitudes of the N200 peaks are more prominent, though not shown in the figures for clarity. The same interactions are observed with the increase in drift rates, representing evidence accumulation. Again, the effects on 40 Hz peaks are weaker and depend on the subjects. We did not observe the effects of the boundary separation (caution) on the neural signals. The effect can be reversed with slower non-decision times and lower drift rates. The strongest effects can be seen when both parameters influence neural signals. This demonstrates the effectiveness of the designs of the hierarchical latent variables inferred from choice-RTs and the disentangled latent space produced by the EEG data. 

  \clearpage
  \captionof{figure}{Drift-diffusion single-trial parameter estimations from correct responses of subject s1. The parameters are constrained by the subject priors resulting from a Bayesian MCMC modeling (without EEG data). Scatter plots illustrate the relationship between the parameters and the observed choice-RTs for each trial. The top two rows are posterior inferences from neural signals, while the bottom two are from behaviors. The left column shows the drift-rate ($\delta$) estimates, the middle column shows boundary ($\alpha$) estimates, and the right column presents non-decision time ($ndt$) estimates. The correlations between the choice-RTs and the inferred DDM parameters are consistent with what is expected. On top of each panel are the Spearman correlation coefficients ($\rho$). The covariances of the inferred parameters are indicated by circles, which correspond to contours having one standard deviation. For clarity, each circle is magnified 300 times.}
  \label{fig:ddm_params}
  
  \clearpage
  \captionof{figure}{Performance of the model in reconstructing 98 EEG channels of subject s1 by averaging $\approx$ 800 predicted EEG trials from $\approx$ 800 choice-RTs in the test set. Time point zero denotes the time point of stimulus onset. The first row displays the original (blue) and generated (orange) trial-averaged EEG data at the pooled electrodes. The x-axis denotes the time in milliseconds from stimulus onset, and the y-axis denotes the signal amplitude. The second, third, and fourth rows are (left) frequency spectra and (right) EEG signals averaged over all test choice-RT trials ($\approx 800/3$ per condition). The signals on the right are low-pass filtered at 15 Hz for clarity of N200 peaks. Each colored line corresponds to one reconstructed EEG channel. In low-noise conditions, the spectra show a strong peak at the Gabor flicker frequency of 30 Hz, and the ERP waveform shows a shorter N200 latency and larger peak amplitude. Under high-noise conditions, the spectra show a strong peak at the noise flicker frequency of 40 Hz, and the ERP waveform shows a longer N200 latency and a smaller peak amplitude.}
  \label{fig:trial_avg_erp}

\begin{figure*}[!htb]
  \centering

  \begin{subfigure}{\linewidth}
    \centering
    \includegraphics[width=1\linewidth]{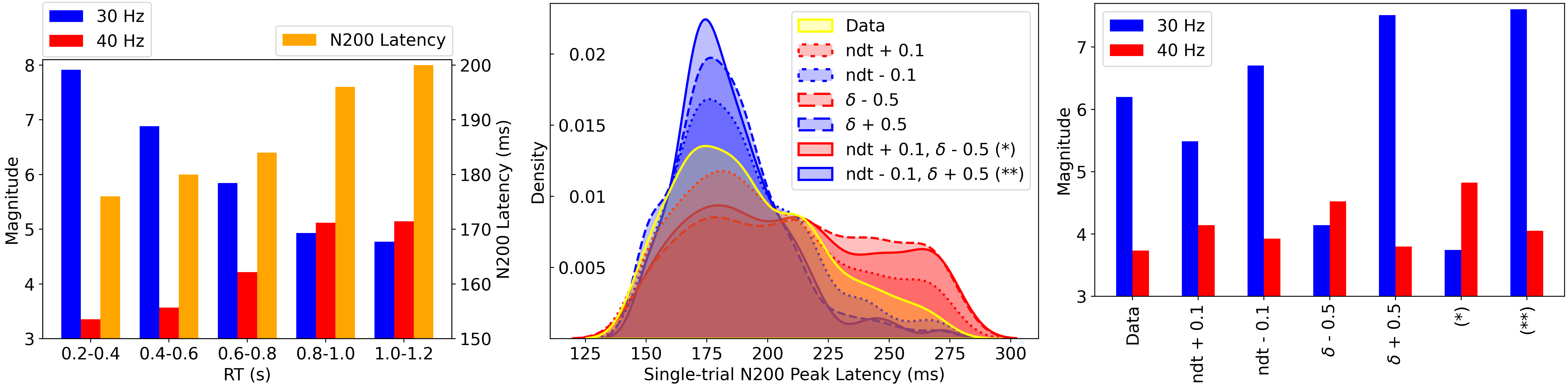}
    \vspace{-5mm}
    \subcaption{Subject s1}
    \label{fig:sensitivity1}
  \end{subfigure}

  \vspace{0.5cm}
  
  \begin{subfigure}{\linewidth}
    \centering
    \includegraphics[width=1\linewidth]{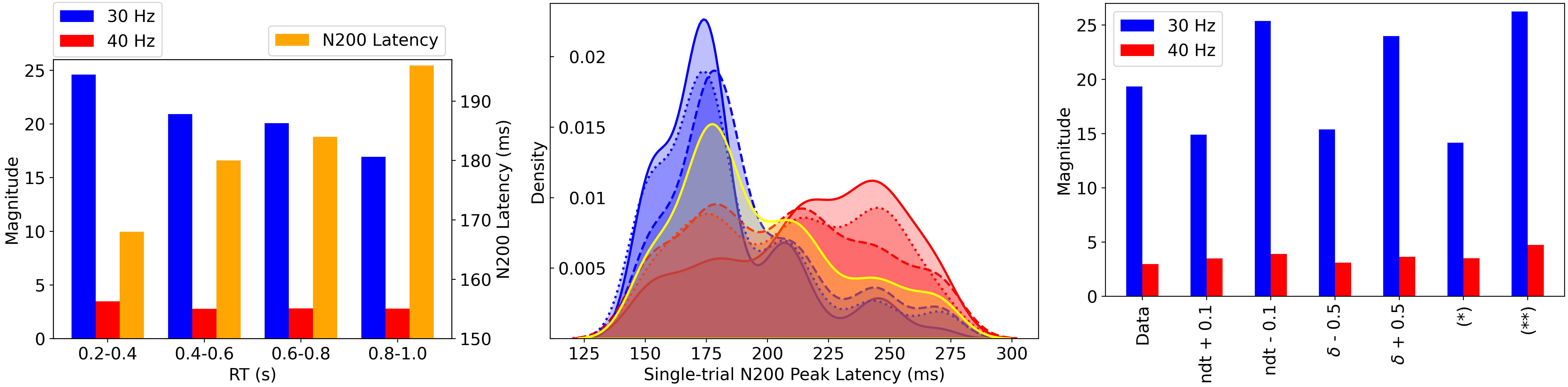}
    \vspace{-5mm}
    \subcaption{Subject s2}
    \label{fig:sensitivity2}
  \end{subfigure}

  \vspace{0.5cm}
  
  \begin{subfigure}{\linewidth}
    \centering
    \includegraphics[width=1\linewidth]{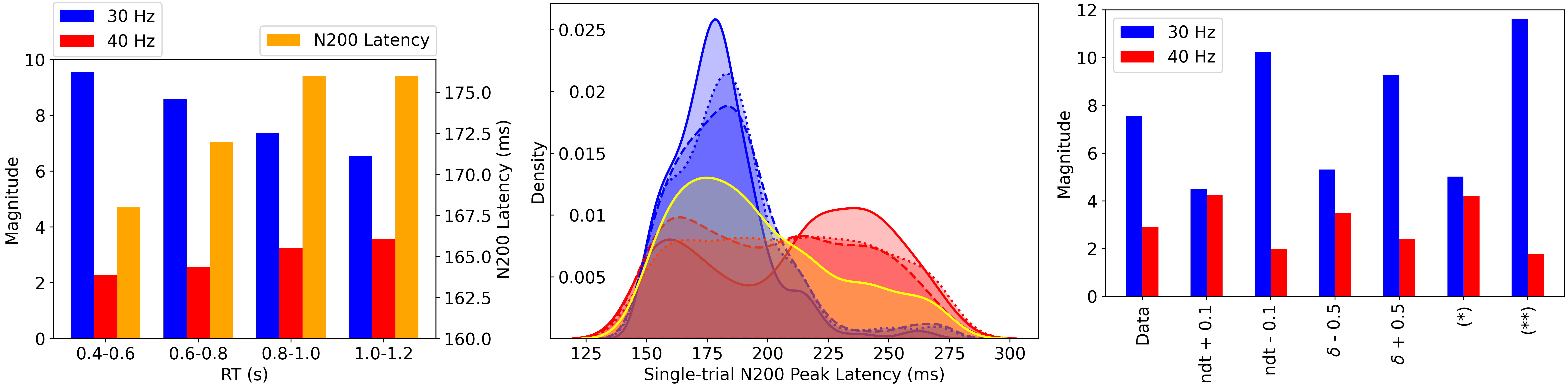}
    \vspace{-5mm}
    \subcaption{Subject s3}
    \label{fig:sensitivity3}
  \end{subfigure}

  \vspace{0.5cm}

  \begin{subfigure}{\linewidth}
    \centering
    \includegraphics[width=1\linewidth]{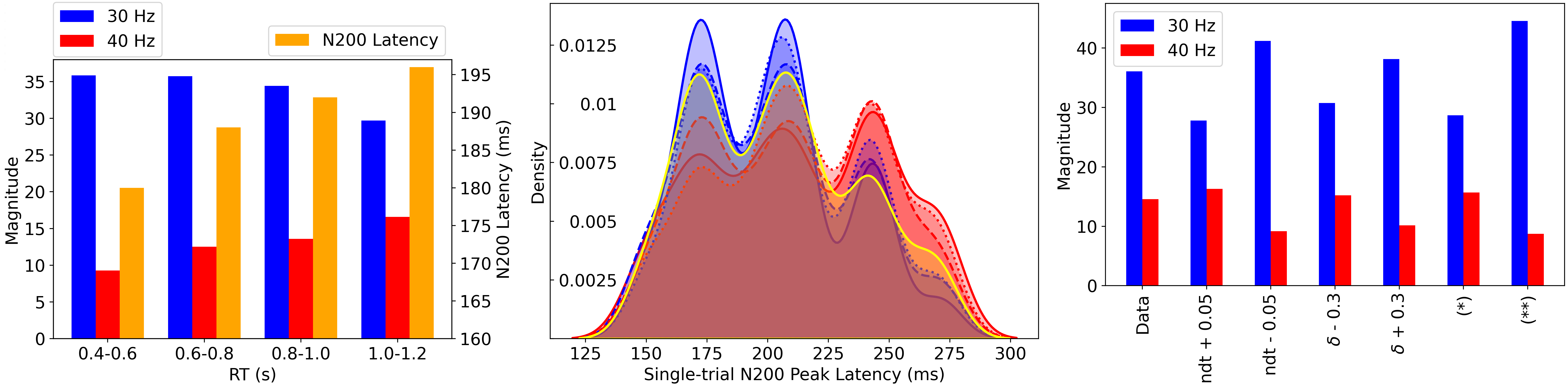}
    \vspace{-5mm}
    \subcaption{Subject s4}
    \label{fig:sensitivity4}
  \end{subfigure}
  \end{figure*}
  \clearpage
  \captionof{figure}{Sensitivity analysis of choice-RTs and latent drift-diffusion parameters on EEG signal generation in four subjects. The left column presents the effects of choice-RTs on the output neural signals. The blue bars represent the power at 30 Hz, while the red bars represent the power at 40 Hz. The orange bars show the N200 latencies. The middle column shows the changes in the single-trial N200 distribution w.r.t to \emph{hypothetical} changes in the cognitive parameters. The yellow distribution represents the reference data, while the blue and red ones correspond to modified parameter settings that decrease or increase the N200 latencies, respectively. The modification in subject s4 (ndt $\pm$ 0.05, $\delta \pm$ 0.3) is different from other subjects. The right column characterizes the changes in 30 Hz and 40 Hz peaks w.r.t to the changes in the same cognitive parameters.}
  \label{fig:sensitivity}

\section{Conclusion}
In this work, we proposed a joint behavioral and EEG modeling approach driven by a cognitive model of decision making. 
The experimental results demonstrate the effectiveness of our Neurocognitive VAE in simultaneously modeling high-dimensional EEG signals and low-dimensional behavioral data.
Remarkably, the model learns essential task-relevant neural features, e.g. N200 peaks and SSVEP, without explicit specification in the optimization objective. 
Furthermore, the model captures how these features modulate behavior, specifically discovering relationships between brain activity and behavior consistent with other models based on prior knowledge.
This suggests that the Neurocognitive VAE helps uncover neural signals linked to behavioral data by mapping to a structured latent space.
Compared to the aforementioned published joint models \citep{nunez2015individual, nunez2017attention, nunez2019latency, lui2021timing, turner2013bayesian, turner2016more}, our end-to-end model is capable of inferring task-relevant EEG features from behavior without prior knowledge of which features to optimize.
The structured latent space allows the learning of behavioral variability to drive the EEG data generation process, leading to the prediction of the structure of EEG features in relation to the stimuli used in the experiments (N200 and SSVEP) and the behavioral performance (choice-RT). In addition, the model allows us to directly map the variability of cognitive parameters to neural signals, allowing for theoretical predictions that guide future experimental studies. 
It should be noted that our framework does not serve to refine the functional form of process-oriented computational models. Instead, it presumes a set of fixed assumptions; in the DDM, a constant drift rate and boundary separation within trials. 
Importantly, our framework can be generalized to encompass any other neural measures combined with any cognitive model to explain behavior, provided that the cognitive model expresses a closed-form likelihood of behavioral data. Importantly, by parameterizing the likelihood by a deep neural network receiving neural data as input, trial-level parameter inferences are made possible. 
In this research, we assume a DDM posterior with a diagonal covariance matrix. This could lead to an overestimation of the variance of the marginal posteriors if the true posterior has dependencies. It would be beneficial to investigate the use of a full covariance matrix as an alternative.
{\color{black}{It is important to mention that our validation process focused on correct responses. Due to the low number of incorrect responses compared to correct ones, we lack confidence in interpreting the results in this study for the incorrect trials, although the direction of the trial-level parameter fits was consistent with the results for correct trials. We anticipate future research to explore strategies to address the class imbalance problem in deep learning models \citep{johnson2019survey}.}}
Further work with a larger dataset is needed to demonstrate that we can extend the model to new individuals. In principle, this would potentially allow us to predict brain activity in clinical populations with known behavioral differences.

\section*{Data and Code Availability Statement}
The dataset analyzed during the current study is available on \url{https://zenodo.org/record/8381751}, and the implementation of the model is in the following repository \url{https://github.com/khuongav/neurocognitive_vae}.

\section{Supplementary Materials}

\subsection{Neural Network Architectures and Training Hyperparameters}

The inferential and generative processes are parameterized by deep neural networks, as shown by the flows in Figure \ref{fig:architecture}. Table \ref{tab:architecture} details the architectures of the five networks. 
The input EEG signals are of size 98 x 250 (1 second of data of 98 channels at 250 Hz). The feature extraction layers in the EEG and cognitive encoders are similar to \citet{vo2022composing}. All the feature maps have 128 channels. Leaky ReLU (lReLU) activation functions are applied to all layers, with a slope of 0.1 to stimulate easier gradient flow. Batch normalizations (BN) \citep{ioffe2015batch} are used in each convolutional layer of the encoders and decoders. Self-attention layers \citep{zhang2019self} are applied in the encoders and decoders to better account for long-range relationships in time series. $\boldsymbol{c}$ are noise condition embeddings as one-hot vectors (size 3). The size of $\boldsymbol{z}_N$ is set at 32 as increasing the dimension did not lead to any improvement in performance on a validation set. 

In Equation 5.4, the term $\log p(\boldsymbol{y} \mid \boldsymbol{z}_C)$ is weighted by $\lambda=2$ to scale up the likelihood of low-dimensional behavior. The KL terms are weighted by $\beta=20$. The KL terms are normalized to balance the KL divergence loss and the reconstruction loss. Please refer to Sections 4.2 and A6 of \citep{higgins2016beta} for further information. The optimization of $q_{\phi_C}\left(\boldsymbol{z}_C \mid \boldsymbol{x}\right)$ is divided into two stages. We first optimize the network w.r.t drift rate $\delta$ and boundary $\alpha$, while non-decision time $\tau$ is set to $0.93\cdot RT_{min}$ for each subject, approximating the results of the Bayesian MCMC modeling \cite{nunez2019latency}. Having trained $\phi_C$ for $\delta$ and $\alpha$, we can proceed to train only the last fully connected layer that predicts $\tau$. 
This procedure is to circumvent the difficulty of simultaneously optimizing the network for the boundary and the non-decision time on the experimental data. We used Adam \citep{kingma2014adam} for optimizations, with a learning rate of 5e-4 and exponential decay rates $\beta_1$ = 0.9 and $\beta_2$ = 0.999.

\begin{table*}
\scriptsize
\caption{Neural network parametrization}
\label{tab:architecture}
\hskip-2.0cm\begin{tabular}{clcc}
\hline
\multicolumn{1}{c|}{\textbf{$\text{Encoder}_\text{N} - q_{\phi_N}(\boldsymbol{z}_N \mid \boldsymbol{x})$}} &
  \multicolumn{1}{l|}{} &
  \multicolumn{1}{c|}{\textbf{$\text{Encoder}_\text{C} - q_{\phi_C}(\boldsymbol{z}_C \mid \boldsymbol{x})$}} &
  \textbf{Decoder - $p_{\theta}(\boldsymbol{x} \mid \boldsymbol{z}_N, \boldsymbol{z}_C)$} \\ 

\multicolumn{1}{c|}{\textbf{maps EEG signals to neural latents}} &
  \multicolumn{1}{l|}{} &
  \multicolumn{1}{c|}{\textbf{maps EEG signals to cognitive latents}} &
  \textbf{reconstructs EEG signals} \\ \hline
  
\multicolumn{1}{c|}{Dropout(0.3)} &
  \multicolumn{1}{l|}{} &
  \multicolumn{1}{l|}{} &
  Get $\boldsymbol{z}_C$ \\
\multicolumn{1}{c|}{Conv 1, lReLU, 128 x 250} &
  \multicolumn{1}{l|}{} &
  \multicolumn{1}{c|}{Conv 1, lReLU, 128 x 250} &
  Linear 128, lReLU \\ \cline{1-2}
\multicolumn{1}{c|}{Conv 6, BN, lReLU} &
  \multicolumn{1}{l|}{\multirow{2}{*}{X 2}} &
  \multicolumn{1}{c|}{Conv 6, BN, lReLU, Dropout(0.7)} &
  Linear 32, lReLU \\
\multicolumn{1}{c|}{Conv 6, Stride 2, BN, lReLU} &
  \multicolumn{1}{l|}{} &
  \multicolumn{1}{c|}{Conv 6, Stride 2, BN, lReLU, Dropout(0.7)} &
  Concat $\boldsymbol{z}_N$, $\boldsymbol{c}$ \\ \cline{1-3}
\multicolumn{1}{c|}{Self Attention} &
  \multicolumn{1}{l|}{} &
  \multicolumn{1}{c|}{Self Attention} &
  Conv Transp 8, Stride 4, 512 Channels, BN, lReLU \\ \cline{1-3}
\multicolumn{1}{c|}{Conv 6, BN, lReLU} &
  \multicolumn{1}{l|}{\multirow{2}{*}{X 2}} &
  \multicolumn{1}{c|}{Conv 6, BN, lReLU, Dropout(0.7)} &
  Conv Transp 8, Stride 4, 256 Channels, BN, lReLU \\
\multicolumn{1}{c|}{Conv 6, Stride 2, BN, lReLU} &
  \multicolumn{1}{l|}{} &
  \multicolumn{1}{c|}{Conv 6, Stride 2, BN, lReLU, Dropout(0.7)} &
  Self Attention \\ \cline{1-3}
\multicolumn{1}{c|}{Reshape 2048, Concat $\boldsymbol{c}$} &
  \multicolumn{1}{l|}{} &
  \multicolumn{1}{c|}{Reshape 2048, Concat $\boldsymbol{c}$} &
  Conv Transp 6, Stride 3, 128 Channels, BN, lReLU \\
\multicolumn{1}{c|}{Linear 32 (mean $\boldsymbol{z}_N$)} &
  \multicolumn{1}{l|}{} &
  \multicolumn{1}{c|}{Linear 1 (mean $\delta$), Linear 1 (logvar $\delta$)} &
  Conv Transp 6, Stride 3, 128 Channels, BN, lReLU \\
\multicolumn{1}{c|}{Linear 32 (logvar $\boldsymbol{z}_N$)} &
  \multicolumn{1}{l|}{} &
  \multicolumn{1}{c|}{Linear 1, Softplus (mean $\alpha$)} &
  Self Attention \\
\multicolumn{1}{c|}{} &
  \multicolumn{1}{l|}{} &
  \multicolumn{1}{c|}{Linear 1 (logvar $\alpha$)} &
  Conv Transp 10, Stride 2, 98 Channels \\
\multicolumn{1}{l|}{} &
  \multicolumn{1}{l|}{} &
  \multicolumn{1}{c|}{Linear 1, Softplus (mean $ndt$)} &
  \multicolumn{1}{l}{} \\
\multicolumn{1}{l|}{} &
  \multicolumn{1}{l|}{} &
  \multicolumn{1}{c|}{Linear 1 (logvar $ndt$)} &
  \multicolumn{1}{l}{} \\ \hline
\multicolumn{1}{l}{} &
   &
  \multicolumn{1}{l}{} &
  \multicolumn{1}{l}{} \\ \hline

\multicolumn{1}{c|}{\textbf{$\text{Encoder}^2_{\beta} - q^2_{\beta}\left(\boldsymbol{z}_N \mid \boldsymbol{z}_C\right)$}} &
  \multicolumn{1}{l|}{} &
  \multicolumn{1}{c|}{\textbf{$\text{Encoder}^1_{\beta} - q^1_{\beta}\left(\boldsymbol{z}_C \mid \boldsymbol{y}_{i}\right)$}} &
  \textbf{} \\

\multicolumn{1}{c|}{\textbf{maps cognitive latents to neural latents}} &
  \multicolumn{1}{l|}{} &
  \multicolumn{1}{c|}{\textbf{maps behaviors to cognitive latents}} &
  \textbf{} \\ \cline{1-3}
  
\multicolumn{1}{c|}{Linear 128, lReLU} &
  \multicolumn{1}{l|}{} &
  \multicolumn{1}{c|}{Linear 128, lReLU} &
   \\
\multicolumn{1}{c|}{Linear 128, lReLU} &
  \multicolumn{1}{l|}{} &
  \multicolumn{1}{c|}{Linear 128, lReLU} &
   \\
\multicolumn{1}{c|}{Concat $\boldsymbol{c}$} &
  \multicolumn{1}{l|}{} &
  \multicolumn{1}{c|}{Concat $\boldsymbol{c}$} &
   \\
\multicolumn{1}{c|}{Linear 64} &
  \multicolumn{1}{l|}{} &
  \multicolumn{1}{c|}{Linear 6} &
   \\ \hline
\end{tabular}
\end{table*}

\subsection{Simulation Studies}

\begin{figure}[!htb]
  \centering
    
  \begin{subfigure}{\linewidth}
    \centering
    \includegraphics[width=0.7\linewidth]{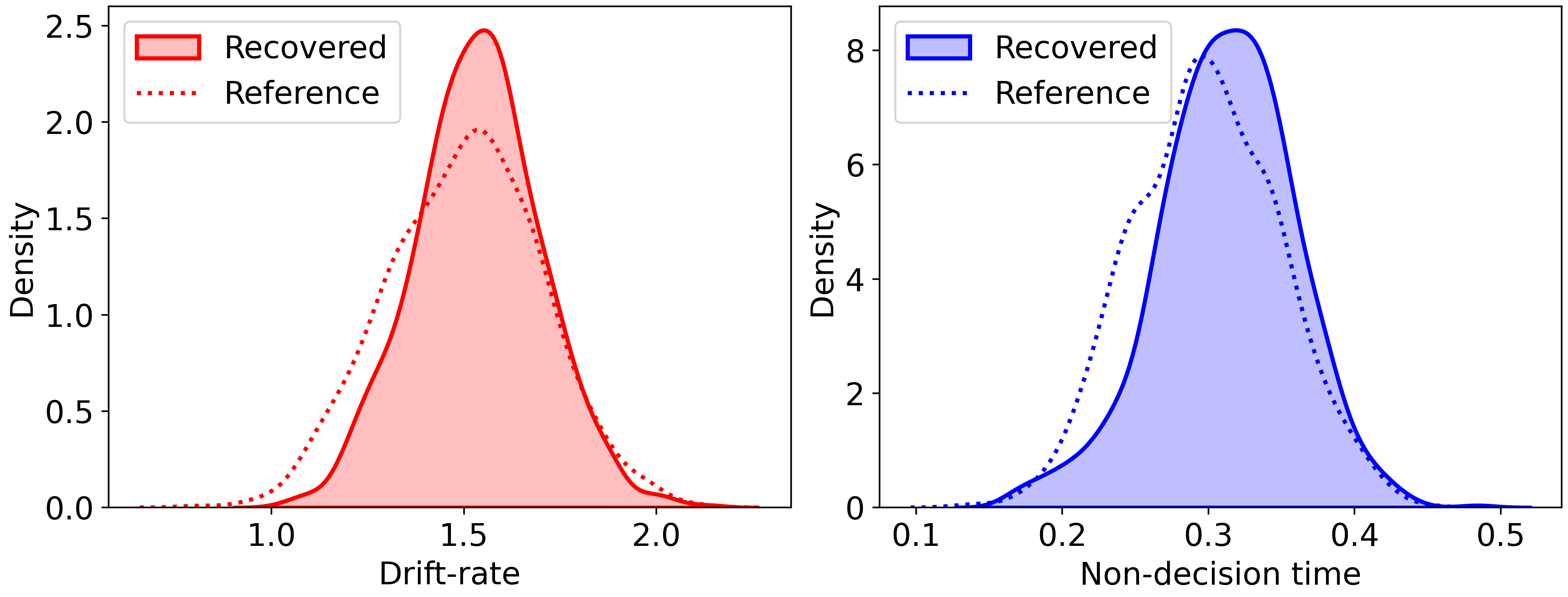}
    \subcaption{Parameter distributions}
    \label{fig:sim1}
  \end{subfigure}

  \vspace{0.4cm}

  \begin{subfigure}{\linewidth}
    \centering
    \includegraphics[width=0.7\linewidth]{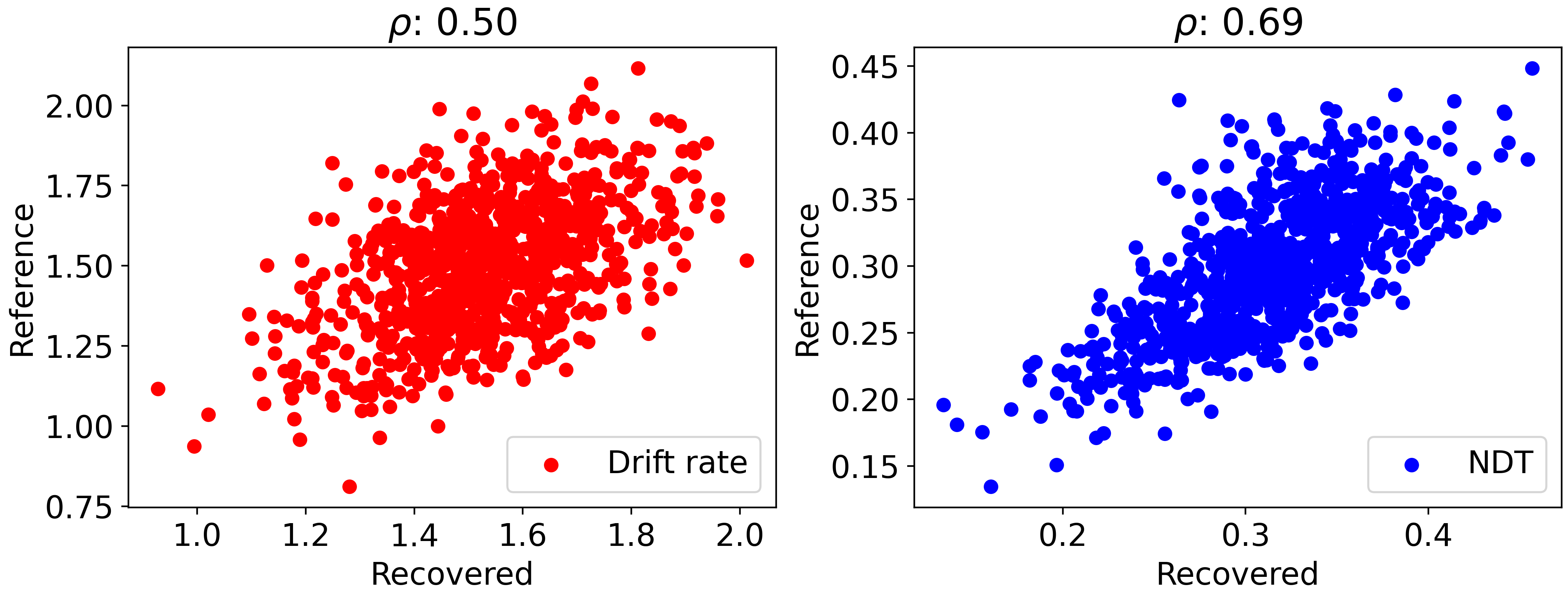}
    \subcaption{Parameter correlations}
    \label{fig:sim2}
  \end{subfigure}

  \caption{{\color{black}{Drift-diffusion parameter estimates from neural signals in a simulation of trial-level choice RTs and EEG signals. The top panels show the overlap between the recovered and the original distributions of trial-specific drift-rate and NDT. The reference values for the drift rate and NDT are drawn from the normal distributions $\mathcal{N}(1.5, 0.2)$ and $\mathcal{N}(0.3, 0.05)$, respectively. The bottom scatter plots illustrate the relationship between the recovered parameters and the original parameters each trial. $\rho$ are the Spearman correlation coefficients.}}}
  \label{fig:sim}
\end{figure}

{\color{black}{We assessed our ability to recover true non-decision time (NDT) and drift rate by simulating response time data and EEG signals. Response time data were simulated from a drift-diffusion model with trial-to-trial variability in NDT and evidence accumulation rate (i.e., drift rate). To simulate EEG signals with a known relationship with DDM parameters, we specifically focused on N200 due to the significant associations between N200 latency and NDT reported 
by \cite{nunez2019latency}.
In our new experiments, we additionally observed a substantial relationship between drift rate and N200 latency, which we included in the simulation. Boundary separation was not included in the simulation, as we did not find any neural correlates of variability in boundary separation, and those are usually only found in tasks with trial-level accuracy feedback \citep{cavanagh2014frontal, nunez2024tutorial}. 

To simulate single-trial EEG signals, we shifted the true averaged ERP waveform based on each sample of trial-level NDT, using a linear regression slope of 1, as in \citet{nunez2019latency}. EEG noise was obtained from the original data, using independently sampled segments that did not include responses to stimuli. The resulting ERP and EEG waveforms were then combined to generate artificial EEG signals for each trial that carried the N200 latency information and was associated with choice and response time. 

It is evident from the results in Figure~\ref{fig:sim} that the model can accurately recover the original distributions of trial-specific parameters. In particular, the generating and recovered distributions strongly overlap, and the correlation plots indicate that our single-trial estimates of cognitive parameters exhibit good correlations with the reference parameters.}}

\subsection{Experimental Tasks}

\citet{nunez2019latency} incorporated data from two experiments to test the hypothesis that N200 peak-latencies track Visual Encoding Time (VET). Both experiments required participants to determine whether a Gabor stimulus had high or low spatial frequency content. The tasks took place in a dark room with participants fixating on a small spot while responding to the stimuli presented on a 61 cm LED monitor.

\begin{figure}[h]
\centerline{\includegraphics[scale=0.3]{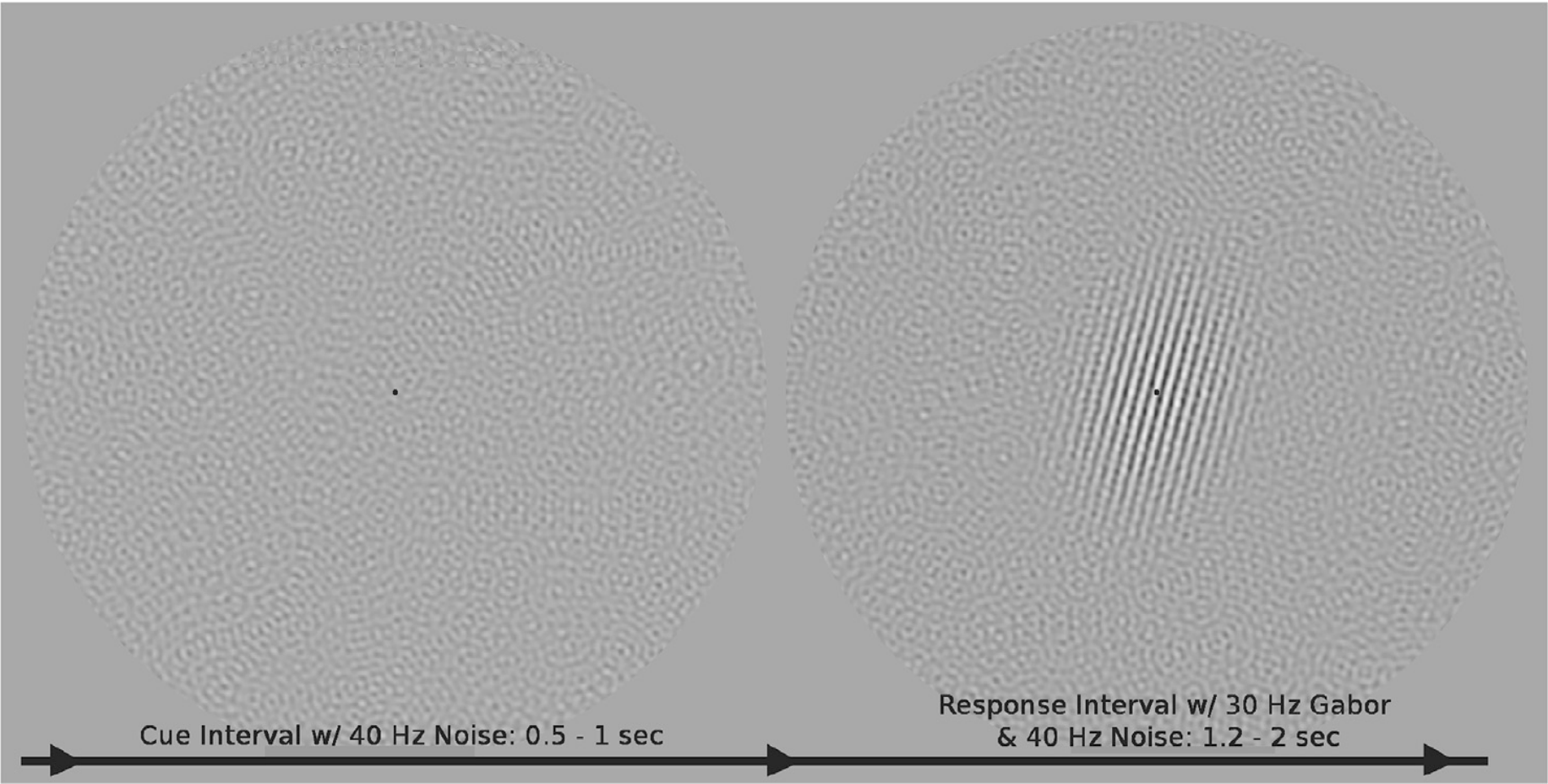}}

\caption{Example stimuli of the cue and response intervals of medium noise conditions \citep{nunez2019latency}. In the response phase, participants identified the spatial-frequency target represented by each Gabor, using their left hand to press a button for a target with a low spatial frequency (2.4 cpd) and their right hand for a target with a high spatial frequency (2.6 cpd). N200 waveforms were calculated time-locked to the onset of the Gabor stimulus during the response intervals. The visual noise altered at a frequency of 40 Hz, while the Gabor signal modulated at 30 Hz, inducing 40 Hz and 30 Hz electrocortical responses that monitor attention to both noise and signal.}

\label{fig:exp}
\end{figure}

In these experiments, Gabors were sinusoidal grating patterns with a Gaussian falloff of contrast. The high and low spatial frequencies of the target Gabors were 2.4 and 2.6 cycles per degree visual angle (cpd) respectively. The experiments involved three conditions of visual noise contrast: high, medium, and low. Visual noise was displayed both before and concurrently with the Gabor targets at regular intervals. Example stimuli are given in Figure \ref{fig:exp}. Participants used a button box to respond, pressing with the left hand for low spatial frequency targets and the right hand for high spatial frequency targets. They maintained fixation on a central spot while identifying the spatial frequency of the Gabor stimuli embedded in noise.

EEG data was recorded using a 128-channel Geodesic sensor net. The visual noise changed at 40 Hz, and the Gabor signal flickered at 30 Hz, evoking specific electrocortical responses. The primary objective was to assess whether N200 peak-latencies recorded by EEG reflected VET across varying visual noise conditions, thereby shedding light on the timing involved in perceptual decision-making processes.

\subsection{Decision-Making Models - The Drift-Diffusion Model (DDMs)}

The Drift-Diffusion Model (DDM) is a sequential sampling model of decision making. The model assumes that decision-making is the result of the accumulation of evidence in favor of one option or another. The evidence is represented by a random walk process, where the evidence accumulates over time, and the decision is made when the accumulated evidence exceeds a threshold.

\begin{figure}[h]
\centerline{\includegraphics[scale=0.5]{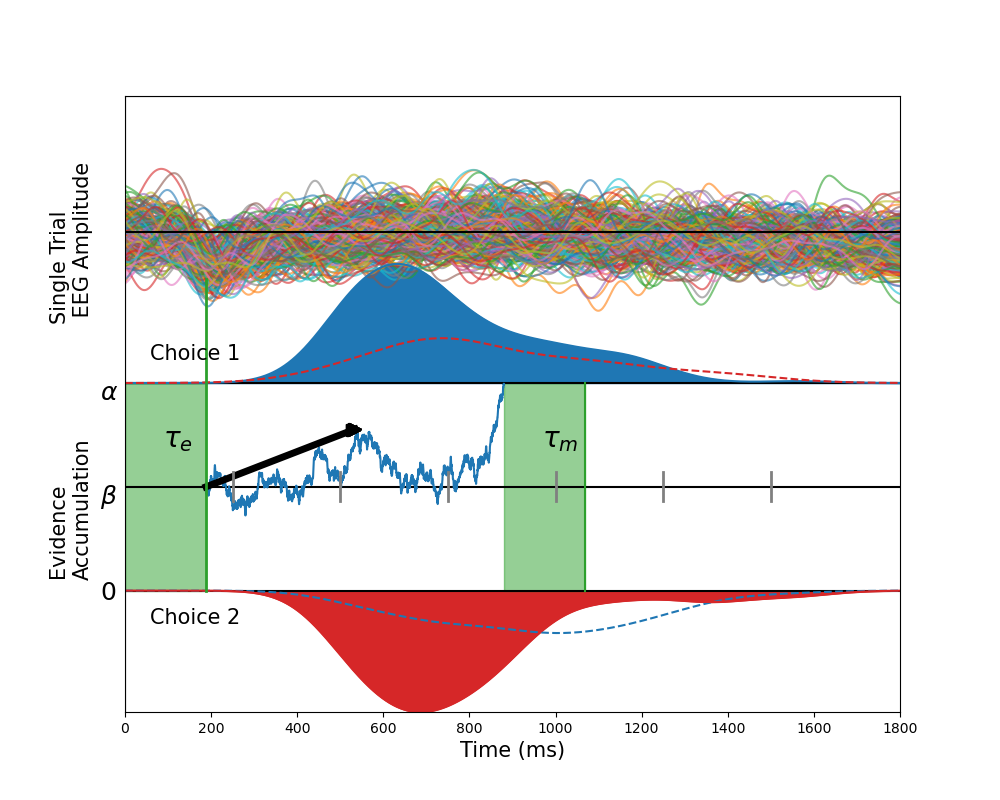}}

\caption{The DDM is illustrated in action during a two-choice task, with non-decision time shown in green. Following the visual encoding period, the decision variable (DV) begins evidence accumulation and reaches either the upper or lower limit for each trial. The black vector depicts the average rate of evidence accumulation. The blue curve depicts the distribution of response times when choice 1 is successfully picked, while the red curve depicts the distribution of reaction times when choice 2 is correctly selected. When the DV drifts towards the incorrect boundary owing to random noise, incorrect decisions are made. The distribution of reaction times for incorrect trials is depicted by the dotted curve. EEG data for each trial, processed using singular value decomposition to highlight N200, is shown on top that track the start of evidence accumulation.}

\label{fig:ddm}
\end{figure}

Mathematically, the DDM is described by a set of equations that govern the accumulation of evidence and the decision-making process. The basic equation for the DDM following a Wiener process (Figure \ref{fig:ddm}) is
\begin{equation}
dx = \delta dt + \varsigma dW
\end{equation} 

where $dx$ is the evidence step, $dt$ is the time step and $\varsigma dW$ denotes a Gaussian noise with a scale $\varsigma$. The drift rate $\delta$ and the diffusion coefficient $\varsigma$ are parameters that describe the average rise in change over a unit interval during evidence accumulation and the instantaneous variation in the rate of change, respectively. The variance can be fixed at 1 for mathematical explicitness and simplicity. The distance between two options is described as the boundary separation, or $\alpha$. The beginning position of evidence accumulation, which shows a bias toward one of the two options, is encoded by the parameter $\beta$. When $\beta$ is 0.5, the beginning point is halfway between the two borders, and the evidence-building process can begin unbiasedly between the two options. Visual encoding time prior to evidence accumulation and motor execution time following evidence accumulation could be written as $\tau_e$ and $\tau_m$, respectively. Only the total of the two processes, non-decision time $\tau$, can be observed with behavior alone.

The probability density function (pdf) of the Wiener diffusion model is bivariate (with one dimension for the latency ($t$) and one for the binary choice ($c$)); its analytical form can be approximated as
\begin{equation}
\left\{\begin{aligned}
& \text { Wiener }(t, c=0 ; \alpha, \beta, \tau, \delta) \\ & = \frac{\pi}{\alpha^2} e^{-\frac{1}{2}\left(2 \alpha \beta \delta+\delta^2(t-\tau)\right)} \times \sum_{k=1}^{+\infty}\left[k \sin (\pi k \beta) e^{-\frac{1}{2} \frac{k^2 \pi^2}{\alpha^2}(t-\tau)}\right] \\
& \text { Wiener }(t, c=1 ; \alpha, \beta, \tau, \delta) \\ & =  \text { Wiener }(t, c=0 \mid \alpha, 1-\beta, \tau,-\delta)
\end{aligned}\right.
\end{equation} 
Efficient methods for the computation of the Wiener diffusion model density and distribution functions exist \citep{navarro2009fast}, making it a highly tractable model.
In this work, $\beta$ is set at 0.5, so that the starting point is always unbiased at $z$ = $\beta \alpha$.


\chapter{Conclusions}

\section{Contributions}

The growing accessibility of medical time series data is propelling the advancement of mathematical models and techniques that can analyze it broadly and efficiently. This dissertation is a move towards this goal, motivated by the recent achievements in probabilistic modeling and deep learning.

In this dissertation, we have constructed integrated frameworks that combine concepts from latent variable models, state-space models, and deep learning to model multidimensional dependencies in physiological signals. These frameworks are capable of modeling complex data distributions across various applications. This is accomplished through the development of probabilistic models that utilize deep neural networks to parameterize the underlying conditional distributions. Deep learning architectures serve as powerful function approximators, enabling the model to automatically extract features essential for wide applicability. Recent developments in deep learning can be seamlessly integrated into this framework.

The approaches proposed in this dissertation offer versatile frameworks for representing and learning from diverse types of physiological measures. By efficiently processing unlabeled datasets, these models enable the discovery of hidden structures and patterns, facilitating data-driven hypothesis generation:

\begin{enumerate}
\item \textit{Deep State-Space Model for Heart Electrical Waveforms} \citep{vo2023ppg}. This application has significant potential for clinical diagnoses, especially since it allows for heart disease assessment through wearable devices. The use of optically obtained signals as inputs adds to the innovation, potentially making diagnosis simpler and more accessible.
\item \textit{Brain Signal Modeling with Probabilistic Graphical Models and Deep Adversarial Learning} \citep{vo2022composing}. Combining these two approaches is promising for encoding neural oscillations' complexity while maintaining interpretability. Moreover, applying these techniques to epilepsy seizure detection as an unsupervised learning problem could lead to earlier and more accurate diagnoses, improving patient outcomes.
\item \textit{Joint Modeling of Physiological Measures and Behavior} \citep{vo2024deep}. This approach shows potential in tackling the modern challenge in amalgamation  of diverse medical data sources. By analyzing the relationship between physiological measures and behavior, our method could uncover new insights into brain function and potentially revolutionize our understanding of neurocognitive processes.
\end{enumerate}

\section{Future Work}

Every progression in a field brings forth a set of unanswered queries, usually more complex than the ones preceding them. Regarding the deep latent variable models (DLVMs) presented in this thesis, several open questions exist, the resolution of which could enhance our comprehension of their working principles as well as to better exploit their modeling power.

\begin{itemize}
\item Ensuring patient safety when implementing medical machine learning methods for clinical applications necessitates robust models. A key characteristic of robust models is their resilience to out-of-distribution (OOD) data, meaning they can still provide accurate predictions when encountering data that differ from the training set. Such OOD data might include samples from varied patient demographics, different medical equipment and laboratory methods. Future research should focus on assessing the adaptability of the DLVMs to OOD data and enhancing model resilience against such data by leveraging established strategies in the field \citep{zhou2022domain, wang2022generalizing}.

\item Probabilistic graphical models offer a principled approach to incorporating prior knowledge and structured frameworks into the model, utilizing current message-passing algorithms for approximate inference. A crucial yet difficult task for broader adoption of DLVMs is developing a message-passing library that seamlessly works with existing deep learning libraries \citep{johnson2016composing, bendekgey2024unbiased}.

\item Choosing the optimal model parameterization for a specific application can be challenging. This category of models derives all the complexities associated with defining the precise network architecture, such as the number of layers, units, and activation functions, from its deep learning components. Therefore, identifying a systematic method for hyperparameter optimization that is effective across different applications is essential \citep{yu2020hyper, he2021automl}.

\item  Recent progress in physics-informed deep learning \citep{raissi2019physics, nabian2020physics} integrates the advantages of deep learning methods with physical principles to improve both model efficacy and generalization. In this approach, deep learning models are enhanced with a regularization term that serves as prior knowledge, reflecting the fundamental laws and penalizing deviations from these governing equations. Investigating DLVM approaches that adhere to any specified law of electrophysics, as characterized by stochastic differential equations, would be advantageous.

\end{itemize}

\clearpage
\phantomsection

\bibliographystyle{apalike}
\bibliography{thesis}



\end{document}